\documentclass[10pt,letterpaper]{article}
\usepackage[top=0.85in,left=2.75in,footskip=0.75in]{geometry}

\usepackage{amsmath,amssymb}

\usepackage{changepage}

\usepackage[utf8x]{inputenc}

\usepackage{textcomp,marvosym}

\usepackage{cite}

\usepackage{nameref,hyperref}

\usepackage[right]{lineno}

\usepackage{microtype}
\DisableLigatures[f]{encoding = *, family = * }

\usepackage[table]{xcolor}

\usepackage{array}

\usepackage{rotating}

\usepackage{float}

\newcolumntype{+}{!{\vrule width 2pt}}

\newlength\savedwidth

\raggedright
\usepackage[aboveskip=1pt,labelfont=bf,labelsep=period,justification=raggedright,singlelinecheck=off]{caption}

\makeatletter
\renewcommand{\@biblabel}[1]{\quad#1.}
\makeatother

\usepackage{lastpage,fancyhdr,graphicx}
\usepackage{epstopdf}
\fancyheadoffset[L]{2.25in}
\fancyfootoffset[L]{2.25in}
\begin{document}	
\vspace*{0.2in}

\graphicspath{{./figures-pace-corrected-png/}}

\begin{flushleft}
{\Large
\textbf\newline{A Statistical Model of Word Rank Evolution}
}
\newline
\\
Alex John Quijano\textsuperscript{1*},
Rick Dale\textsuperscript{2},
Suzanne Sindi\textsuperscript{1},
\\
\bigskip
\textbf{1} Applied Mathematics, University of California Merced, Merced, California, USA
\\
\textbf{2} Department of Communications, University of California Los Angeles, Los Angeles, California, USA
\\
\bigskip

%
%





* aquijano4@ucmerced.edu \\

\end{flushleft}





\section*{Abstract}
The availability of large linguistic data sets enables data-driven approaches to study linguistic change. The Google Books corpus unigram frequency data set is used to investigate the word rank dynamics in eight languages. We observed the rank changes of the unigrams from 1900 to 2008 and compared it to a Wright-Fisher inspired model that we developed for our analysis. The model simulates a neutral evolutionary process with the restriction of having no disappearing and added words. This work explains the mathematical framework of the model - written as a Markov Chain with multinomial transition probabilities - to show how frequencies of words change in time. From our observations in the data and our model, word rank stability shows two types of characteristics: (1) the increase/decrease in ranks are monotonic, or (2) the rank stays the same. Based on our model, high-ranked words tend to be more stable while low-ranked words tend to be more volatile. Some words change in ranks in two ways: (a) by an accumulation of small increasing/decreasing rank changes in time and (b) by shocks of increase/decrease in ranks. Most words in all of the languages we have looked at are rank stable, but not as stable as a neutral model would predict. The stopwords and Swadesh words are observed to be rank stable across eight languages indicating linguistic conformity in established languages. These signatures suggest unigram frequencies in all languages have changed in a manner inconsistent with a purely neutral evolutionary process.


\section*{Author summary}
Dr. Alex John Quijano completed his PhD in Applied Mathematics (2021) at the University of California Merced. Dr. Suzanne Sindi is a professor in Applied Mathematics at the University of California Merced. Dr. Rick Dale is a professor in the Department of Communications at the University of California Los Angeles.



\section*{Introduction}

Natural languages are deeply connected to human culture. There is no doubt that the subtleties of word usage and evolution in a language are very difficult to understand and predict. Words can naturally gain or lose meanings in time, or words can change their meaning entirely. Important historical events may serve as symbolic beginnings of cultural change that we can now see ``echo'' from them in word frequency changes. The term ``gay" has changed its meaning from ``cheerful", ``happy", or ``joyful" to many contexts connected to homosexuality, and the word ``broadcast" has changed its meaning from ``spreading seeds for sowing" to ``spreading radio waves to convey information" \cite{Li2019,Bamler2017,hamilton2016diachronic}. Although the term ``gay" retains some of its original meanings, it has become sexualized as a result of societal shift over the last half-century. Similarly, the word ``broadcast" has kept its basic meaning of ``spreading" but has altered in context as a result of technological advancements. Another example includes the word ``risk". The contextual meaning of the word ``risk" changed from negative emotion words such as ``danger", ``fear", and ``hazard" to more positive emotion words such as ``prevalence" and ``prevention". The word still retains its objective meaning of ``the presence and exposure of danger" but the context has changed from the presence of threats to a more scientific context of prevention of threats \cite{Li2020}. Other examples include ``cell", ``car", ``monitor", ``nuclear", and ``option" \cite{Li2019}. The examples above are English words known to have been used frequently in time allowing their meanings to evolve. Roughly speaking, the frequency and their ranks showed us some insights into the evolution of word meanings. The changes in word ranks are influenced by a multitude of factors. It has been hypothesized that languages undergo evolutionary pressures to adapt to social systems similar to organisms evolving to fit into their environments \cite{Bentz2018,Tamariz2016,lupyan2010language}. Studies have also shown that language adapt to to their environment through the process of iterated learning \cite{Smith2020,Winters2015} and through language transmission \cite{Pagel2019,newberry2017detecting,Hammarstrom2016,Crema2016,Acerbi2014}. In this work, we use statistical modeling of word rankings to investigate the mechanism of language evolution.
	
Pagel et al. (2007) \cite{Pagel2007} and (2019) \cite{Pagel2019} demonstrated that choosing words requires more than just repeating what others have said. Language speakers appear to have a bias that drives them to employ words that are used disproportionately more frequently by the majority of people. Words might be pushed out, allowing a single word to dominate all others. Pagel et al. (2019) used this explanation to show how languages organize themselves and stay reasonably stable. Word frequency is a measurement of word popularity while word ranking is a stable measure of popularity simplifying the complexity of language dynamics. The relationship between frequencies and ranks is inversely proportional and has been observed in many naturally occurring systems such as networks, language, and genes \cite{Iniguez2021,Ghoshal2011}. In this study, we explore how word ranks change in time while exploring the conditions for which words remain stable or become volatile.

Despite the complicated nature of written language , the frequency of word usage follows a pattern. Popularized by George Kingsley Zipf, word frequencies on a large body of text follow a power-law $r \propto 1/k$ where $r$ is the relative frequency and $k$ is the rank \cite{Zipf1935, Zipf1950}. Until recently, historical text was challenging to work with. However, the Google Ngram data \cite{Lin2012} provides historical $n$-gram sequences of $n$ words from digitized text of the Google books corpus. We observed from the Google unigram ($1$-gram) data that the ranks of words change in time. While word frequencies may increase/decrease in time, the ranks corresponding to those frequencies may remain stable. With the inspiration of molecular evolutionary biology, the statistical model presented in this study offers an explanation of the volatility/stability conditions of word rank evolution.

\begin{figure}[!htbp]
	\centering
	\caption[The Wright-Fisher inspired model.]{\textbf{The Wright-Fisher inspired model.} Words (shown as squares) at time $t+1$ chosen - with replacement - from the previous time $t$ assuming three words in the vocabulary with increasing corpus size. Different colored squared represents different words. Because we seek to model the evolution of stable words, we require words to be sampled at least once which results in constant number of vocabulary words for all $t$.}
	\label{fig:wright-fisher-model}
	\includegraphics[width=\textwidth]{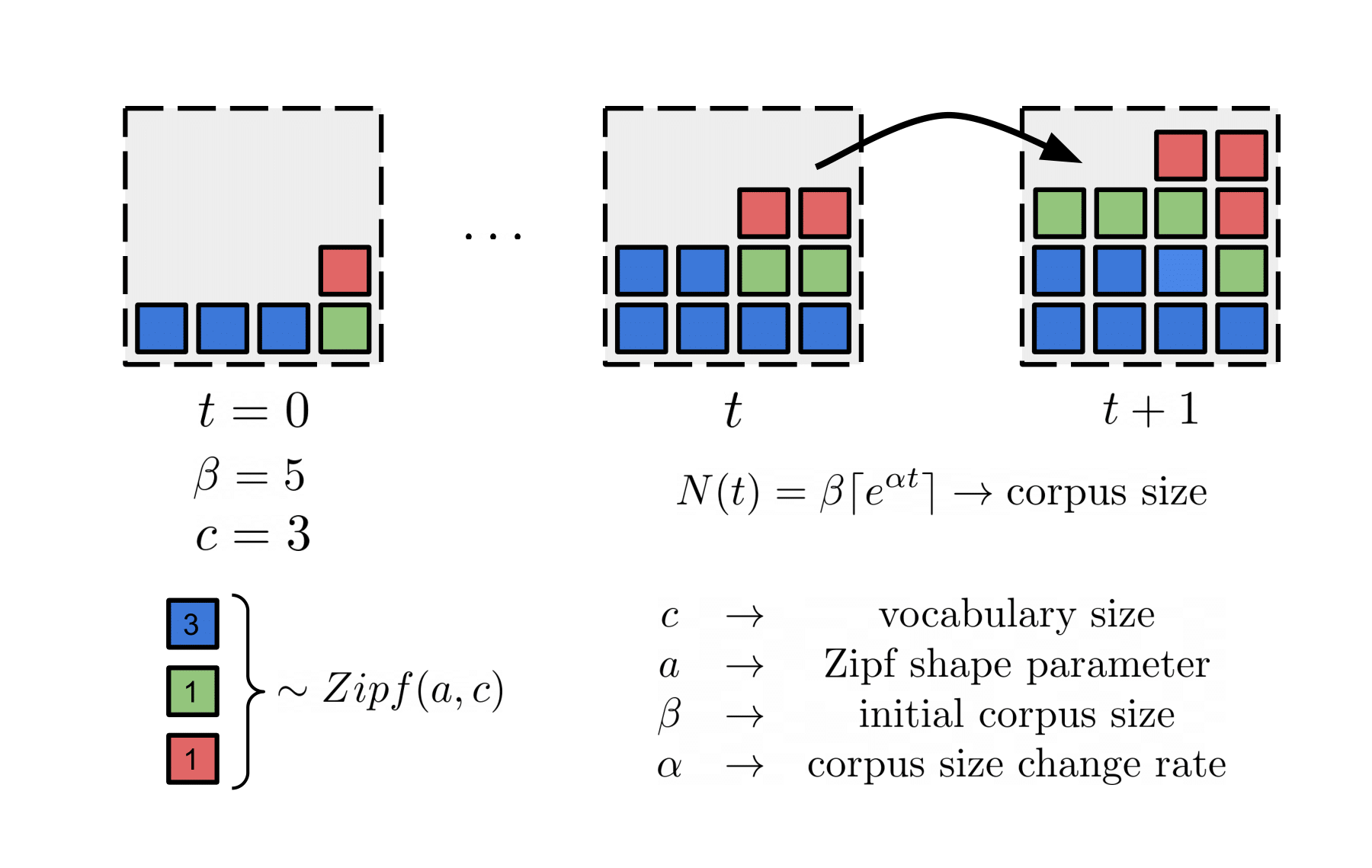}
\end{figure}

We develop a Wright-Fisher (WF) inspired neutral model as a base model to the Google unigram time-series data. We utilize this model to better understand how languages change through time. This neutral model is inspired by the theory of molecular evolution by Kimura \cite{Kimura1983} and the Wright-Fisher model \cite{yang2014molecular,Ewens2012}. A version of this model was previously presented by Sindi and Dale \cite{Sindi2016} who simulated the evolution of a growing corpus where each word in the vocabulary has equal fitness and chosen from the previous generation at random. This type of model is largely dominated by drift. Drift is a neutral process of evolution that is dominated by frequency dependent sampling. The outcome of the word frequency at the current generation is determined entirely by chance and not by other evolutionary forces like mutation, migration, natural selection, and random ``mating". The model assumes that the number of vocabulary words stays the same, meaning words are not added nor removed in time, and the initial frequency distribution follows Zipf's law. It also assumes the corpus size grows exponentially. In Fig.~\ref{fig:wright-fisher-model}, we illustrate this process using blue, green, and red squares as words. The model has four parameters that can be adjusted which are the vocabulary size $c$, Zipf shape parameter $a$, initial corpus size $\beta$, and the corpus size change rate $\alpha$.

Previous studies have used the Google Ngram dataset and other linguistic corpora to study and model how word frequencies evolve in time. Turney et al. (2019) \cite{turney2019natural} demonstrated that language change is not random, but natural selection is the main driver for how language changes. For instance, there is a rise in concreteness in American English where specific words and phrases are used to minimize vagueness \cite{Hills2016,Hills2015}. Concreteness is particularly important when communicating because they have generally have stable meanings. Non-concrete words such as emotion words are generally more abstract. Recent studies have shown that there is a decline in the use of positive emotion words in fictional works written in American and British English \cite{Morin2016,Acerbi2013}. We can speculate that people tend to resonate with emotions in fiction because they relate it in real life. Even with the decline of emotion word usage in fictional writing, there is a universal positivity bias consistent in ten languages across different sources - in books and in social media \cite{Dodds2015}. A study have shown that the rate of change of word usage is decreasing overtime in the English language, meaning that some words - especially concrete words - are approaching a steady-state \cite{petersen2012languages}. However, recent studies have shown that - along with some words approaching stability in frequency - there is a rise of novel words which contributes to a rise entropy in language \cite{Sun2021,Pilgrim2021}. The rise of concreteness, rise in entropy, and the decline of the usage of emotion words are evident of selection in language - at least hypothesized. In this study, we attempt to understand these behaviors through statistical modeling of word ranks under the assumption of neutral evolution.

Previous studies have modeled the rank diversity of words. Cocho et al. \cite{Cocho2015} developed a Gaussian random walk model and compared it to the Google unigram dataset of six European languages. They found that the ranks of words in their experiments approach a log-normal distribution and the diversification of word ranks is the result of the random walks. Morales et al. \cite{Morales2018} continued the rank diversity analysis to include n-grams where $n > 1$. For six European languages, they found that it is important to consider studying languages at higher scales. Bigrams pose challenges, considering standard approaches to automatically quantifying word meanings through \emph{document} statistics\cite{deerwester1990indexing} and analyzing meaning changes through machine learning \cite{Bamler2017,hamilton2016diachronic}.

In this study, we analyze unigram dataset of a century's worth of word rank changes in several languages. These languages are English, Simplified Chinese, French, German, Italian, Hebrew, Russian, and Spanish. We also consider three other variants of English which are American English, British English, and English Fiction. These languages are eight of the most spoken and written language in the world, though admittedly they are a small subset of the world’s thousands of languages. Nevertheless, they also represent several language families (Indo-European, Sino-Tibetan, Afroasiatic). There are many perspectives of studying language evolution which includes grammatical variations, sentence structure ordering (e.g. subject-verb-object), pronunciation variations, number of speakers, word frequencies, and other linguistic concepts that explains how language works \cite{Hammarstrom2016}. We seek to characterize word frequency changes of our chosen languages using word-rank statistics. 

We use unigram time-series similar to the studies involving word frequency change \cite{Younes2019, Koplenig2018, Ruck2017, koplenig2017impact, montemurro2016coherent, Pechenick2015, kulkarni2015statistically, Kim2015, Bentley2014, Acerbi2013, Goldberg2013, Bentley2012, Michel2011}, building on our prior work which, like Turney et al., uses Google unigram as a testbed for devising new tests of selection in the cultural domain \cite{Sindi2016}. This study uses the Swadesh words and stopwords - set of words with stable ranks or stable meanings - to perform a comparative analysis between the eight languages. We first develop a neutral model of frequency change and, consistent with past studies \cite{Pagel2019, turney2019natural, newberry2017detecting, o2017inferring, Sindi2016, Blythe2012, Reali2010, Pagel2007}, we observe departure from neutrality in all languages we study. We then generalize our neutral model to consider what we believe to be a \emph{minimal} model of linguistic change: a word's frequency changes as a function of its previous frequency and the frequencies of other closely related words. This work also explains the mathematical framework of the model to answer why words change ranks in time. By using the WF inspired model we articulate several of the volatility/stability conditions for word-frequency change and show that it is surprisingly consistent across several languages. The result shows that \emph{word community} matters, not just frequency - how a word is situated in the entire space of possible in a language gives it a position in a landscape. Some regions of this landscape may introduce capacity for semantic change, and other regions may restrict it.

\section*{Methods}

\subsection*{The Wright-Fisher (WF) inspired model}

The Wright-Fisher model is an evolutionary model in molecular biology that simulates genetic drift, an evolutionary process of random sampling of alleles between discrete generations \cite{Ewens2012}. Our model considers words to be alleles and samples words in discrete generations with an increasing corpus size. As shown in Fig.~\ref{fig:wright-fisher-model}, words are sampled from the previous generation according to their frequency until we sample sufficiently many words as specified by the corpus size. Our model assumes fixed vocabulary words in time, meaning that the number of distinct words is always constant as corpus size increases. The initial distribution of word frequencies is chosen according to the Zipf distribution and we track the raw counts of the words, their empirical probabilities, and corresponding word ranks.

A rank of 1 refers to the most frequent word. Throughout the paper, we often use the term ``high rank" which refers to word ranks close to the rank of 1 while the use of ``low rank" refers to the word ranks close to the value $c$. In other words, the use of ``high rank" means that a word is high up in rank similar to the meaning of ``high ranking institution" which indicates a low valued number. In addition, we also use in this paper the term ``go up in rank" which refers to the change of ranks from low rank to high rank while ``go down in rank" means the opposite.

Unlike the traditional Wright-Fisher model, we do not allow words (alleles) to disappear. Because we seek to model the variation in the frequencies (ranks) of stable words we ensure that each word is sampled at least once. In the following, we explain the theoretical properties of our Wright-Fisher inspired model and the quantities we follow in time.

It is known that most of the natural languages follow Zipf’s law \cite{Piantadosi2014,Baixeries2013,petersen2012languages,Manin2009}. The general assumption of the shape parameter of the Zipf distribution is assumed to be $1$ given large text data \cite{Baixeries2013} and the WF model only yields Zipf's Law under specific parameter values \cite{o2017inferring}. However, it is worth mentioning that the value of this parameter is not fixed, and it is shown that it varies with time and with linguistic complexity \cite{Baixeries2013, Koplenig2018}. It has been shown by Ruck et al. \cite{Ruck2017} that their neutral model of vocabulary change replicates Zipf's law. If the system follows Zipf's law, then Heap's law can be considered where the size of the corpus has a non-linear relationship with the number of vocabulary words \cite{Lu2010}. However, Heap's law cannot be replicated using the WF model \cite{Ruck2017}. Their model has to be modified where - instead of sampling from the previous generation - they sample from \emph{all} previous generations to replicate Heap's law. This work uses the word frequency distribution at the initial time point to estimate the shape parameter of Zipf's law. There are other research works where they discussed fitting the curves of the log-transformed Zipf with a more generalized Zipf distribution called the Zipf-Mandelbrot distribution \cite{Manin2009, Piantadosi2014}.

As noted above, Zipf’s law states that the word rank is inversely proportional to its frequency. More specifically, for a set of vocabulary words $V = \{w_1, w_2, \cdots, w_c\}$ of $c$ words and with corresponding ranks $K = \{k_1, k_2, \cdots, k_c\}$ of the same length, the probability mass function for the Zipf distribution is given by
\begin{equation}
	P(Y = k_w; a, c) = \frac{1/k_{w}^a}{\sum_{w=1}^{c} (1/k_{w}^a)}
	\label{eq:zipf-pmf}
\end{equation}
where $a \ge 0$ is the power law parameter characterizing the distribution and $Y$ is the discrete random variable of the word rank of a randomly sampled word from a corpus. The Zipf distribution is used to compute the initial probability for each word sample at time $t = 0$. That is, each word in $V$ is sampled randomly based on its probability from the Zipf distribution. Words are sampled until we reach the corpus size. The corpus size $N$ at time $t + 1$ increases exponentially at rate $\alpha \ge 0$ and initial corpus size $\beta \ge c$ is given by
\begin{equation}
	N(t;\alpha,\beta) = \beta \left\lceil e^{\alpha t} \right\rceil.
	\label{eq:corpus-size-function}
\end{equation}
The rate of an exponentially increasing corpus size has been estimated and applied to similar studies by Ruck et al. (2017) \cite{Ruck2017} and Petersen et al. (2012) \cite{petersen2012languages}. We incorporate Eq.~\ref{eq:corpus-size-function} to model the sum of all occurring words in the corpus.

\subsection*{Zipf's law in relation to the corpus size}

Zipf's law (a power law) is defined in our problem as
\begin{equation}
	r_{w} \propto \frac{1}{k_{w}^{a}}, \hspace{10px} r_{w} \ge 1
	\label{eq:power-law}
\end{equation}
where $k_{w}$ is the rank of word $w$ and $r_{w}$ is the raw frequency at the initial time \cite{Clauset2009,Newman2005,Zipf1935,Zipf1950}. It means that the most frequent word $k_1=1$ has $r_1 \propto 1$. The Zipf probability mass function shown in Eq~\ref{eq:zipf-pmf} is used to compute the probabilities of words. At the initial time, we sample the words according to those probabilities. The expected value (or the expected rank of a randomly sampled word) is given by
\begin{equation}
	E[Y] = \sum_{w=1}^{c} k_{w} P(k_{w}; a, c) = \frac{\sum_{w=1}^{c} (1/k_{w}^{a-1})}{\sum_{w=1}^{c} (1/k_{w}^a)}.
	\label{eq:zipf-expected-value}
\end{equation}
where $Y$ is the discrete random variable of the word rank of a randomly sampled word from a corpus. When sampling from the Zipf distribution, it is possible that some words will have the same proportions. It means that some words can have the same rank but to simplify the problem, the ranks of the words are unique. For example, two words with the same raw frequency have unique ranks in sequence.

Since $\beta$ is the initial corpus size, the sample proportion of a word $w$ given its raw frequency $r_{w}$ at the initial time is given by
\begin{equation}
	\hat{p}_{w} = \frac{r_{w}}{N(0;\alpha,\beta)} = \frac{r_{w}}{\beta} \hspace{10px} \text{where} \hspace{10px} \beta = \sum_{w=1}^c r_{w}.
	\label{eq:zipf-data-pmf}
\end{equation}
Then, the sample mean word rank at the initial time is written as
\begin{equation}
	\bar{k} = \sum_{w=1}^{c} r_{w} \frac{r_{w}}{\beta}.
	\label{eq:zipf-data-expected-value}
\end{equation}

There are bound to be some differences between the expected values because of noise when sampling. Therefore, the difference can be characterized by
\begin{equation}
	E[Y] - \bar{k} = \frac{\sum_{w=1}^{c} (1/k_{w}^{a-1})}{\sum_{w=1}^{c} (1/k_{w}^a)} - \sum_{w=1}^{c} r_{w} \frac{r_{w}}{\beta}
	\label{eq:zipf-pmf-data-expected-value-difference}
\end{equation}
where as $\beta$ increases the $E[Y] - \bar{k} \to 0$ for fixed $c$, where it means that the mean rank $\bar{k}$ approaches to the expected rank $E[Y]$ by the law of large numbers. If $c$ increases while $\beta$ is fixed ($\beta > c$), then $r_{w} \to 1$ for all $w$. If $\beta=c$, then $r_{w} \propto 1$ for all $w$.

\subsection*{The Wright-Fisher (WF) inspired model written as multinomial probability transitions}

Due to the stochastic nature of our model, the randomness of word count evolution and its processes can be thought of as a Markov chain with multinomial probability transitions. The mathematical foundations of classical WF model is based on the binomial probability mass function, which is a reduced form of the multinomial probability mass function \cite{Ewens2012}. Kandler et al. (2013) \cite{Kandler2013} incorporated the binomial into their version of the Wright-Fisher model to detect departures from neutrality of evolving pottery decorations. They further generalized their model using a multinomial approach to infer selection \cite{Kandler2015}. These are examples of modeling cultural transmission - not of language specifically. Here, we use the multinomial approach in our WF inspired model as means to explain the rank changes observed in the unigram frequency data.

Consider the state space $\mathbf{\vec{r}}_t \in \left\{ r_{w,t} \in \{1,2,\cdots,N_{t}-1\} \text{ for } w = \{1,2,\cdots,c\} \right\}$ where $\mathbf{\vec{r}}_t$ is a vector of word raw frequencies at time $t$, $c$ is the number of vocabulary words, and $N_{t}$ is the corpus size at time $t$. We know that $\sum_{w=1}^{c} r_{w,t} = N_{t}$. The corpus size $N_{t}$ increases exponentially at rate $\alpha \ge 0$ with initial corpus size $\beta \ge c$ which is written in Eq~\ref{eq:corpus-size-function}.

The probability of transitioning from $\mathbf{\vec{r}}_{t-1} = (a_1,a_2,\cdots,a_c)$ to $\mathbf{\vec{r}}_{t} = (b_1,b_2,\cdots,b_c)$ is given by the multinomial probability mass function,
\begin{multline}
	Mult\left( \mathbf{\vec{r}}_{t} = (b_1,b_2,\cdots,b_c) | \mathbf{\vec{r}}_{t-1} = (a_1,a_2,\cdots,a_c) \right) = \\ \frac{(N_{t}-c)!}{b_1! b_2! \cdots b_c!} \left( \frac{a_1}{N_{t-1}} \right)^{b_1} \left( \frac{a_2}{N_{t-1}} \right)^{b_2} \cdots \left( \frac{a_c}{N_{t-1}} \right)^{b_c}.
	\label{eq:multinomial-probability-pmf}
\end{multline}
with initial state $\mathbf{\vec{r}}_{0} \sim Zipf(a)$ where $a \ge 0$ is the Zipf shape parameter (See Eq~\ref{eq:zipf-pmf}).

The multinomial random vector has components with a binomial distribution for frequency $r_{w,t}$ in $\mathbf{\vec{r}}_{t}$,
\begin{equation}
	\begin{split}
		R_{1,t} & \sim Bin{\left(N_t,\hat{p}_{w,t-1}\right)} \\
		R_{2,t} & \sim Bin{\left(N_t,\hat{p}_{w,t-1}\right)} \\
		& \vdots \\
		R_{c,t} & \sim Bin{\left(N_t,\hat{p}_{w,t-1}\right)}
	\end{split}
	\label{eq:binomial-components}
\end{equation}
where $R_{w,t}$ is the discrete random variable of a word's raw frequency and $\hat{p}_{w,t-1} = \frac{r_{1,t-1}}{N_{t-1}}$, which is the sample proportion of word $w$ at $t-1$. The binomial probability mass function is given by
\begin{multline}
	Bin(R_{w,t}=r_{w,t}; N_t, N_{t-1},c) = \binom{N_t-c}{r_{w,t}} \left( \hat{p}_{w,t-1} \right)^{r_{w,t}} \left(1 - \left( \hat{p}_{w,t-1} \right) \right)^{N_t - r_{w,t}}.
	\label{eq:binomial-pmf}
\end{multline}

Since the corpus size is prescribed in our WF model, the counts of words are weakly dependent on one another. However, the counts of any word taken alone are binomial. The binomial probability mass function at the initial time is given as
\begin{equation}
	Bin(R_{w}=r_{w}; \beta, c) = \binom{\beta-c}{r_{w}} p_w^{r_{w}} \left(1 - p_w \right)^{\beta - r_{w}}.
	\label{eq:binomial-pmf-initial}
\end{equation}
where the value of the probability $p_w$ is from the Zipf probability mass function in Eq. \ref{eq:zipf-pmf}. It is important to note that even though the word frequencies can be independently divided into their binomial components, the overall behavior of the Wright-Fisher model is heavily dependent on the corpus size and the vocabulary size.

The expected value and the variance for the binomial probability mass function are given as follows:
\begin{eqnarray}
	E[R_{w,t}] & = & (N_t-c)p_{w,t-1} \hspace{10px} \text{and} \\
	Var[R_{w,t}] & = & (N_t-c)p_{w,t-1}(1-p_{w,t-1}).
	\label{eq:binomial-expected-value-and-variance}
\end{eqnarray}
The expected value and variance are heavily dependent on the probability and the corpus size function. Since it is defined that the corpus size function is an exponentially increasing function in time shown in Eq.~\ref{eq:corpus-size-function}, then the expected value and variance also increases in time.

\subsection*{Rank change potential}

A discrete binomial distribution can be approximated by a normal distribution with mean $\mu_{w,t} = (N_t-c)p_{w,t}$ and variance $\sigma_{w,t}^2 = (N_t-c)p_{w,t}(1-p_{w,t})$. Approximately 100\% of the distribution lies within the interval $\mu_{w,t} \pm 4\sigma_{w,t}$. Note that $\sum_{w=1}^c p_w = 1$ where the values of $p_w$ came from the Zipf probability mass function in Eq.~\ref{eq:zipf-data-pmf} with shape parameter $a$ and number of vocabulary words $c$. The interval is the segment where a word is most likely to have a frequency when randomly sampled. These segments can overlap, and so it becomes likely that words can change ranks. We count the number of overlaps by computing the length of this segment overlap. The following is how we compute the length of the overlap of two segments for word $w$ and word $v$:
\begin{equation}
	l_{wv,t} = min{\left( \mu_{w,t}+4\sigma_{w,t} , \mu_{v,t}+4\sigma_{v,t} \right)} - max{\left( \mu_{w,t}-4\sigma_{w,t} , \mu_{v,t}-4\sigma_{v,t} \right)}.
	\label{eq:intersection-length}
\end{equation}

Let $r_w$ and $r_v$ be the ranks of word $w$ and $v$ respectively. By counting the number of words that overlap word $w$, the net potential rank change is given by
\begin{equation}
	\widehat{ol}_{w} = \sum_{v=1,v \ne w}^c
	\begin{cases}
		1 & l_{wv} > 0 \text{ and } r_w < r_v \\
		-1 & l_{wv} > 0 \text{ and } r_w > r_v \\
		0 & otherwise.
	\end{cases}
	\label{eq:overlap}
\end{equation}

In other words, the net number of words that overlap with word $w$ is by summing the number of overlaps above and below its word rank. If $\widehat{ol}_{w} > 0$, then word $w$ has the potential to go down in rank. If $\widehat{ol}_{w} < 0$, then word $w$ has the potential to go up in rank.

\subsection*{The Google unigram data}

A subset of historical word frequency time-series data was taken from the Google Ngram Corpus, a collection of digitized books summarized by counting n-grams. The Google n-gram data spans from 1-gram (or unigram) to 5-grams, but we only consider the unigram data for this dissertation. The database provides unigram (single-word) occurrences for over one hundred years in eleven languages (Four of these eleven languages are variations of the English language.) In total, there are eight distinct languages in the data set used here: Chinese, English, French, German, Hebrew, Italian, Russian, and Spanish. Google parsed millions of books from the Google books corpus and counted how many times a unigram occurred in a given year. These data sets were generated in 2012 (version 2) and 2009 (version 1). Version 2 includes parts-of-speech annotations \cite{Lin2012}. See \nameref{S1_Appendix} for the data source and easy access of the data.

To minimize bias in the data set, we attempted to select common words from each language. For the English language, we used unigrams that occurred in at least 500 volumes each year. Because volume counts vary across languages, the other languages required different filtering parameters. The raw data from Google went through three layers of processing: (1) filter layer, (2) consolidation layer, and (3) normalization layer. First, we select unigram data within the desired year and count range for both unigram count and volume count. Second, we converted each unigram into lowercase and summed the frequencies while listing all part-of-speech (POS) annotations into one vector. For example, the word `solo' can be a noun, verb, adjective, or adverb. The POS annotation information and the frequency of the unigrams `solo', 'Solo', or 'SOLO' are then summed into one frequency for the unigram `solo'. For each word, we then count the number of occurrences in each year with the restriction that there should be non-zero counts in all years. This means that any word with a raw count of zero at a particular year is not included. After the raw frequencies are done, we then computed the word proportions in each year. Finally, we convert the word proportions into z-scores for each word. In \nameref{S4_Appendix}, we discuss the data processing in detail.

We then categorize some words in each language. Words like ``water", ``sun", ``moon" and ``night" are considered to describe basic concepts. These words are part of a famous set of words called the \textbf{Swadesh words} which can be obtained using the Natural Language Tool Kit (NLTK) module \cite{bird2009natural}. The list was named after Morris Swadesh, who started creating a list of these words to compare different languages historically and culturally \cite{Swadesh1971}. Some words in this list have multiple meanings which do not have one-to-one relationship between all languages. Additionally, we identify the stopwords in each language. \textbf{Stopwords} are a set of frequently used words - often meaningless - such as prepositions and conjunctions (e.g. ``a", ``and", ``but", ``the", ``is", and ``are"). Intuitively, these words can be stable in ranks because the basic concepts that the words describe tend to be common across languages.

\subsection*{Word rank change quantification}

The most important aspect of our analysis is the rank of words. Rank is a discrete integer measure that determines how frequently a word was used relative to all other words. Compared to the raw frequency, ranks have the advantage of localizing the relative position of the word frequencies in time. Below, we provide details on three metrics we used in this study, which are the sum of rank change, rank change variance, and the Rank Biased Overlap (RBO). The RBO was previously applied to measure the similarity of ranked search results of different internet search engines \cite{Webber2010}. Previous studies have used the amount of turnover, which is the number of new words entering the top word list, which is also equivalent to the number of words exiting \cite{Ruck2017,Acerbi2014}. While our study is not focused on the turnover metric, we discuss it briefly in the discussion section in the context of our model and languages, and how it relates to the RBO.

The Table~\ref{tb:rank-example} below shows the assigned unique rank for example words $w_1$, $w_2$, $w_3$, and $w_4$. The word $w_1$ remained in the 1st rank for all five timestamps while the word $w_3$ changed rank from 4th in $t_1$ to 2nd in $t_2$. The rank for each word is assigned using their proportion, and each word has a unique rank. If two words have the same proportion, the words are alphabetized (for actual words) or sorted (for numerals), and then the ranks are assigned in sequence accordingly. This procedure is rare and is unlikely to introduce bias.

\begin{table}[!htbp]
	\centering
	\caption[Word rank matrix example.]{\textbf{Word rank matrix example with four words and five time stamps.}}
	\label{tb:rank-example}
	\begin{tabular}{|c|c|c|c|c|c|}
		\hline
		& $t_0$ & $t_1$ & $t_2$ & $t_3$ & $t_4$ \\ \hline
		$w_1$ & \textbf{1}                          & \textbf{1}                          & \textbf{1}                          & \textbf{1}                          & \textbf{1}                          \\ \hline
		$w_2$ & 2                          & 2                          & 3                          & 2                          & 2                          \\ \hline
		$w_3$ & 3                          & \textbf{4}                          & \textbf{2}                          & 3                          & 3                          \\ \hline
		$w_4$ & 4                          & 3                          & 4                          & 4                          & 4                          \\ \hline
	\end{tabular}
\end{table}

We also considered a different style of quantifying the ranks. Table~\ref{tb:ranked-list-example} below shows a ranked list for each timestamp. Instead of assigning an integer for the ranks, the words are listed according to their rank. It gives the same information from Table~\ref{tb:rank-example} but with the words instead of integers. The rank matrix shows the time series for ranks in each word, while the ranked list shows the overall structure of word ranks in time.

\begin{table}[!htbp]
	\centering
	\caption[Word ranked lists example.]{\textbf{Word ranked lists example with four words and five time stamps.}}
	\label{tb:ranked-list-example}
	\begin{tabular}{|l|c|c|c|c|c|}
		\hline
		& $t_0$ & $t_1$ & $t_2$ & $t_3$ & $t_4$ \\ \hline
		1 & $\mathbf{w_1}$                      & $\mathbf{w_1}$                      & $\mathbf{w_1}$                      & $\mathbf{w_1}$                      & $\mathbf{w_1}$                      \\ \hline
		2 & $w_2$                      & $w_2$                      & $\mathbf{w_3}$                      & $w_2$                      & $w_2$                      \\ \hline
		3 & $\mathbf{w_3}$                      & $w_4$                      & $w_2$                      & $\mathbf{w_3}$                      & $\mathbf{w_3}$                      \\ \hline
		4 & $w_4$                      & $\mathbf{w_3}$                      & $w_4$                      & $w_4$                      & $w_4$                      \\ \hline
	\end{tabular}
\end{table}

Formally, we denote this rank information as $\mathbf{K}$ with dimensions $c \times T$ for the word ranks and $\mathbf{RL}$ with dimensions $c \times T$ for the ranked list. There are two metrics that describe the overall structure of the rank changes for each word in $\mathbf{K}$. First, we compute the $\mathbf{\Delta K}$ by taking $\mathbf{K_{t}} - \mathbf{K_{t-1}}$ for each $w$. The dimensions of $\mathbf{\Delta K}$ is now $c \times (T-1)$. The first metric is the sum of rank change of word $w$ which is computed by
\begin{equation}
	\sum_{t=0}^{T-2} \Delta k_{w,t}
	\label{eq:sum-of-rank-change}
\end{equation}
where $\Delta k_{w,t}$ is an entry in the matrix $\mathbf{\Delta K}$ for word $w$ at time $t$.
The second metric is the rank change variance of word $w$ which is computed by
\begin{equation}
	\frac{1}{T-1} \sum_{t=0}^{T-2} \left( \Delta k_{w,t} - \overline{\Delta k_{w,t}} \right)^2
	\label{eq:rank-change-variance}
\end{equation}
where $\overline{\Delta k_{w,t}}$ is the mean computed as
\begin{equation}
	\overline{\Delta k_{w,t}} = \frac{1}{T-1} \sum_{t=0}^{T-2} \Delta k_{w,t}.
	\label{eq:rank-change-mean}
\end{equation}
The sum of rank change metric is the net change of the ranks within the time frame which is 109 years. It can capture words that changed in ranks - either monotonically up or down trends - within the time frame. This type of measure ignores cases where a word change ranks often or the ranks go up and down with some variance. For example, if a word initially has rank 10 and changed its rank to 100 in 50 years and went back to rank 10 in an additional 50 years, the net change would be zero. The rank change variance is a second metric to verify that most stopwords and Swadesh words are consistently stable across languages.

Finally, we use the Rank Biased Overlap - or RBO - to measure the overall temporal structure of the ranked list matrix $\mathbf{RL}$. The RBO is a similarity measure to compare two indefinite ranked lists. For a given ranked list matrix $\mathbf{RL}$, we compare $\mathbf{RL}_{t}$ and $\mathbf{RL}_{t+1}$ to each other, $\mathbf{RL}_{t}$ and $\mathbf{RL}_{t+10}$ to each other, and $\mathbf{RL}_{0}$ and $\mathbf{RL}_{t}$ to each other. That means that the RBO is measured for the matrix $\mathbf{RL}$ to see the overall collective changes in the ranked list in time. The RBO measure is in the range $[0,1]$. $RBO=1$ means that both lists are the same, while $RBO=0$ means that both lists are completely different. Following from the work of Webber et al. \cite{Webber2010}, below is the general computation of the RBO similarity measure of two ranked lists.

Given two sets of infinite rankings $S$ and $T$, the RBO is computed generally as
\begin{equation}
	RBO(S,T,p) = (1-p)\sum_{d=1}^{\infty} p^{d-1} \dot A_d
	\label{eq:rbo}
\end{equation}
where the parameter $p$ is the steepness of the weights. This parameter falls in the range $[0,1]$. Smaller $p$ means that less items are considered in both of the list while larger $p$ means that more items are considered in both of the lists. If $p=0$, it means that only the top ranked item for both lists are considered which means that the RBO can only be either 0 or 1. If $p=1$, all items are considered and that RBO fall between the range $[0,1]$. The term $A_d$ is called the \textit{agreement} measure at some depth $d$. The \textit{agreement} is computed by
\begin{equation}
	A_{S,T,d} = \frac{| S_{:d} \cap T_{:d} |}{d}
	\label{eq:rbo-agreement}
\end{equation}
where the operation $| S_{:d} \cap T_{:d} |$ is called the cardinality of the intersection at depth $d$ of the sets $S$ and $T$. In other words, that term is the number of times the two sets have common items up to the depth of $d$. In our analysis, we set $p=1$. See \nameref{S2_Appendix} for the implementation of the RBO.

\section*{Results}

The results are divided into three parts. First, we establish and investigate the relationship between our model simulations and its theoretical limit as a multinomial/binomial distribution. Then we examine the similarities and differences between our WF simulations and the Google unigram data for eight languages.

\subsection*{The relationship between WF model and multinomial sampling}

\textbf{Results summary.} Recall that the initial frequency distribution of the WF model is sampled from the Zipf distribution. The probabilities of sampling a word are inversely proportional to its ranks. There are two key results in this section.
\begin{itemize}
	\item If a word is assigned a rank, then the probability of sampling that word is $P(Y;a,c)$ which is defined in Eq.~\ref{eq:zipf-pmf}. For multiple independent trials, each word follows the binomial distribution. However, these binomials are overlapping, and the chances of having a word having an error from its predefined rank is non-zero.
	\item The binomial distributions suggest that the overlaps are the main cause of a rank error when sampling at the initial time. The results show that if the corpus size increases, the binomial overlaps decrease, while the binomial overlaps increase if the vocabulary size gets closer to the corpus size. 
\end{itemize}
Below, we explain the details of our findings.

The individual components of the multinomial probability distributions are binomial distributions. As shown in Eq.~\ref{eq:binomial-components}, the frequency of a word $w_i$ can be sampled using the binomial distribution with probability mass function shown in Eq.~\ref{eq:binomial-pmf} (or Eq.~\ref{eq:binomial-pmf-initial} at $t=0$). There are three parameters we can vary. The first parameter is the Zipf shape parameter called $a$. This parameter controls the distance between the success probabilities of the most frequent words versus the rest. These probabilities are inserted into the individual Binomials in Eq.~\ref{eq:binomial-pmf-initial} for each word in the vocabulary. If $a=0$, then the Zipf distribution reduces to the uniform distribution. As we increase $a$, the distance between the binomials gets stronger especially for the most frequent word which is the highest ranked word. See \nameref{S1_Fig} for the visualization of the binomials while the Zipf shape parameter $a$ is varied. By increasing the vocabulary size, we observed that lower ranked words have overlapping binomials. We also observed that by increasing the corpus size while the vocabulary size is fixed, the binomial overlaps get weaker. We demonstrate this in detail through visualization in \nameref{S2_Fig}. We tracked these binomial overlaps for each word. We observed that higher ranked word tend to overlap with lower ranked words and low ranked words tend to overlap with higher ranked words. We demonstrate this through visualization in \nameref{S3_Fig} and comparing it to simulations of the WF model in \nameref{S4_Fig}. The main point is that - in our model - the binomial overlaps would predict that low ranked words have the potential to go up ranks, and high ranked words have the potential to go down in ranks.

\begin{figure}[!htbp]
	\centering
	\caption[The net potential rank change.]{\textbf{The average potential rank change.} The Subfigures below show the net potential rank change (Eq.~\ref{eq:overlap}) based on a few examples of the corpus size $\beta$ and the vocabulary size $c$. The Subfigure indicates that as the corpus size increases, the net potential decreases while an increase in vocabulary size increases the net potential. A positive $\widehat{ol}$ means a word has the potential to go down in rank, while a negative $\widehat{ol}$ means a word has the potential to go up in rank. The normalization of the $\widehat{ol}$ values is by dividing the values by the vocabulary size.}
	\label{fig:word-by-overlap}
	\includegraphics[width=\textwidth]{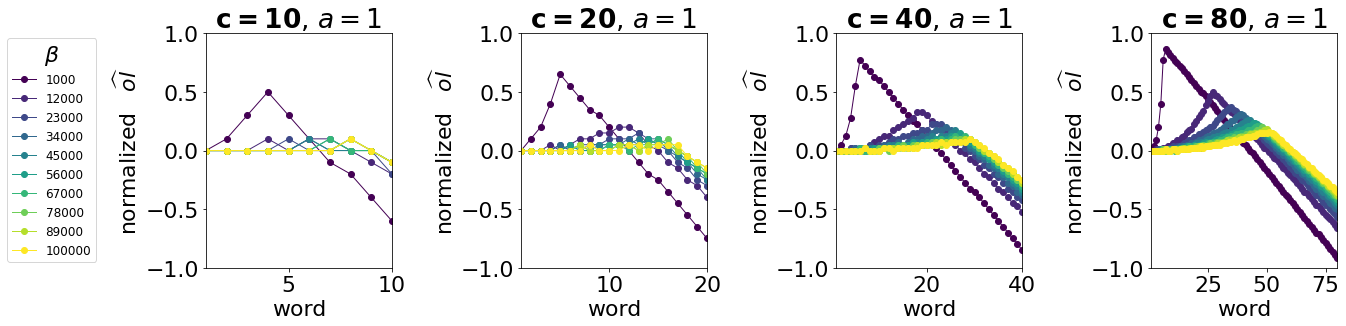}
\end{figure}

The net potential rank change - as explained in Eq.~\ref{eq:overlap} - is the potential for a word to change ranks based on its current rank. We look at these net potential values more broadly in terms of the initial corpus size $\beta$ and the vocabulary size $c$ in Fig.~\ref{fig:word-by-overlap}. For example, in the Subfig labeled $c=20$, the potential at higher ranks is positive, meaning that these words can go down in ranks. The words with negative potential mean that these words have the potential to go up in ranks. We can see in the Subfigs that as the corpus size increases, the net potential for each rank decreases. The potential of words to change ranks decreases as the corpus size increases. By comparison, we can see in the Subfigs that as the vocabulary size increases, the net potential for each ranks increases. Again, more words mean more competition, and the potential of words to change ranks increases. We observe global maximum and minimum values for each case which is the limit of word rank change potential.

\subsection*{Word rank evolution under the WF model}

\textbf{Results summary.} The Wright-Fisher inspired model is a statistical model that relies on sampling words. As explained in the previous section and the Methods section, the multinomial probability transitions of the Markov chain of transitioning from $t-1$ to $t$ would result in the accumulation of sampling errors. These sampling errors are the fundamental reason for evolutionary drift. As explained in the Methods section, a smaller ratio $c/\beta$ means that the initial corpus size is significantly larger than the vocabulary size, which means that the words are less likely to have rank errors in sampling. If the vocabulary size is closer to the initial corpus size, the ratio $c/\beta$ goes to 1, which means that the words are forced to have a frequency of just 1. There are two key results in this section.
\begin{itemize}
	\item We observed that the ratio between $c/\beta$ is an important indicator of the behavior and structure of a single WF inspired model simulation. A significantly small ratio of around $10^{-5}$ in magnitude results in the words to change ranks less regardless of their initial ranks. 
	\item In contrast, a larger ratio around $10^{-2}$ in magnitude results in words changing ranks more often, especially for lower-ranked words. This is due to the binomials overlapping for lower-ranked words for vocabulary sizes greater than two.
\end{itemize}
Below, we show the details of our findings.

\begin{figure}[!htbp]
	\centering
	\caption[100 WF simulations with parameters $\alpha=0.01$, $\beta=1.00\times10^{5}$, $c=1000$, and $a=1$.]{\textbf{100 WF simulations with parameters $\alpha=0.01$, $\beta=1.00\times10^{5}$, $c=1000$, and $a=1$.} The Subfigures below are the results of 100 WF simulations of the given parameter set. Subfig. \textbf{(a)} is the time-series visualization of the raw word counts, proportions, standardized scores, and ranks. These figures show the time-series simulation outcomes of the five example words within the vocabulary. The words shown are the 1st, 10th, 100th, 500th, and 1000th initially ranked words. The results show that the highest initially ranked word has no outcomes where it changed ranks. Subfig. \textbf{(b)} shows the box plot rank distributions of the selected words at the initial time. Subfig. \textbf{(e)} and \textbf{(f)} are the rank change distributions of the selected words. The $\widehat{ol}$ line is the rank change potential from Eq.~\ref{eq:overlap}. As expected, the rank change distributions of the lower-ranked words have higher variances than the high-ranked words. The low ranked words have little or no rank changes and have low variance distributions as expected. Subfig. \textbf{(c)}, \textbf{(d)}, and \textbf{(g)} shows the RBO trends of the WF simulations. It shows that the overall ordered ranks have a consistent pattern for all 100 WF simulations. This particular example have $c/\beta=1.00\times10^{-2}$. Normalize rank means that the ranks are divided by the vocabulary size, which is the variable $c$. For the normalized variance, the variances are divided by the maximum variance.}
	\label{fig:wf_c1K_a1_alpha0.01_beta100K_T109}
	\includegraphics[width=\textwidth]{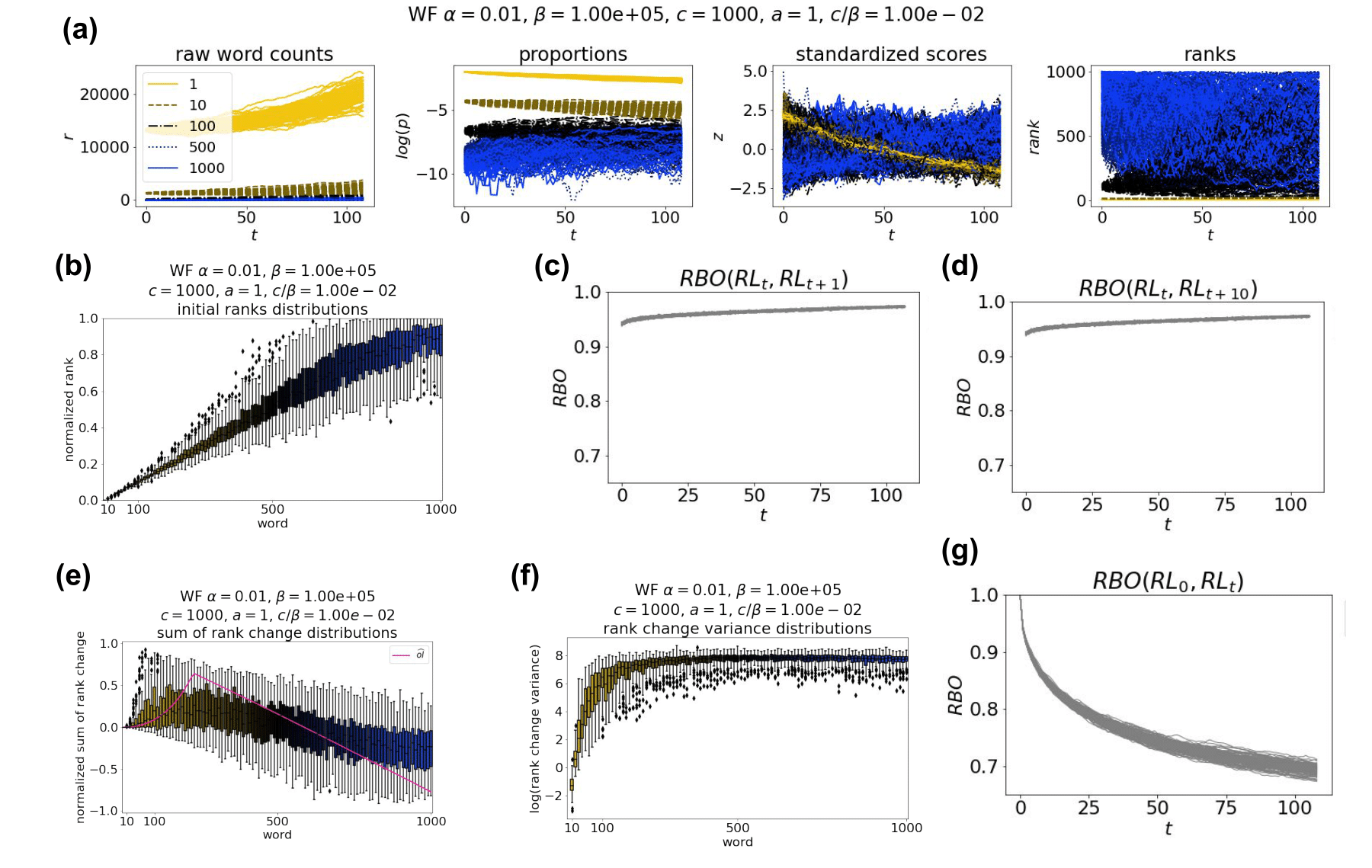}
\end{figure}

First, we look at a 100 WF simulation using parameters $\alpha=0.01$, $\beta=1.00\times10^{5}$, $c=1000$, and $a=1$. For all of the WF simulations shown in this section, the total time is $T=109$. The Subfigs in Fig.~\ref{fig:wf_c1K_a1_alpha0.01_beta100K_T109} shows the results of these 100 WF simulations. The time-series behaviors in Subfig \textbf{(a)} show that the high-ranked words behave in a more deterministic way because these words have fewer chances of overlapping with other words for them to change ranks. In contrast, the low-ranked words show chaotic behaviors because these words have more chances of changing ranks in time. After all, as we have shown in the binomial analysis - the probabilities overlap increases for lower-ranked words if vocabulary size is large. At the initial, we can see in Subfig \textbf{(b)} that the rank distributions of the high-ranked words are very narrow while the low-ranked words have wider rank distributions. We can also verify that the low ranked words change in ranks more often than the high ranked words by looking at Subfig \textbf{(e)} and \textbf{(f)}. For more simulation outcomes with smaller parameter values, see \nameref{S5_Fig} for simulations with varied $\beta$ and \nameref{S6_Fig} for simulations with varied $c$.

The RBO trends of the simulations in Fig.~\ref{fig:wf_c1K_a1_alpha0.01_beta100K_T109} show that the ordered rank list shows consistent changes in time. It means that - even though the words are changing ranks - the overall structure of the word ranks are predictable. For example, Subfig \textbf{(c)} and \textbf{ (d)} shows that the RBO trends for $RL_{t}$ and $RL_{t+1}$ are consistently have the same pattern for all 100 WF simulations. The RBOs are greater than $0.90$. It means more than 90\% of the words at $RBO_{t}$ retained their rank or have little change in rank at $RBO_{t+1}$.
In comparison, the RBO for $RL_{0}$ and $RL_{t}$ in Subfig \textbf{(g)} shows decreasing RBO trends which means that the ordered ranks changed significantly since the initial time point. This means that small accumulated rank changes by year result in big changes in ranks for a longer time length. This is due to the accumulated sampling errors of the binomials. We can see that at $t=100$, only around 70\% of words retained their rank or had little change in rank since the initial. Even though we see over 90\% of words retained their rank at the current time from the previous time, the overall structure of the ordered ranks can change significantly through time. This is due to the behavior of the low ranks words to change in ranks more often than high-ranked words, but the corpus size is large enough for more words to have fewer chances of changing their ranks significantly.

\begin{figure}[!htbp]
	\centering
	\caption[One WF simulations with parameters $\alpha=0.01$, $\beta=1.00\times10^{5}$, $c=1000$, and $a=1$.]{\textbf{One WF simulations with parameters $\alpha=0.01$, $\beta=1.00\times10^{5}$, $c=1000$, and $a=1$.} The Subfigures below are the results of one WF simulation of the given parameter set. Subfig \textbf{(a)} is the time-series visualization of the raw word counts, proportions, standardized scores, and ranks. These figures show one time-series outcome of the five example words within the vocabulary. Subfig \textbf{(b)} is the initial distribution of word ranks with annotated words from Subfig \textbf{(a)}. The label of the word corresponds to its initial rank. Subfig \textbf{(c)} is the normalized distribution of the sum of rank changes of all 1000 words, while Subfig \textbf{(d)} is the normalized distribution of the rank change variance. The mode of this distribution is close to zero, but the second mode is around 0.50. Subfig \textbf{(e)} is a scatter plot of the initial ranks versus the sum of rank change of the words. The $\widehat{ol}$ line is the rank change potential from Eq.~\ref{eq:overlap}. It shows that high initially ranked words tend to go down in ranks while low initially ranked words tend to go up. Similarly, in Subfig \textbf{(f)}, high initially ranked words have lower variances while low initially ranked words have higher variances. Subfig \textbf{(c)}, \textbf{(d)}, and \textbf{(g)} shows the RBO trends of the one WF simulation. This particular example have $c/\beta=1.00\times10^{-2}$. The normalization of the sums is by dividing the values by the vocabulary size, while the normalization of the variances is by dividing the values by the maximum variance.}
	\label{fig:wf-one_c1K_a1_alpha0.01_beta100K_T109}
	\includegraphics[width=\textwidth]{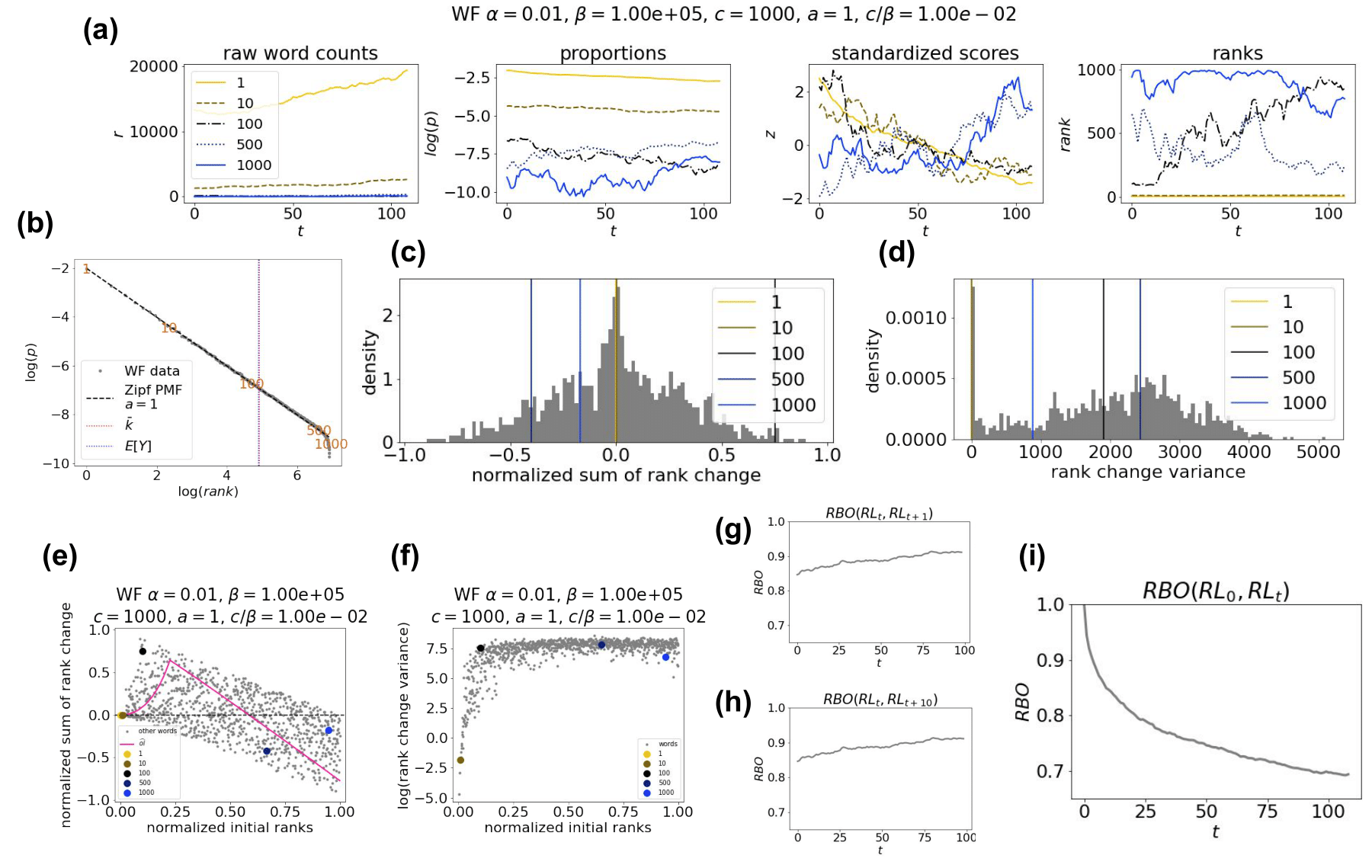}
\end{figure}

The results in Fig.~\ref{fig:wf_c1K_a1_alpha0.01_beta100K_T109} are the 100 different outcomes of the WF inspired model with parameters $\alpha=0.01$, $\beta=1.00\times10^{5}$, $c=1000$, and $a=1$. t. Next, we look at just one outcome of this particular WF model with the given parameters. As you can see in Fig.~\ref{fig:wf-one_c1K_a1_alpha0.01_beta100K_T109} Subfig \textbf{(a)}, the rank time-series for the 1st and 10th initially ranked words have no changes in ranks, and we know that for high initially ranked words, there is high certainty that these words have no other different outcomes. The 100th, 500th, and 1000th initially ranked words are observed to have rank changes, and we know that for low initially ranked words, there are other possible outcomes than the one observed. In this particular case, we look at the structure of the multiple time series as a whole. On Subfig \textbf{(b)}, this is the initial rank distribution of the words. Results show that at the lower-ranked words, there is an error in the samples as expected. On Subfig \textbf{(c)}, this is the normalized distribution of the sum of rank change. The left tail represents the words that went up in ranks, and the right tail represents words that went down in ranks. The normalization process divides the values by the vocabulary size to make the sums in the range $[-1,1]$. The distribution indicates that most of the words have little or no rank changes. This includes the 1st initially ranked word or the highest-ranked word, which is located at zero. In contrast, the highest-ranked word is also located at zero when looking at the variance distribution shown in Subfig \textbf{(d)}. There are also a lot of words with zero variance in their rank changes. This distribution is also normalized by dividing the values by the maximum variance to make the range $[0,1]$. The variance distribution - unlike the sum distribution - is skewed and has two modes in this case. The initial rank distribution is important on how the rank changes in time. We see in Subfig \textbf{(e)} that words closer to high ranked words tend to go down in ranks while words closer to the low ranked words tend to go up in ranks. There is still uncertainty on how the rank changes based on the rank distribution, but it will certainly remain in rank given a large enough initial corpus size for the highest-ranked word. Similarly, for the lowest-ranked word, the only way it can change is to go up in rank. On Subfig \textbf{(f)}, we see the initial rank versus the rank change variance. This figure tells us the volatility of how to word change in ranks. For example, the 10th initially ranked word is observed to changed down in ranks and have low variance. This means that the word is almost consistent in changing its rank upward. In comparison, the 500th initially ranked word is observed to change rank inconsistently were at around $t=50$ to $t=60$, the word radically changed its direction. We see in Subfig \textbf{(f)} that the variances concerning the initial ranks are high and uncertain for lower-ranked words.

Next, we look at an extreme case of the model where we significantly increase the initial corpus size. In Fig.~\ref{fig:wf-one-sim_c1000_a1_alpha0.01_beta100000000_T109}, we observe a radical change in the behavior of the WF inspired model. As mentioned in the binomial analysis from the previous section, a larger corpus size would result in a less likely change in ranks because their binomial distributions would overlap less. We can see the effects of an extreme initial corpus size in Subfig \textbf{(a)} where the low ranked words are stabilizing, unlike the behaviors in the less extreme case. The high-ranked words also remained stable in ranks. In Subfig \textbf{(e)}, the sum of rank changes of words are mostly low and zero sums regardless of their initial ranks. This means that almost all of the words have not changed in ranks significantly. The rank change variance is also low for high-ranked words shown in Subfig \textbf{(f)}. The initial ranks versus the rank change variance shown in Subfig \textbf{(f)} exhibit a predictable pattern where high-ranked words have less variance while low ranked words have more variance. In general, each word's variances here are low, but the normalization process of the values made the curve obvious. The curve also explains that low-ranked words still have some variance even with extreme initial corpus size. There is a special case happening for the lowest-ranked words where we observed that the variance starts to go down. It would be reasonable to predict that for an infinite amount of initial corpus size, the sum of rank change variances would go to zero, and the variances would also go to zero. For Subfigs \textbf{(g)}, \textbf{(h)}, and \textbf{(i)}, the RBO curves for this extreme case of WF inspired model shifted up because the initial corpus size is large enough for the words to have less rank change.

\begin{figure}[!htbp]
	\centering
	\caption[An ``extreme" case of One WF simulations with parameters $\alpha=0.01$, $\beta=1.00\times10^{8}$, $c=1000$, and $a=1$.]{\textbf{An ``extreme" case of One WF simulations with parameters $\alpha=0.01$, $\beta=1.00\times10^{8}$, $c=1000$, and $a=1$.} The Subfigures below are the results of one WF simulation of the given parameter set. This is an extreme case with significantly higher $\beta$ of the one shown in Fig.~\ref{fig:wf-one_c1K_a1_alpha0.01_beta100K_T109}. Subfig \textbf{(a)} is the time-series visualization of the raw word counts, proportions, standardized scores, and ranks. These figures show one time-series outcome of the five example words within the vocabulary. Subfig \textbf{(b)} is the initial distribution of word ranks with annotated words from Subfig \textbf{(a)}. The label of the word corresponds to its initial rank. Subfig \textbf{(c)} is the distribution of the sum of rank changes of all 1000 words, while Subfig \textbf{(d)} is the distribution of the rank change variance. The mode of this distribution is still close to zero, but the entire distribution contracted towards zero. Subfig \textbf{(e)} is a scatter plot of the initial ranks versus the sum of rank change of the words. The $\widehat{ol}$ line is the rank change potential from Eq.~\ref{eq:overlap}. It shows that, given that $\beta$ is extremely high while other parameters are fixed, there a few rank changes regardless of initial ranks. Similarly, in Subfig \textbf{(f)}, high initially ranked words have significantly lower variances relative to low initially ranked words. We can see clearly where the words follow a curve, and the words are not as scattered as from the previous case, but in general, the words, in this case, are very low. Subfig \textbf{(c)}, \textbf{(d)}, and \textbf{(g)} shows the RBO trends of the one WF simulation. This particular example have $c/\beta=1.00\times10^{-5}$.}
	\label{fig:wf-one_cK_a1_alpha0.01_beta100KK_T109}
	\includegraphics[width=\textwidth]{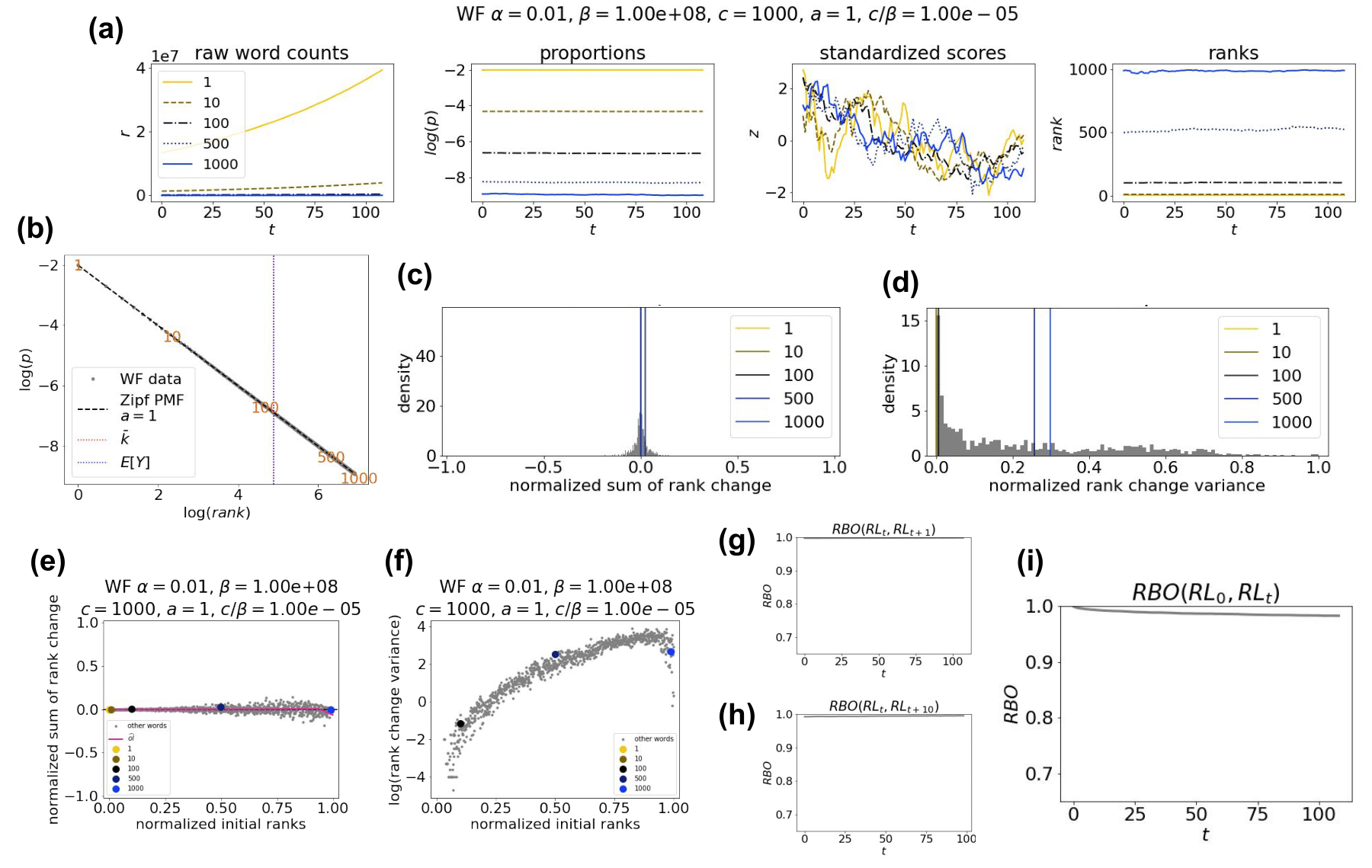}
\end{figure}

We further investigate the behaviors of the RBO curves by changing the parameters of the WF inspired model. The RBO measure is the similarity between two ranked lists, which is defined in Eq.~\ref{eq:rbo}. Here, we explain six different cases of varied parameter values while other parameters are fixed., we show six different cases of varied parameter values while other parameters are fixed. Case 1 is when the corpus size rate is varied. This is when the corpus size of the WF inspired model increases exponentially in time. That means that - while the vocabulary size stays fixed - the words' behavior in time gets less likely to change ranks. We see in the RBO trends that the curve slightly shifts upward as $\alpha$ increases, which means that the words are stabilizing their rank in time. Similarly, in Case 2, while the initial corpus size is varied and increasing, the entire RBO curves shifted more strongly upward because the corpus size is large enough to begin with for words to stabilize rank more quickly. In contrast, Case 3 is when the vocabulary size $c$ is varied and increasing. We see in this case that the RBO curves shift downward as $c$ increases. Note that the vocabulary size parameter $c$ is prescribed at the initial time and is unchanging in time. This means that - while the initial corpus size stays fixed - the increase in vocabulary size at the initial time would let words strongly compete with each other, and they would have more than likely to change ranks in time. Cases 4 and 5 is when the ratio $c/\beta$ is fixed. Results show that the RBO curves does not shift for these cases. Even with the vocabulary size and initial corpus size is increasing while the ratio is fixed, the behavior of the words collectively are consistent. Similar to Cases 4 and 5, Case 6 - where the Zipf shape parameter is varied and increasing - the RBO curves stayed the same, meaning that no matter what the initial state distribution is defined, the RBO curves will be consistent. See \nameref{S7_Fig} for more details and visualizations.

For a fixed ratio of $c/\beta$ while $c$ and $\beta$ are changing, the overall structure and behavior of the word rank change give consistent RBO curves for the WF inspired model. This means that the ratio of $c/\beta$ indicates how an entire language space is behaving compared to the WF model. For instance, we expect similar word rank change behaviors for a real language with $c/\beta=1.00\times10^{-5}$ compared to the ``extreme" case we showed in Fig.~\ref{fig:wf-one_cK_a1_alpha0.01_beta100KK_T109} with ratio $c/\beta=1.00\times10^{-5}$. The languages we observed do have a small ratio $c/\beta$ down to $10^{-6}$ in magnitude. Still, we see different rank change behaviors in the natural languages compared to the ``extreme" case of the WF inspired model. In the next section, we explain the details of our results of the languages, comparing it to the model.

\subsection*{Word rank evolution in Google unigram data for eight languages}

\textbf{Results summary.} The Google unigram data contains the time-series of unigrams from 1900 to 2008 in eight languages. The WF inspired model simulates the evolutionary drift process in which the frequency of the unigrams changes based entirely on frequency-dependent re-sampling. We compared the languages from the Google unigram data, and the ``extreme" case of the WF inspired model and found these three key results.
\begin{itemize}
	\item Most of the words in each of the eight languages have relatively little or no rank changes compared to other words, suggesting that the languages we considered are mostly stable.
	\item Some words change in ranks significantly from 1900 to 2008 in two ways: (a) by accumulating small increasing/decreasing rank changes in time and (b) by shocks of increase/decrease in ranks. Most of the stop words and Swadesh words appear to be stable in ranks for each of the eight languages.
	\item The word ranks in the languages as a whole change more significantly than the ``extreme" case of the WF inspired model despite the ratio $c/\beta$ being as low as the languages. This suggests that words in each of the languages collectively behave contrary to a neutral evolutionary process.
\end{itemize}
The above three summaries are the most important results in this section. Below we explain the details of the results and show that the languages deviate from neutral evolution.

First, we look at an overview of the Google Ngram data in Table.~\ref{tab:la-summary-results}. After processing the Google unigram data, the resulting vocabulary for each language is in the $c_{data}$ column of the Table. The resulting initial corpus size is in the $\beta_{data}$ column of the Table. The vocabulary sizes of the languages varied a lot, where the highest is 18737 for the English language and 180 for the Simplified Chinese language. The overwhelmingly sizeable initial corpus size results in the ratio $c/\beta$ to be roughly $10^{-6}$. The vocabulary sizes are that way since the processing includes where we remove the words with zero counts in any year. The data we present here is a representative sample of the words used every year from 1900-2008 for each language. It is essential to mention three other variants in the English language here, namely American English, British English, and English Fiction. The English language is the combined information of all variants. They have overlapping vocabulary words. There are also Spanish and French words in the English set because English speakers often borrow words from these languages historically and culturally.

\begin{figure}[!htbp]
	\centering
	\caption[Time-series visualization of six example words of the English data and the distributions of sum of rank change and rank change variance of all English words in the data.]{\textbf{Time-series visualization of six example words of the English data and the distributions of sum of rank change and rank change variance of all English words in the data.} Subfig \textbf{(a)} is the time-series of six example words in the English vocabulary. This figure shows the words ``a", ``the", ``gay", ``farm", ``cession", and ``jobs". We see that the most frequent words ``a" and ``the" did not change ranks in time. Subfig \textbf{(b)} is the initial rank distribution with the fitted Zipf shape parameter $a$. Subfig \textbf{(c)} is the normalized distribution of the sum of rank changes showing words that went up in ranks (list \textbf{A}), little or no rank change (list \textbf{B}), and words that went down in ranks (list \textbf{C}). The sums are normalized by dividing the values by the vocabulary size. We see that most words in the list (list \textbf{A}) are stop words and Swadesh words. Subfig \textbf{(d)} is the normalized distribution of the rank change variances showing words with little or no variance (list \textbf{A}), words with average variances (list \textbf{A}), and words with extreme variances (list \textbf{C}). We also see that all of the words in (list \textbf{A}) are stop words and Swadesh words. The variances are normalized by dividing the values by the maximum variance.}
	\label{fig:eng_time-series_distributions}
	\includegraphics[width=\textwidth]{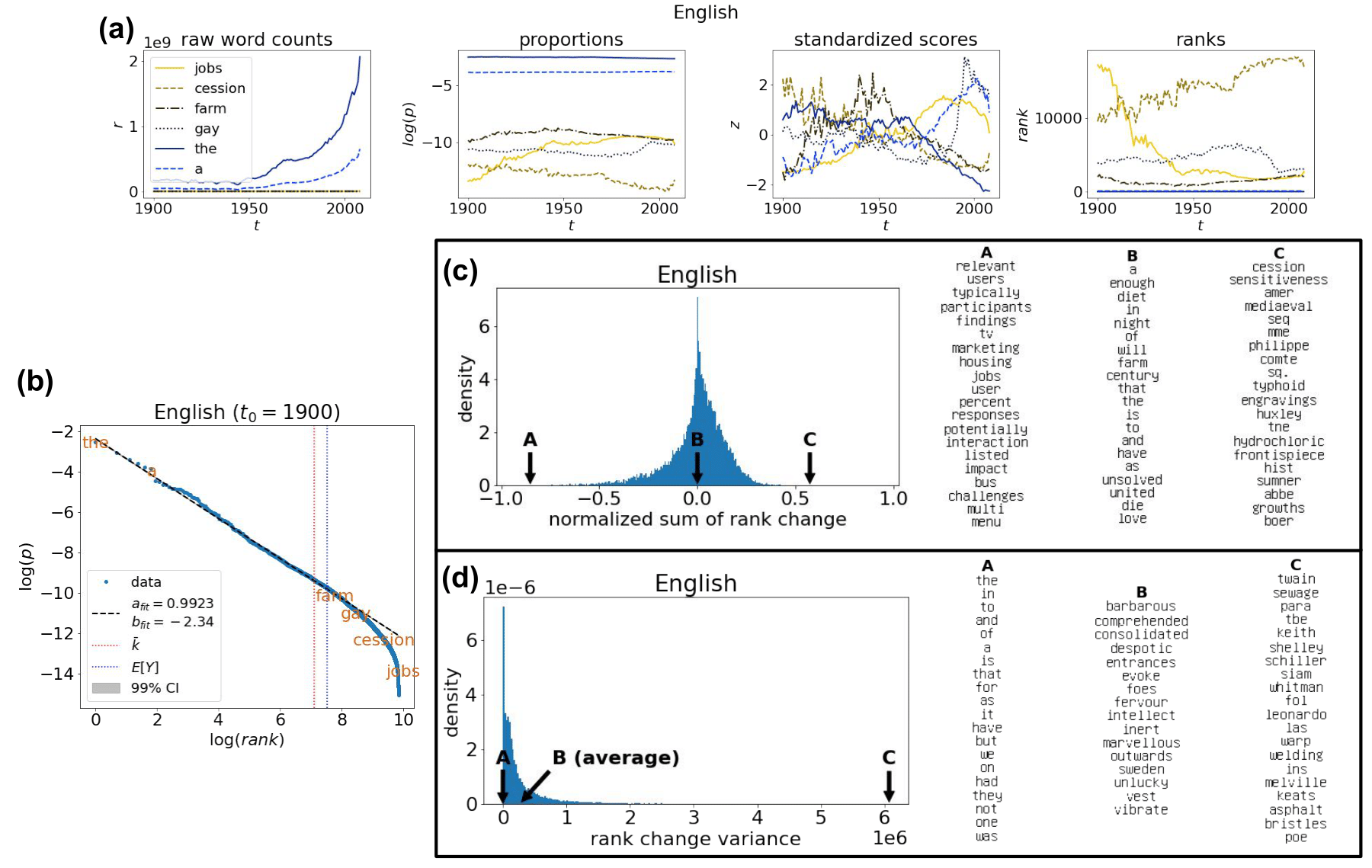}
\end{figure}

In Table.~\ref{tab:la-fitting-results}, we look at the fits of the corpus function in Eq.~\ref{eq:corpus-size-function} and the Zipf probability mass function in Eq.~\ref{eq:zipf-pmf}. Recall that the corpus size function governs the corpus size (or total unigram frequency of words) at time $t$ with parameters $\beta$ which is the initial corpus size, and $\alpha$, which is the corpus size rate increase. These parameters are estimated such that these numbers fit into the corpus size time-series data of each language. Using the log-transformed corpus size time-series and the log-transformed corpus size function, we fit these parameters using the non-linear least-squares method (See \nameref{S3_Appendix} for more details). Results show that the $\alpha$ values are roughly close to each other between the languages. The language with the highest rate of corpus size change is Simplified Chinese which is much higher than other languages. See \nameref{S8_Fig} for the visualization of the corpus size function fitted against each language data corpus size time-series. The initial Zipf distribution shape parameter $a$ for each language varies but roughly close to 1 (See \nameref{S3_Appendix} for more details), and the data is truncated such that the highest-ranked word up to the word closest to the sample mean word rank of the data (see Eq.~\ref{eq:zipf-data-expected-value}) is fitted while the rest is ignored. Since we know that lower-ranked words are prone to sampling errors, this method of semi-data truncation will fully minimize the error of the fit to achieve optimal fit for the most frequently used words. Fig.~\ref{fig:eng_time-series_distributions} Subfig \textbf{(b)} shows the fitted log-transformed Zipf function fitted against the English initial log-transformed rank distribution. In this figure, the lower ranks presented a significant error from the fitted line, but these data points are ignored in the fit. The higher-ranked words are almost perfectly on the fitted line. The resulting fitted Zipf shape parameter for the English initial rank distribution is $0.9923 \approx 1$. See \nameref{S9_Fig} for the visualization of the log-transformed Zipf function fitted against each language data. The Zipf probability mass function Eq.~\ref{eq:zipf-pmf} is non-linear, but applying the log transform of the function yields a linear function where the shape parameter is the slope of that linear function. For our analysis, we have large enough data to assume that this parameter is close to $1$ as shown in fitting the Zipf distribution on the language data (See Table.~\ref{tab:la-fitting-results}).

\begin{table}[!htbp]
	\centering
	\caption[Table of the Google unigram data vocabulary sizes ($c_{data}$) and initial corpus sizes ($\beta_{data}$).]{\textbf{Table of the Google unigram data vocabulary sizes ($c_{data}$) and initial corpus sizes ($\beta_{data}$).} This Table includes the number of available stop words ($c_{stop}$) and Swadesh words ($c_{swad}$) in each language. The last column is the vocabulary to corpus size ratio $c_{data}/\beta_{data}$.}
	\label{tab:la-summary-results}
	\begin{tabular}{@{}|c||c|c|c|c|c|c|c|c|c|c|c|@{}}
		\hline
		\textbf{Language} &
		$c_{data}$ &
		$c_{stop}$ &
		$c_{swad}$ &
		$\ln{(\beta_{data})}$ &
		$\frac{c_{data}}{\beta_{data}}$ \\[5pt] \hline
		\textbf{English}            & 18737 & 571 & 202 & 21.4598 & $9e^{-6}$ \\ \hline
		\textbf{American English}   & 16410 & 568 & 202 & 21.3127 & $9e^{-6}$ \\ \hline
		\textbf{British English}    & 4759 & 592 & 171 & 20.2960   & $7e^{-6}$ \\ \hline
		\textbf{English Fiction}    & 5651 & 478 & 193 & 19.0564      & $3e^{-5}$ \\ \hline
		\textbf{Simplified Chinese} & 180 & 49 & 30  & 13.7613       & $1.9e^{-4}$ \\ \hline
		\textbf{French}             & 12168 & 116 & 193 & 20.5723  & $1.4e^{-5}$ \\ \hline
		\textbf{German}             & 5871 & 113 & 142  & 19.7012  & $1.6e^{-5}$ \\ \hline
		\textbf{Italian}            & 4446 & 123 & 121 & 18.9443 	  & $2.6e^{-5}$ \\ \hline
		\textbf{Hebrew}             & 3000 & 313 & 144 & 16.8523    & $1.44e^{-4}$ \\ \hline
		\textbf{Russian}            & 828 & 238 & 53  & 18.0485   & $1.2e^{-5}$ \\ \hline
		\textbf{Spanish}            & 10661 & 140 & 174 & 19.1625  & $5.1e^{-5}$ \\ \hline
	\end{tabular}
\end{table}

\begin{sidewaystable}[!htbp]
	\centering
	\caption[Table of the fitted parameter values of the corpus size function and the Zipf probability mass function.]{\textbf{Table of the fitted parameter values of the corpus size function and the Zipf probability mass function.} This Table shows the fitted estimate of the parameter values for the corpus size function from Eq.~\ref{eq:corpus-size-function} and the Zipf shape parameter for the initial distribution in Eq.~\ref{eq:zipf-pmf}. The 99\% confidence intervals are computed using the student-t distribution.}
	\label{tab:la-fitting-results}
	\begin{tabular}{@{}|c||c|c|c|c|@{}}
		\hline
		\textbf{Language} &
		$\ln{(\beta_{fit})}$ {\footnotesize (99\% CI)} &
		$\frac{c_{data}}{\beta_{fit}}$ &
		$\alpha_{fit}$ {\footnotesize (99\% CI)} &
		$a_{fit}$ {\footnotesize (99\% CI)} \\[5pt] \hline
		\textbf{English}                & 20.7866 {\footnotesize (20.5867, 20.953)} & $1.80e^{-5}$ & 0.0239 {\footnotesize (0.021, 0.0267)} & 0.9923 {\footnotesize (0.9879, 0.9967)} \\ \hline
		\textbf{American English}   & 20.6247 {\footnotesize (20.4173, 20.7964)} & $1.80e^{-5}$ & 0.0215 {\footnotesize (0.0185, 0.0245)}	 & 1.0056 {\footnotesize (1.0016, 1.0095)} \\ \hline
		\textbf{British English}       & 19.5205 {\footnotesize (19.2605, 19.7267)}  & $1.6e^{-5}$ & 0.0179 {\footnotesize (0.0142, 0.0215)} & 1.0388 {\footnotesize (1.0316, 1.0461)} \\ \hline
		\textbf{English Fiction}       & 18.0306 {\footnotesize (17.735, 18.2585)}  & $8.30e^{-5}$ & 0.0306 {\footnotesize (0.0265, 0.0347)} & 1.0776 {\footnotesize (1.0662, 1.0889)} \\ \hline
		\textbf{Simplified Chinese}  & 9.4742 {\footnotesize (8.595, 9.9347)}   & $1.38e^{-2}$ & 0.1068 {\footnotesize (0.0975, 0.1162)}	 & 0.8244 {\footnotesize (0.6986, 0.9501)} \\ \hline
		\textbf{French}                 & 19.8126 {\footnotesize (19.5795, 20.0015)} & $3.0e^{-5}$ & 0.0109 {\footnotesize (0.0075, 0.0142)} & 0.9969 {\footnotesize (0.9881, 1.0058)} \\ \hline
		\textbf{German}               & 19.3325 {\footnotesize (19.0549, 19.5496)} & $2.4e^{-5}$ & 0.0120 {\footnotesize (0.0082, 0.0159)} & 1.0317 {\footnotesize (1.0201, 1.0432)}	 \\ \hline
		\textbf{Italian}                 & 18.3050 {\footnotesize (18.1225, 18.4593)} & $5.0e^{-5}$ & 0.0208 {\footnotesize (0.0181, 0.0235)} & 0.9948 {\footnotesize (0.9856, 1.0039)} \\ \hline
		\textbf{Hebrew}               & 15.8291 {\footnotesize (15.5314, 16.0581)}  & $4.01e^{-4}$ & 0.0342 {\footnotesize (0.03, 0.0383)}	 & 0.9771 {\footnotesize (0.9664, 0.9878)}	 \\ \hline
		\textbf{Russian}               & 17.3915 {\footnotesize (17.0154, 17.6642)} & $2.30e^{-5}$ & 0.0319 {\footnotesize (0.0269, 0.0369)}	 & 0.9151 {\footnotesize (0.8937, 0.9365)} \\ \hline
		\textbf{Spanish}               & 18.9342 {\footnotesize (18.8706, 18.994)} & $6.4e^{-5}$ & 0.0265 {\footnotesize (0.0255, 0.0274)} & 0.9378 {\footnotesize (0.9267, 0.9489)} \\ \hline
	\end{tabular}
\end{sidewaystable}

As an example, we show the Google unigram time-series trends of the words ``jobs", ``cession", ``farm", ``gay", ``the", and ``a" in Fig~\ref{fig:eng_time-series_distributions} Subfig \textbf{(a)}. The first thing to notice that the words ``the", ``and" and ``a" are examples of the most frequently used word and their ranks in time remained constant. These types of vocabulary words are called \textbf{stop words} which can be obtained from Ranks NL website, \url{https://www.ranks.nl/stopwords}. This group of words is mostly function words because they have little lexical meaning but serve an important function, to construct a meaningful and grammatical sentence. Other stop word examples are ``to", ``and", and ``they". Second, the word ``gay" shows an increase in frequency around the 1960s and its rank went up a few ranks. The most significant rank change for this example is with the word ``jobs". As you can see in the fourth subplot in Fig~\ref{fig:eng_time-series_distributions} Subfig \textbf{(a)}, the rank trend for ``jobs" went from one of the lowest ranks in 1900 to one of the highest ranks in the year 2000. Similarly in an opposite way, the word ``cession" decreased in rank. In contrast to the rank trend of ``jobs", the rank trend for the word ``farm" generally remained in the higher ranks but there are some changes as well. Significant industrial and technological changes in the past have contributed to many cultures changing and adapt to the modern workforce which explains the word ``jobs" has gained ranks significantly.

In Fig.~\ref{fig:eng_time-series_distributions} Subfigs \textbf{(b)} to \textbf{(c)}, we show more details on other words and their rank change behaviors. Subfig \textbf{(a)} show that for the most frequent words, ``the" and ``a", their ranks remained constant while lower-ranked words have some changes in their ranks. The word ``jobs" initially in the lower rank (Subfig \textbf{(b)}) but became high ranked in time. The sum of these rank changes is a metric for each word on how their rank changes in time. For example, the word ``jobs" has a negative sum of rank change which is higher than most words. Negative sums mean that a word changes up in ranks, while a positive-sum means a word change down in rank. The value of the sum is the magnitude of the change. We can see the normalized distribution of these sums of rank change for the English language in Subfig \textbf{(c)}. The sums are normalized by dividing the values by the vocabulary size. This will transform the sums in the range $[-1,1]$. Shown in Subfig \textbf{(c)}, the set \textbf{A} of words are the words that changed up in ranks in 109 years. Words such as ``jobs", ``job", ``user", ``users", ``marketing", and ``housing" are words with socio-economic meanings. Because of technological advancements in the past century, these words changed up in ranks and became widely used. On the opposite end of the distribution, the set \textbf{C} are words that changed down in ranks in 109 years. These words became less frequent in usage because of changes in culture. For example, the word ``mediaeval" with the emphasis in the extra letter ``a" between ``i" and ``e" became less frequent in usage compared to the word ``medieval". It is probably because of a spelling preference between American versus British where more people prefer writing it without the ``a" for simplicity. Other words such as ``cession" and ``typhoid" also significantly went down in ranks. In addition, words such as ``phillipe",``huxley", ``sumner", ``abbe", and ``boer" are names that became less popular today than 100 years ago. The set \textbf{B} of words is mostly the stop words such as ``a", ``in", ``off", ``that", ``the", ``is", and ``it". These are the most frequently used words and more stable in ranks than other words. On Subfig \textbf{(d)}, this is the normalized distribution of the rank change variance. Normalized means that the values are divided by the maximum variance. We can see this Subfig that set \textbf{A} of words are mostly stop words because they are stable in ranks. The set \textbf{B} are words with an average variance, while the set \textbf{C} are words with the highest variance. For words with the highest variance, they are mostly names and nouns such as ``twain", ``keith", and ``shelly". We can speculate that names are more susceptible to changes based on seasonality. Furthermore, the words listed in these lists for both Subfig \textbf{(c)} and \textbf{(b)} are words in the English set of the Google Ngram data. The American English, British English, English Fiction are special sets of English Language data to separately distinguish different vocabulary usage of American culture, British culture, and the words used in fictional writings. The English set is a combination of all three. The distributions of other English data set and other languages are shown in \nameref{S10_Fig} to \nameref{S17_Fig}.

\begin{figure}[!htbp]
	\centering
	\caption{\textbf{The initial ranks versus the sum of rank change and rank change variance of the English data compared with the WF inspired model.} Looking at the Subfig by row, Subfig \textbf{(a)} shows the initial ranks versus the sum of rank changes of the English data compared with the ``extreme" case of the model. This figure also shows the stop words and Swadesh words on where they are located in the English scatter plot. Subfig \textbf{(b)} shows the initial ranks versus the rank change variance of the English data compared with the extreme case of the model. It shows that these groups of words are the most frequently used words, and most of them have little or no change in ranks with minimal variance. We also see a big difference in the shape of the scatter plots between the English data and the WF inspired model even though they have both almost the same ratio down to $c/\beta = 1.00\times10^{-5}$. The normalization of the sums is by dividing the values by the vocabulary size, while the normalization of the variances is by dividing the values by the maximum variance. Most of the words for both the English and the ``extreme" case of the WF model have very low-rank change variances. We note that due to the max normalization of the variances, the English data appears flat compared to the WF model mainly because English and other languages have outliers, and most of the variances are low.}
	\label{fig:eng_time-series_initial-vs-sum-variance}
	\includegraphics[width=\textwidth]{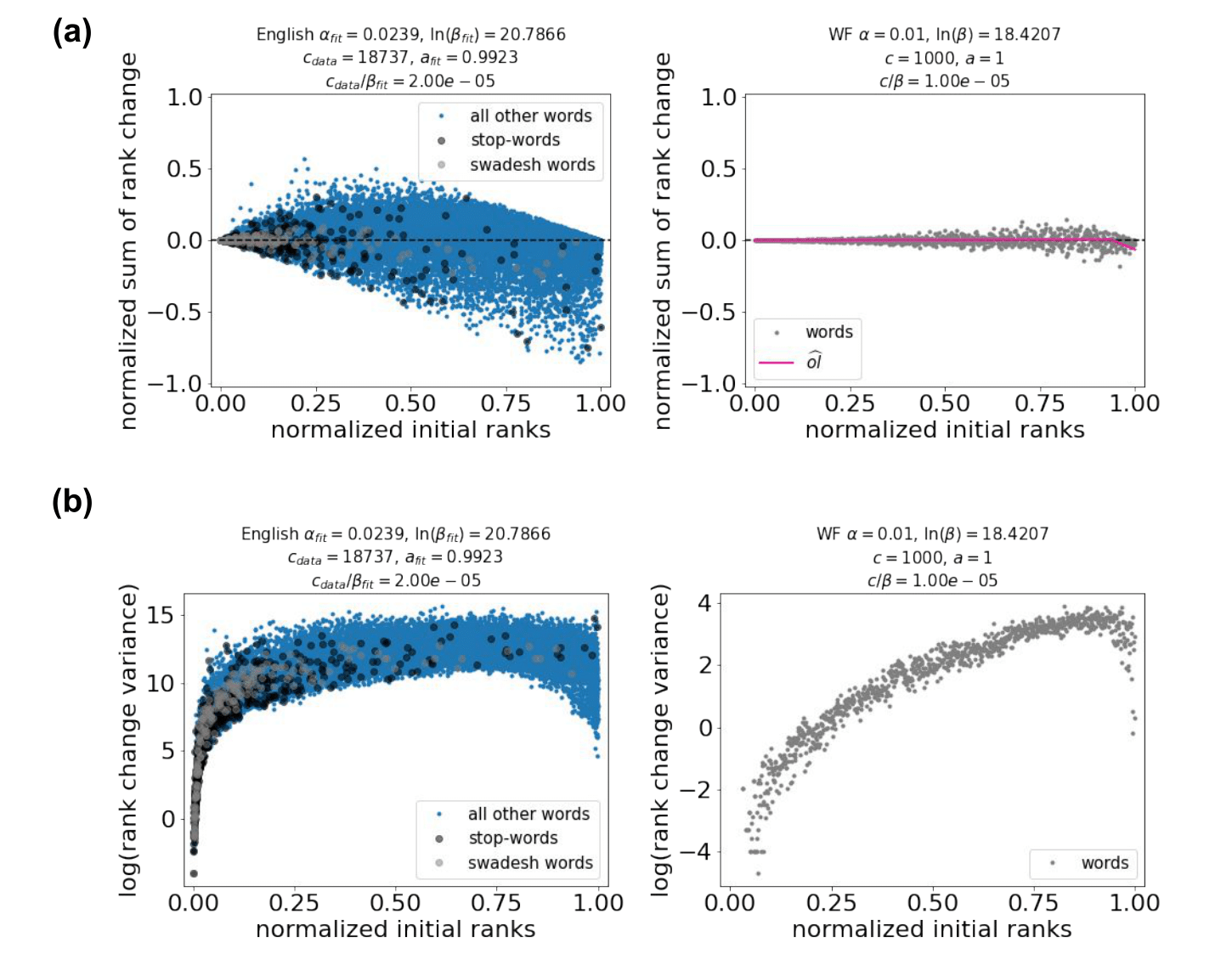}
\end{figure}

Now that we considered the distributions of the sum of the rank change, we consider the initial rank distribution compared to the sums. From our analysis in the WF simulations, the initial distribution is important since it determines the overall behavior on how the rank changes in time. We show the results of the English data set compared with the ``extreme" case of the WF inspired model single simulation in Fig.~\ref{fig:eng_time-series_initial-vs-sum-variance}. Subfig \textbf{(a)} left subplot shows the scatter plot of the initial rank versus the sum of rank change for the English language. The words in the high ranks initially tend to go down in ranks while the words in the low ranks initially tend to go up in ranks. Most of the words have sums closer to zero, which means that the words in the English language tend to be stable. However, some words did change in ranks significantly. We can also see that the stop and Swadesh words are in the high ranks, and most of them have little or zero sums of rank change, which means that these words are stable. The English data set has ratio $c/\beta = 1.00\times10^{-5}$. That means that there is a significantly smaller vocabulary words than corpus size. Compared to the ``extreme" case of the model - shown on the right of subplot \textbf{(a)} - where we set the ratio to be the same $c/\beta = 1.00\times10^{-5}$, the results are different from the English data. \textit{Why?} As we mentioned in the Methods section, the binomial components of the multinomial distributions get separated as the corpus size increases. For smaller corpus sizes, the binomials overlap, and there are higher chances for words to change ranks. The ``extreme" case of the WF model has significantly more corpus size than the vocabulary size. That is why the sum of rank change for each word is close to zero. Compared to the English data, where it also has more corpus size than vocabulary size, the English sum of rank changes has high sums, and most of its words have little sums. Even though the ratio is the same, the difference between the English data and the WF inspired model is striking. The WF inspired model mainly models purely drift processes. The smaller the corpus size, the stronger the drift effect. The higher the corpus size, the drift effect is weaker. The English data shows that some words are behaving contrary to drift.

\begin{figure}[!htbp]
	\centering
	\caption{\textbf{The initial ranks versus the sum of rank change of language data compared with the extreme case of the WF inspired model.} Each of the subplots below shows the initial rank versus the sum of rank change for each language in the Google unigram data with stop words and Swadesh words annotated on the scatter plot. First, we see that these words are consistent for all languages where they are the highest ranks and have little or no rank change in time. Second, all of them exhibit a rectangular shape showing that high-ranked words tend to go down in ranks while low-rank words tend to go up in ranks. Finally, the extreme case of the WF inspired model shows a radically different outcome. The ratios $c/\beta$ are low for all the languages and the ``extreme" case of WF inspired model.}
	\label{fig:rc-sum_vs_initial-ranks_la_4x3}
	\includegraphics[width=\textwidth]{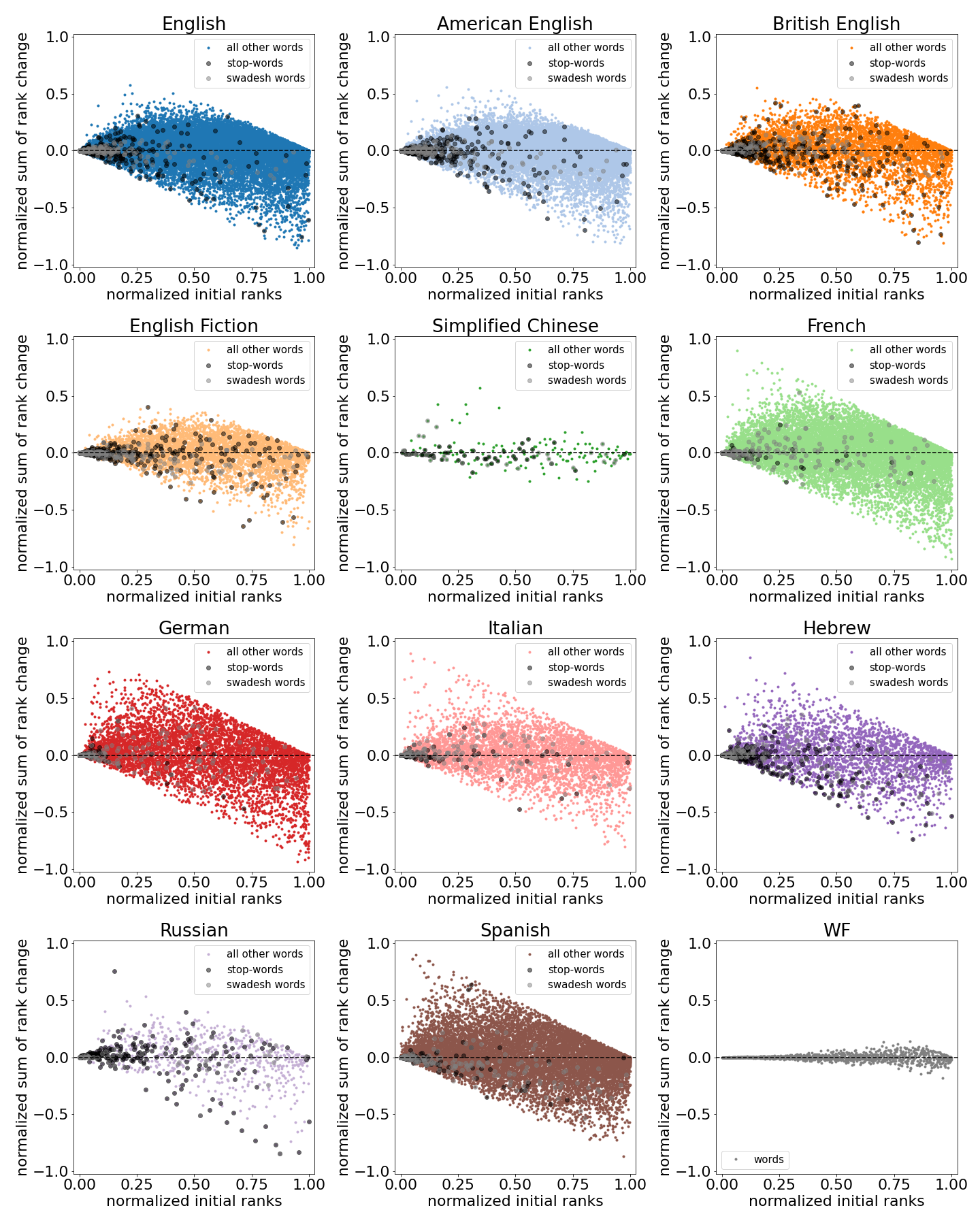}
\end{figure}

In Fig.~\ref{fig:rc-sum_vs_initial-ranks_la_4x3}, it shows the initial ranks versus the sum of rank change for each language in our work. First, most of the stopwords and Swadesh words are consistently high, with low sums for all languages. Second, the rectangular pattern is present in each language where it means that the high initially ranked words tend to go down in ranks while low initially ranks words tend to go up in ranks. Finally, the Simplified Chinese have the smallest vocabulary size, and the results show that some words have high sums while most of the words have small sums. The rectangular pattern is not a coincidence. The Zipf distribution has a shape parameter $a$ that governs how the probability of the most frequent and the smallest are separated. For $a \gg 0$, the most frequent word can only go down in ranks while the smallest can only go up.

Going back in Fig.~\ref{fig:eng_time-series_initial-vs-sum-variance}, next we look at the initial ranks versus the rank change variance in Subplot \textbf{(b)}. The WF inspired model results on the right Subplot show a curved pattern where the initially high-ranked words have low variance relative to the initially low-ranked words. If the corpus size goes to infinity, the curve will go to zero. The English data with the same ratio $c/\beta = 1.00\times10^{-5}$ shows some similarities and differences. First, there is still a small curve exhibited by the English data. The high-ranked words have mostly low variance, while some of the low-ranked words have high variance. This is the effect of normalization where the words with very high variances skewed the curve to appear flat compared to the ``extreme" case of the WF model. Second, most of the words in English have low variance regardless of where the word is ranked initially. This is similar to the ``extreme" case of the WF model where all of of the words have low variances. Finally, the stop words and Swadesh words are in the high ranks at the initial with a low variance which is expected. Since the WF inspired model is the null model for the evolutionary drift process, the comparison suggests that some words in the English data behave contrary to drift. Similar to the results in Subfig \textbf{(a)}, the majority of the words have a low rank change variance, implying that the majority of the terms in the English data are rank stable.

\begin{figure}[!htbp]
	\centering
	\caption[The initial ranks versus the rank change variance of the Language data compared with the extreme case of the WF inspired model.]{\textbf{The initial ranks versus the rank change variance of the Language data compared with the extreme case of the WF inspired model.} Each of the subplots below shows the initial rank versus the rank change variance for each language in the Google unigram data with their respective stop words and Swadesh words. First, we observe that the languages exhibit a subtle curve. Second, the WF inspired model curve is much more obvious than the Language data even though the ratios $c/\beta$ are extremely low for all the languages and the extreme case of WF inspired model. Finally, the stop words and Swadesh words for each of the languages show consistent behavior where they are the highest-ranked words with low variances.}
	\label{fig:rc-variance_vs_initial-ranks_la_4x3}
	\includegraphics[width=\textwidth]{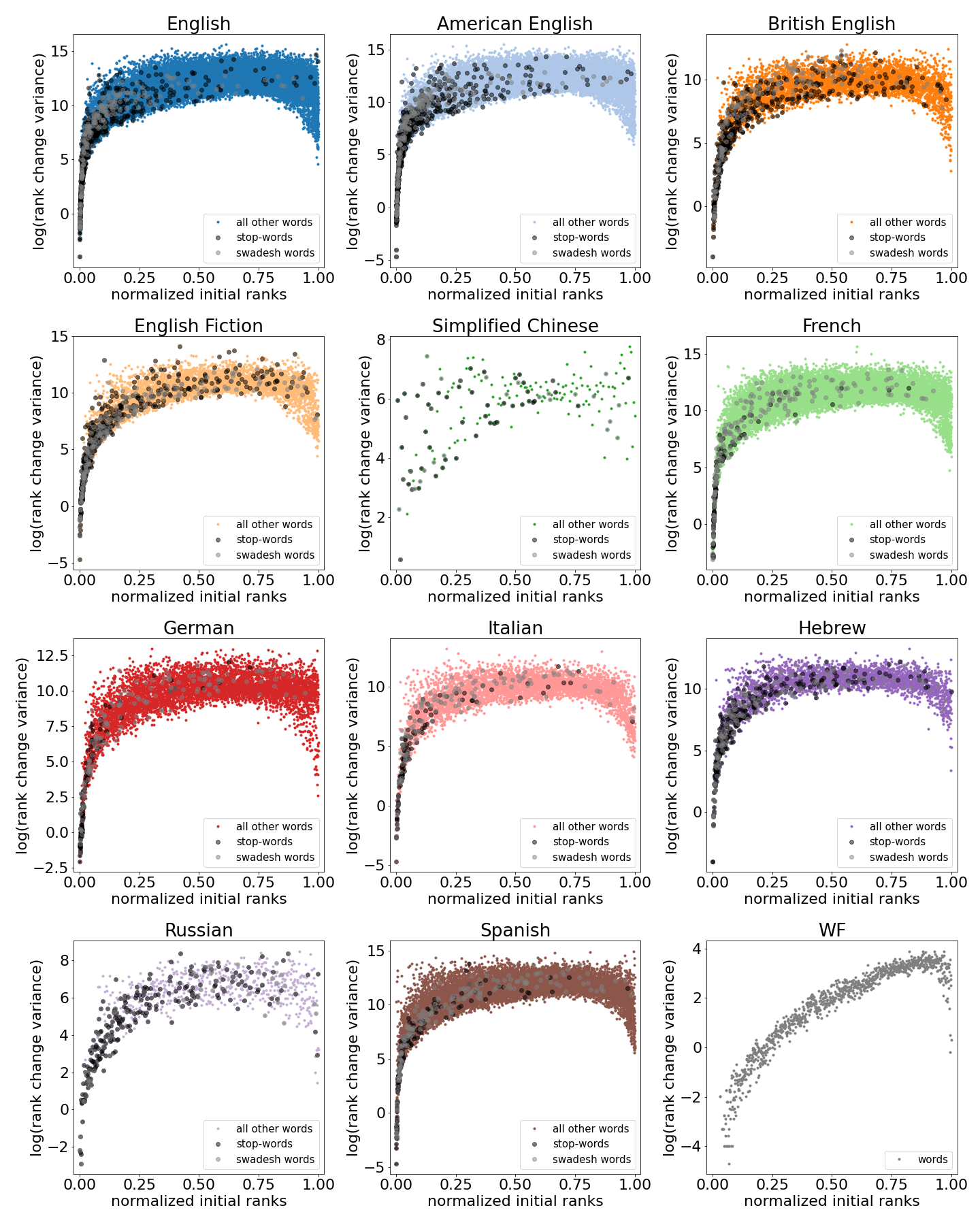}
\end{figure}

On Fig.~\ref{fig:rc-variance_vs_initial-ranks_la_4x3}, it shows the initial ranks versus the rank change variance of the languages we consider in our work. First, each language has this flat curve compared to the WF inspired model. Again, this is due to the normalization process. This curve is consistent for all the languages, which means that most languages are stable regardless of their initial ranks. There are words in the languages with a high variance that skewed the curve to appear flat compared to the ``extreme" case of the WF model. Second, the stop words and Swadesh words in each language are consistently in the high ranks at the initial step and have low variance. These groups of words are expected to be stable in meaning and frequency, and it is shown that most of these words have low sums and low variances in rank change. The results of languages clearly showed that some of the words have high variance regardless of initial ranks compared to the WF inspired model. Generally, both the languages and the ``extreme" case of the WF model have low variances because the initial corpus size is considerable relative to the vocabulary size. The WF inspired model has the ratio $c/\beta=1.00\times10^{-5}$ which is as low as the natural languages. However, the differences in the structure of the scatter plots suggest that the languages that have words - with very high variance relative to other words - behave contrary to an evolutionary drift process.

\begin{figure}[!htbp]
	\centering
	\caption[The RBO curves of the Language data with the extreme case of the WF inspired model.]{\textbf{The RBO curves of the Language data with the extreme case of the WF inspired model.} Subfig \textbf{(a)} shows the RBO curves computed by taking the $RBO$ metric of the ranked list $RL$ at time $t$ versus the ranked list at $t+1$. Subfig \textbf{(b)} shows the RBO curves computed by taking the $RBO$ metric of the ranked list $RL$ at time $t$ versus the ranked list at $t+10$. For both Subfigs \textbf{(a)} and \textbf{(b)}, the RBO curves for the Languages are lower than the extreme case of the WF inspired model, which means that the Languages have words changed in ranks more often than the WF model. Subfig \textbf{(c)} shows the RBO curve computed by taking the $RBO$ metric of the ranked list at the initial time versus the ranked list at time $t$. This also shows that the Languages have words change in ranks much more extremely than the WF inspired model. It is expected for the extreme case of the WF inspired model to behave like the Languages since we set the ratio $c/\beta$ to be close enough like the ratios of the languages. However, we observed a different result which means that some words in the Languages behave differently from an evolutionary drift process.}
	\label{fig:rbo-trends-la-vs-wf}
	\includegraphics[width=\textwidth]{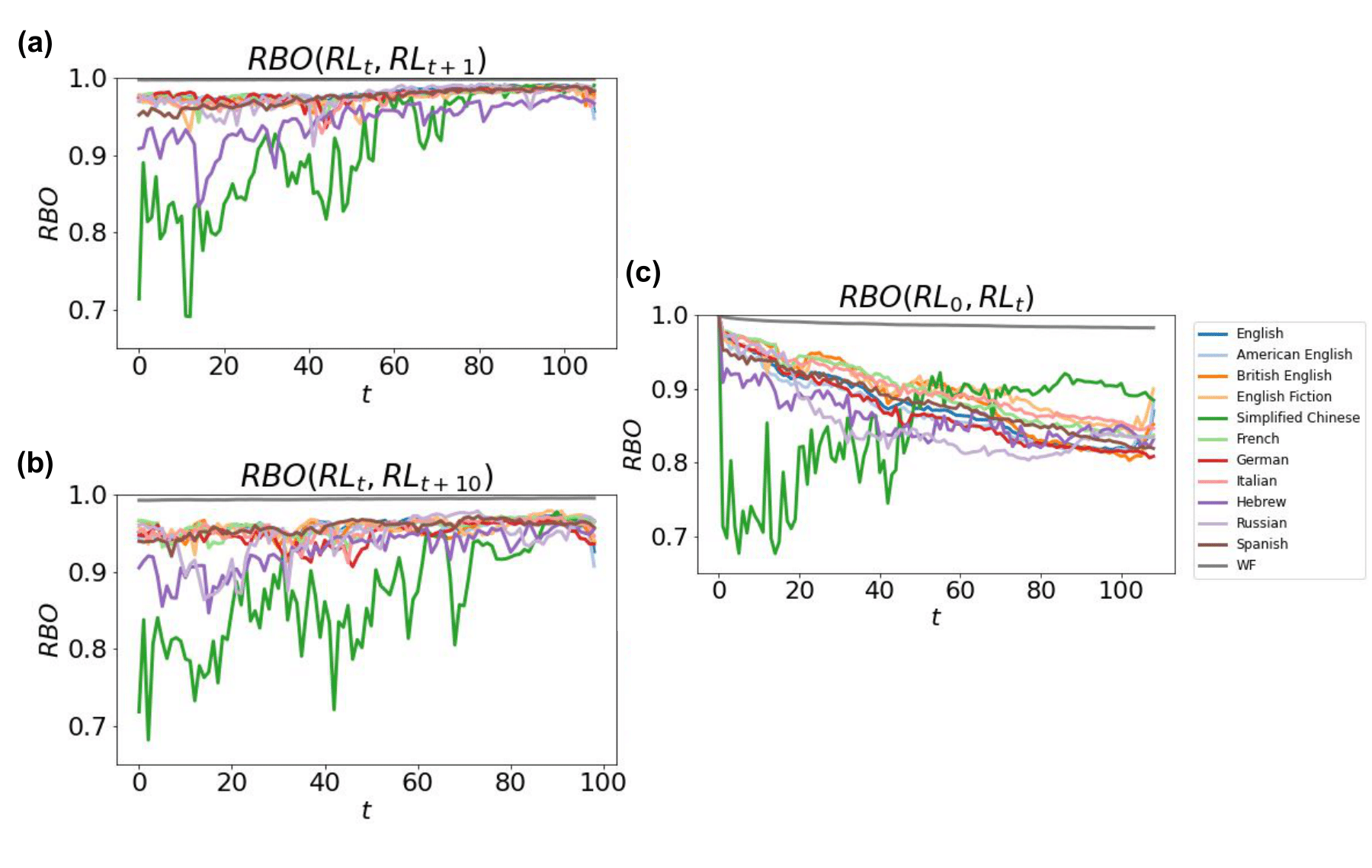}
\end{figure}

To further verify that the languages are different from the WF inspired model, we look to the rank-biased overlap (RBO) curves. RBO computes the similarity between two ranked lists, which is defined in Eq.~\ref{eq:rbo}. Fig.~\ref{fig:rbo-trends-la-vs-wf} show the RBO curves of the languages compared to the ``extreme" case of the WF inspired model. Subfig \textbf{(a)} are RBO curves where the $RL_{t}$ and $RL_{t+1}$ is compared and in Subfig \textbf{(b)} is where $RL_{t}$ and $RL_{t+10}$ is compared. We see in the results that the WF inspired model has an RBO curve that is visually constant close to 1, which means that the ranks of words did not change that much. In comparison, the RBO curves of the languages are much lower, which means that words are changing in ranks significantly. However, the ``extreme" case of the WF model and the languages have a small ratio $c/\beta$ down to $10^{-5}$ in magnitude. As long as the ratio is the same while the vocabulary size and corpus size are different, the RBO curves should be consistent. This is contrary to what we see in the RBO curves of the languages compared to the ``extreme" case of the WF inspired model as shown in Fig.~\ref{fig:rbo-trends-la-vs-wf}. We also observed that the RBO curves appear to be leveling off to the value of $RBO = 1$. This is due to the corpus size increasing in time at the rate of $\alpha$. Recall that as the corpus size increases, the chances of words to change ranks decrease. In Subfig \textbf{(c)} in Fig.~\ref{fig:rbo-trends-la-vs-wf}, it shows RBO curves where the $RL_{0}$ and $RL_{t+1}$. In other words, this computation is comparing the initial ranks and the ranks at time $t$. The results show that as $t$ increases, the RBO is decreasing. It means that the ranks are gradually changing in time. Our observations of the RBO curves tell us that the ranks of the overall structure of the languages are variable. A small accumulation of rank changes leads to the overall ranks changing significantly from the initial ranks. We observed in the results that the RBO curves appear to be leveling off to an unknown RBO value.

\section*{Discussion}

Language users prefer to utilize terms that are used more frequently than other words, according to Pagel et al. (2019) \cite{Pagel2019}. The bias towards word usage based on popularity are evident within a social system leading to stable words \cite{lupyan2010language}. By randomly selecting words, the WF inspired model replicates such phenomena. Higher word frequencies tend to stay at the same place in the rankings, but lower word frequencies fluctuate. Higher ranked words tends to be stable in time than lower ranked words. The model simulates unigram frequency evolution dominated by the evolutionary drift process, a process controlled by frequency-dependent sampling. We demonstrated in our model that smaller corpus size gives stronger drift effects consistent with the neutral theory of molecular evolution. Our findings indicate that the majority of words in the data tends to be rank stable indicating words may fluctuate in ranks but tends to fluctuate around a net change of zero. This may be a purely a statistical effect rather than a linguistic effect because of how we computed the sum of rank change metric. The rank change variance measures the variability of the rank changes where we observed that initially high-ranked words have low variance than low-ranked words.

This current paper, Pagel et al. 2007 \cite{Pagel2007}, and 2019 \cite{Pagel2019} found that frequently used words evolve slowly while infrequently used words evolve faster. Our results were based on word rankings rather than raw frequency. We also found that depending on where the word is initially ranked, high-ranking words tend to fall in rank and low-ranked words tend to rise in rank, while low-ranked words are more likely to shift ranks. Instead of utilizing cognate terms, we compared the rank changes of words across languages using stop-words and Swadesh words. Stop-words and Swadesh words comprise coordinating conjunctions and prepositions, and our results show that these highly rated words have relatively low rank fluctuations. This part of the result is opposite to Pagel et al. 2007 \cite{Pagel2007} findings where prepositions and conjunctions evolve more quickly. We think that these difference in results lies within the fact that they are using raw frequency rather than word ranks. We explained in this current paper that a word frequency may vary a lot but if the word frequency variations are far enough from a nearby word, the chances of overlapping is small, and thus the word rank chances to change is small. This is the case for most of the Stopwords and Swadesh words in our study.

In our model, the corpus size is assumed to exponentially increase but the vocabulary size stays constant in time. This minimal assumption yielded a stable rank behaviors among words. We discovered in our model that when the initial vocabulary size is large enough, there is a lot of rank variation between words. It indicates that the words are more likely to shift rankings, particularly for lower-ranked words. The behavior of the model gets more complicated and influences the evolution of word rankings if the vocabulary size is varied over time - with word births and deaths. According to Petersen et al. (2012) \cite{Petersen2012}, word births can cause an increase in volatility in word growth rates, which is consistent with our findings that a rise in vocabulary words at the initial leads to increased competition among words. As the complexity of a language grows, so does the vocabulary \cite{petersen2012languages}. If more vocabulary words are added into the model, the rank change variances get stronger. We think the entropy also increases as more vocabulary words are added which is consistent with past studies \cite{Sun2021,Pilgrim2021}.

We conducted a parameter estimation analysis for the $c$ and $\beta$ parameters through a scaling ratio $\frac{c}{\beta}$. We attempted to fit a critical scaling parameter $\frac{c}{\beta}$ to simulate language on the scale of real languages. We note that it is not computationally possible to simulate the WF model on the scale of real languages. The parameter study was done to explore what happens as the parameters move more in the direction of real languages. The value of $\frac{c}{\beta}$ is an important indicator of how words behave in the model. Because the multinomial expected values and variances increase in time, these overlaps in the binomials get stronger, introducing swaps in ranks. The unigram ranks for low initially ranked words have higher chances to change ranks than the high initially ranked words. If the ratio $\frac{c}{\beta}$ gets closer to 0 - or $\beta$ is significantly larger than $c$ - the probability of any two words to change ranks decreases. If $\frac{c}{\beta}$ gets closer to 1 - or $c$ gets as large as $\beta$ - the probability of any two words to change ranks increases. The natural languages have extreme $\beta$ and very low $\frac{c}{\beta}$. The WF model would predict that the rank changes in the natural languages would be low, and the rank change variance would be low if a word is initially high ranked. However, the results show that the languages behave contrary to what the WF model predicts using extreme parameter values of $\beta$. The lexicons of all languages behave as if they are projected in a much smaller corpus - they are given considerably more fluctuation than their massive corpus would predict. The corpus size is large such that the WF model predicts extreme stability. Instead, languages may be adapting to fluctuations in cultural and other environmental features that drive rank-order changes.

The Ranked Biased Overlap (RBO) metric, which measures two ranked lists' similarity. Interpreting the RBO can be viewed as the percentage of words that - on average - stayed in ranks. For example, an RBO of 0.95 means that 95\% of words from $t$ and $t+1$ stayed relatively the same rank. It appears that the RBO time-series are increasing in comparing $t$ to $t+1$, and $t$ to $t+10$ ranked lists, meaning words in the data are becoming rank stable in time. This results supports the hypothesis that some words (e.g. concrete words) are becoming rank stable. The RBO timeseries comparing $t=0$ to $t$ are shown to decrease, meaning the initial word rankings have change in time. This result supports the observations that there is a rise in entropy in language. As language expands and becomes more complex, the rankings of words changes. The RBO time-series of the WF model indicates extreme stability in word ranks for very small $\frac{c}{\beta}$. Compared to the WF model with the same vocabulary to corpus size ratio, the average change in ranks is lower in real languages than the WF model, which means that some words tend to deviate from the expected behavior of neutral drift evolution. In general, the RBO is showing us that the language in our analysis is approaching a steady-state distribution (established languages such as English tend to approach stability) but far from the steady-state of the WF model. The standardization of these languages might be the main contributor to why the RBO tends to approach a steady-state distribution. These results seem to support that there is a rise in concreteness in language, where it allows us to communicate complex ideas more effectively while also adding complex words for complex ideas, which increases the entropy of language.

Our study used the RBO metric to study word rank evolution in our WF inspired model and real languages. Word rank evolution was modeled before and used the turnover metric as a measure of how many words are entering the top $y$ list. While our study is not focused on the turnover metric, we conducted some quick computations to check for consistency with our results. We used the generic turnover function $z = ay^b$ by Evans and Giometto (2011) \cite{Evans2011} where $z$ is the turnover, $b$ is the shape, and $a$ is a coefficient. Parameter estimation of $b = 0.86$ (moderately concave down) was presented by Acerbi et al. (2014) \cite{Acerbi2014} and Ruck et al. (2017) \cite{Ruck2017}, where they argued that it represents an estimate of unbiased copying (pure neutrality). The generic turnover function is a simplification from a precise turnover function \cite{Evans2011}. If $b > 1$ (concave up), meaning there is a conformity bias of word rank evolution. If $b < 1$ (concave down), meaning there is an anti-conformity bias. Our quick WF-inspired simulations show that there is an association of the ratio $c/\beta$ in our model with the turnover parameter $b$. If $c/\beta$ is close to 1, then $b > 1$ (conformity). If $c/\beta$ is close to 0, then $b < 1$ (anti-conformity). See \nameref{S18_Fig} for the turnover shapes of our WF inspired model. We also estimated the value of $b$ of the real languages and the results show all of the languages have concave up turnover shapes except for Simplified Chinese, which exhibits moderate concave down. The expectation for the turnover shapes of these languages should be similar to an extreme conformity bias if the languages follow neutrality due to its ratios $c/\beta$ being extremely small. Most languages of our study exhibit turnover curves that show moderate conformity with $b > 0.86$, meaning languages are exhibiting some non-neutral behavior. In comparison to the RBO metric, an RBO value close to 1 is equivalent to conformity bias, meaning little or no change has occurred in the word ranks. An RBO value close to 0 is equivalent to anti-conformity bias, meaning some changes have occurred in the word ranks. We would like to add that the turnover metric merits further investigations in the context of our model but these quick results show consistency with the results of our paper. We show the $b$ parameter estimations of the languages in \nameref{S19_Fig}.

\section*{Conclusion}

We conclude - even in neutral evolution of word frequencies - a word’s rank change shows one of two types of possible characteristics in these data: (1) the increase or decrease in rank is monotonic, or (2) the rank stays the same. High-ranked words tend to be more stable, while low-ranked words tend to be more volatile. Among those words that change rank, some change in two ways: (a) by an accumulation of small increasing/decreasing rank changes in time and (b) by sudden shocks of increase/decrease in ranks. Most of the stopwords and Swadesh words are observed to be stable in ranks for the eight natural languages we studied (this is not meant to imply that these groups have the same meaning). In general, our WF model captures some but not all of these trends, as the sudden change that some words are given to depart from the neutral WF model.

The unigram frequency and ranks showed deviations from the evolutionary drift process. Due to their functionality, stop words and Swadesh words might be the words that are ``fixed" to maintain stability in a language system. Many words have low-rank changes similar to stop words and Swadesh words. We also have seen words that have changed up or down in ranks significantly for all the languages we considered. Since language is tied to culture, these words are probably ``selected" to serve an important function in cultural change. Did these words change in ranks entirely by chance, or is it the result of natural selection? It is difficult to say with certainty that the words in the data that behave in this particular way are naturally selected. Testing for selection requires more than just comparing the data to a null model that simulated drift behavior. However, we are confident that the unigram frequency of words and their ranks does not just behave like the drift evolutionary process but also shows peculiar behaviors unexplained by drift. The deviations from neutrality that we conclude are consistent with results of previous studies \cite{Pagel2019, turney2019natural, newberry2017detecting, o2017inferring, Sindi2016, Blythe2012}. Our work provides a simple mathematical framework of the Wright-Fisher inspired model to explain word rank behaviors.

\section*{Future work}

First, the Wright-Fisher inspired model can be modified to incorporate selective variation and varying vocabulary size in time. This modification could test the model of whether behaviors in the natural languages are caused by external forces such as the environment. By adding a ``fitness" function into the model, it can add more complexity into the model and control how it evolves for each word. This would make the model a better ``fit". In our current model, we assumed that the vocabulary size stayed the same in time. By varying the vocabulary size in time, this can introduce noise into the model and would generate an interesting rank behaviors similar to the real languages.

Second, a future work for this model could include trying out a much shorter time-scale such as in months. The current model simulates word frequency evolution in year time-scale. A recent study suggests that care should be exercised when binning text data into different time scales because it may introduce errors in interpreting the results when testing for selection \cite{Karjus2020}. 

Third, the Google Books Corpus is not the only large-scale historical linguistic corpora. It would be a good idea to see other data sets such as the Corpus of Historical American English (COHA) \cite{Alatrash2020} and the Standardized Project Gutenburg Corpus (SPGC) \cite{Gerlach2020}.

Moreover, linguistic data sets with temporal features taken from online social media platforms such as Twitter, Facebook, or Reddit are possible data sets to consider when studying language evolution in shorter time scales. These data sets can reveal interesting patterns of word ranks since these potential data sets are a representative sample of 21st-century human cultural phenomena around the world. Language transmission has been observed and modeled using Twitter data by Bryden et al. \cite{Bryden2018}. In addition, a study by Carrignon et al. (2019) \cite{Carrignon2019} shows that there is a rapid cultural transmission in the spread of true and false information on Twitter. The WF inspired model and its mathematical framework we presented in this paper could be used to analyze the language change of social movements, popular culture, and political discourse.

Furthermore, the words in the Google Ngram Data were annotated with Part-of-Speech (POS) tags, but the frequencies of these tags are combined. It would be interesting to see the differences in rank changes of these POS word groups.

Finally, the next step to studying word rank evolution is to consider the n-grams for $n > 1$ to see if the n-grams rank changes behave similarly to the Wright-Fisher inspired model.

\newpage

\section*{Supporting Information}

\section*{S1 Appendix}
\label{S1_Appendix}
For easy implementation of downloading and processing the raw files of the Google Ngram data and the computations done in this paper, visit \href{https://github.com/stressosaurus/a-statistical-model-of-word-rank-evolution/}{github.com/stressosaurus/a-statistical-model-of-word-rank-evolution/}.

\section*{S2 Appendix}
\label{S2_Appendix}
For quick and easy implementation of the RBO, you can follow the work by Changyao Chen which is available at \href{https://github.com/changyaochen/rbo}{github.com/changyaochen/rbo}.

\section*{S3 Appendix}
\label{S3_Appendix}
To fit the corpus size time-series into the corpus size function, we use a log transform on Eq.~\ref{eq:corpus-size-function} to make it linear.
\begin{equation}
	\ln{(N(t))} = \alpha t + \ln{(\beta)}
	\label{eq:corpus-size-function-log}
\end{equation}
where $N(t)$ is the corpus size at time $t$, $\alpha$ is the rate of increase, and $\beta$ is the initial corpus size.

To fit the initial frequencies into the Zipf probability mass function, we use a log transform on Eq.~\ref{eq:zipf-pmf}.
\begin{equation}
	\ln{\left(P(Y^{theory})\right)} = -a\ln{\left(r_{w}\right)}  - \ln{\left(\sum_{w=1}^{c} (1/r_{w}^a)\right)}
	\label{eq:zipf-pmf-log}
\end{equation}
where $Y^{theory}$ is the random variable for the ranks, $c$ is the vocabulary size, and $r_w$ is the rank of word $w$.
The above equation indicate that the log transform is a linear equation with slope $-a$ based on the first term. The second term is the intercept. We can rewrite this as
\begin{equation}
	y = -a\ln{\left(r_{i}\right)} + b
	\label{eq:zipf-pmf-linear}
\end{equation}
where $a$ and the $b$ are the parameters we can estimate using the log-transformed data. The shape parameter $a$ is always positive and $b$ can be a negative number. 

The linear models (Eq. \ref{eq:corpus-size-function-log} and \ref{eq:zipf-pmf-linear}) are the equations used to fit the corpus size time-series and the initial frequency distribution respectively using the \textit{scipy.optimize.curve\_fit} module in Python \cite{2020SciPy-NMeth}.

\section*{S4 Appendix}
\label{S4_Appendix}
Following from the work of Sindi and Dale \cite{Sindi2016}, the unigram frequency data is standardized. Given a set of words $V = \{w_1, w_2, \cdots, w_c\}$ and years $Y=\{t_{0},t_{1},t_{2},\cdots,t_{T-1}\}$ where $T$ is the number of years. The frequency of a word $w$ in a corpus at time $t$, $r_{w,t}$, is the number of occurrences of that word in that corpus. In our analysis of the Google unigram data, we selected only words that occur above a desired frequency in the year interval $(1900,2008)$. As such, the vocabulary size, $c$, is fixed as is the number of years $T$ and we thus represent the word frequencies as a matrix:  $\mathbf{R} \in \mathbb{R}^{c \times T}$ where
\begin{equation}
	\label{eq:rscore}
	\mathbf{R}_{w,t} = r_{w,t}, \hspace{10px} r_{w,t} \ge 1.
\end{equation}

In our normalization process, we first convert the frequency matrix $\mathbf{R}$ into a proportion (or relative frequency) matrix $\mathbf{P}$ by normalizing the columns of $\mathbf{R}$ which normalizes word frequencies by year:
\begin{equation}
	\mathbf{P}_{w,t} = \hat{p}_{w,t}, \hspace{10px} \hat{p}_{w,t} = \frac{r_{w,t}}{\sum_{w=1}^{c} r_{w,t}}.
	\label{eq:pscore}
\end{equation}

Finally, we normalize the proportions for each unigram by converting the rows of $\mathbf{P}$ into $z$-scores:
\begin{equation}
	\mathbf{Z}_{w,t} = z_{w,t}, \hspace{10px} z_{w,t} = \frac{\hat{p}_{w,t} - \overline{\hat{p}_{w}}}{\sigma_{\hat{p}_{w}}}
	\label{eq:zscore}
\end{equation}
where $\overline{p_{w}}$ is the mean and $(\sigma_{p_{w}})^2$ is the variance;
\begin{equation}
	\overline{p_{w}} = \frac{1}{T} \sum_{t=0}^{T-1} \hat{p}_{w,t}
	\label{eq:p-mean}
\end{equation}
and
\begin{equation}
	(\sigma_{p_{w}})^2 = \frac{1}{T} \sum_{t=0}^{T-1} \left(\hat{p}_{w,t} - \overline{p_{w}}\right)^2.
	\label{eq:p-variance}
\end{equation}

\section*{S1 Figure}
\label{S1_Fig}
\textbf{The binomial probabilities at the initial time.} This figure shows the binomial probabilities of four words. At $a=0$, the Zipf distribution reduces to the uniform distribution. Since the corpus size is fixed, the mean of each binomial is the same. If $a=0.05$, the binomial distributions of these four words are separated, and they are further separated at $a=1$. The binomials are also separated based on the shape parameter of the Zipf distribution. The word with the highest probability corresponds to the highest-ranked word. The Subfigures below consists three plots where it shows the binomial probabilities of words with varying Zipf shape parameter. The increase in $a$ resulted in separation of the binomials.
\includegraphics[width=\textwidth]{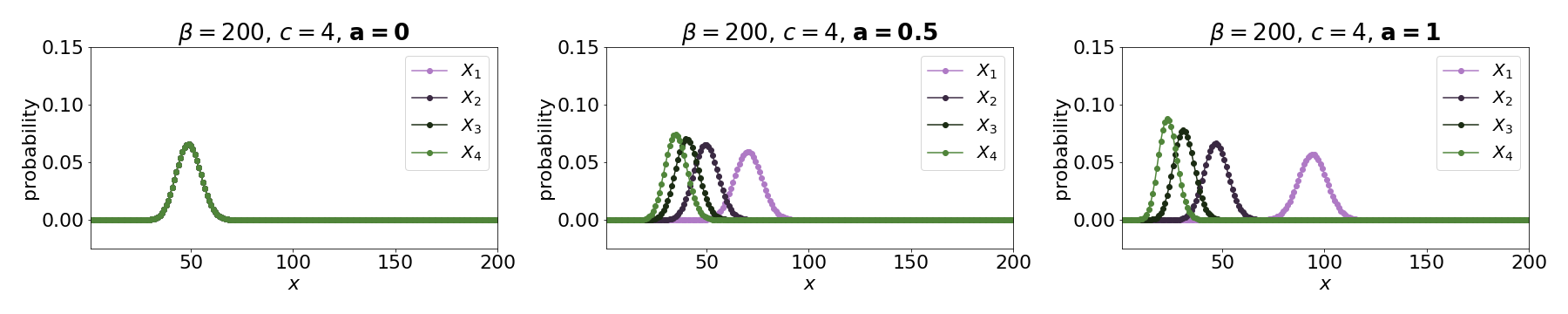}

\section*{S2 Figure}
\label{S2_Fig}
\textbf{The binomial probabilities with varied $c$ and $\beta$ at the initial time.} This figure shows the binomials of the words as the vocabulary size increases. The binomials at $c=2$ are as expected since the Zipf distribution with a shape parameter of $a=1$ or greater would force the two words to be separated. At $c=4$ where the vocabulary size is four, the binomials are now overlapping since the probabilities of some of these words are close enough for them to have a chance of getting the same for close frequencies. At $c=6$ where the vocabulary size is six, the binomials of the less frequent words are forced to overlap since the lowest probabilities of the Zipf distributions have shorter separations between them. This will give less frequent words to have higher chances of having the same or close frequency. In short, the binomial curves are shown to overlap if the vocabulary size is increasing while the corpus size is fixed. The increase in vocabulary size captures the intuitive idea of competing words. The corpus size is the total number of words, meaning that the words have more space to fill in. As seen at $\beta=600$, the separation between the binomial curves has increased, and the overlaps get narrower. This will have less frequent words to have lower chances to have the same or close frequencies. The parameters vocabulary size and corpus size are vital parameters. These parameters can change the separation between the binomials and the probability that two words can overlap in frequency. This figure is showing these behaviors more clearly. It shows that the binomials are more separated if the corpus size increases, while the chances of any two words overlap in frequency increases if the vocabulary size increases. The ratio of $c/\beta$ is a metric that indicates the balance between the vocabulary size and the corpus size. If the ratio is small, the corpus size is large enough to indicate that the binomials of words have fewer cases where there are overlapping frequencies. The Subfigures below are cases of different parameter values showing overlapping binomial distributions, which contribute to rank errors when sampling. Each row of Subfigures has the same ratio. The direction to the right is where the vocabulary size increases, while the direction downwards is where the corpus size increases.
\includegraphics[width=\textwidth]{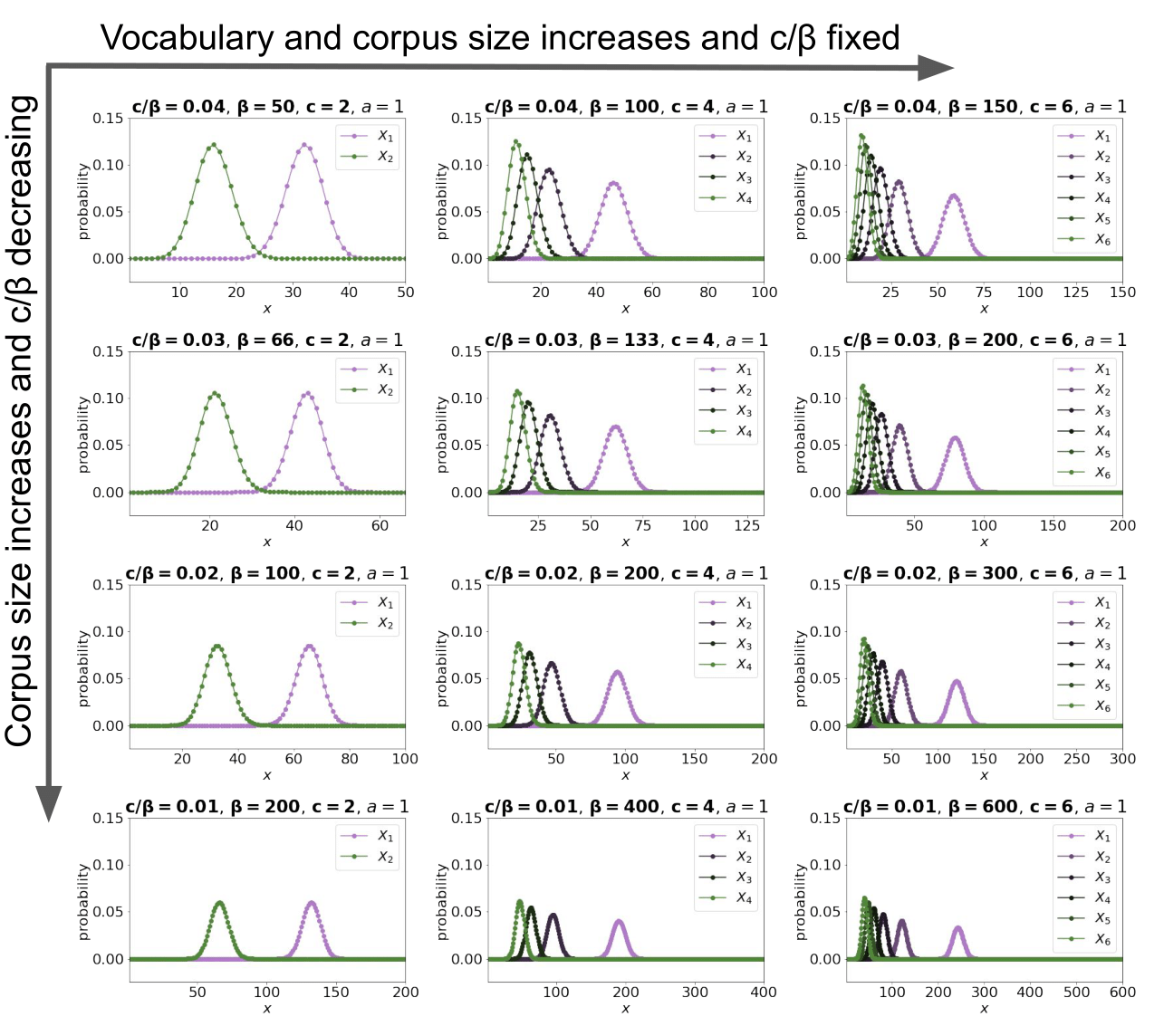}

\section*{S3 Figure}
\label{S3_Fig}
\textbf{The binomial overlaps with varied $c$ and $\beta$ at the initial time.} In this figure, we show the intervals of these binomial cases where each interval is computed with $\mu \pm 4\sigma$. Based on the intervals when the corpus size increases while the vocabulary size is 2, the separation of these intervals widened. The number of overlaps for each interval reduces to zero. Similarly, with the vocabulary size of 4 and 6, the number of overlaps between the intervals decreased. For the interval for the most frequent word, it looks like the separation from it to other intervals widens faster than the less frequent words as the corpus size increases. There are more overlapping intervals if the vocabulary size is large enough, but if the corpus size is much larger than the vocabulary size (or the ratio $c/\beta$ is low), the number of overlaps of these intervals decreases. The Subfigures below are intervals based on 4 standard deviations from the mean of the binomial distributions. Each interval are computed using $\mu \pm 4\sigma$. The value $\widehat{ol}$ as labeled is the net potential rank change. The sign indicates the direction of the rank change. Positive $\widehat{ol}$ means that a word has the net potential to go down in rank while a negative $\widehat{ol}$ is the opposite.
\includegraphics[width=\textwidth]{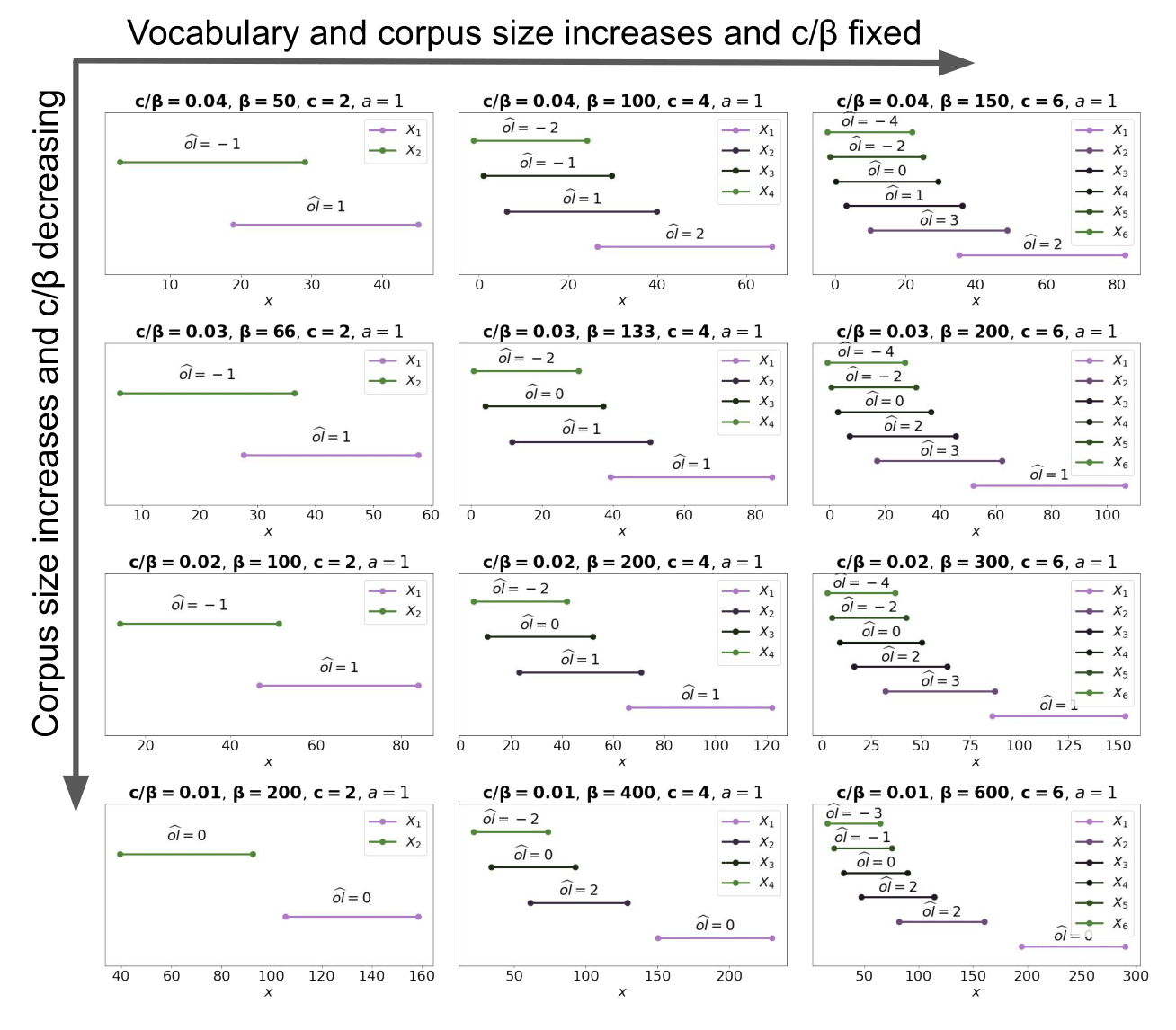}

\section*{S4 Figure}
\label{S4_Fig}
\textbf{The binomial probabilities with 1000 WF simulations with parameters $\beta=200$, $c=4$, and $a=1$.} This figure is showing that at $t=0$ the Wf simulation are right on the binomial curves as expected with the observed separation between the words and their overlaps. Since sampling words from the previous generation simulate the WF, the behavior of the samples at $t=5$ and $t=20$ appears to be widening, and the chances of any two words to overlap increases. The Subfigures below shows the theoretical binomial curves of four words compared against the WF simulations (in shaded bars) at time points $t=0$, $t=5$, and $t=20$ (by row) and at different corpus size rate increase $\alpha=0$, $\alpha=0.01$, and $\alpha=0.03$ (by column). The $\beta$ parameter is the initial corpus size, while the $\alpha$ parameter is the rate of corpus size increase. As time moves forward, the WF simulations are widening. We can also see that as $\alpha$ increases, the WF simulations distributions move to the right. While the multinomial probability transitions of the Markov chain results in the accumulation of sampling errors, the expected value and variance of the raw counts also increases as the corpus size increases in time. The binomial curves are observed to shift right as time increases and as the corpus size increases.
\includegraphics[width=\textwidth]{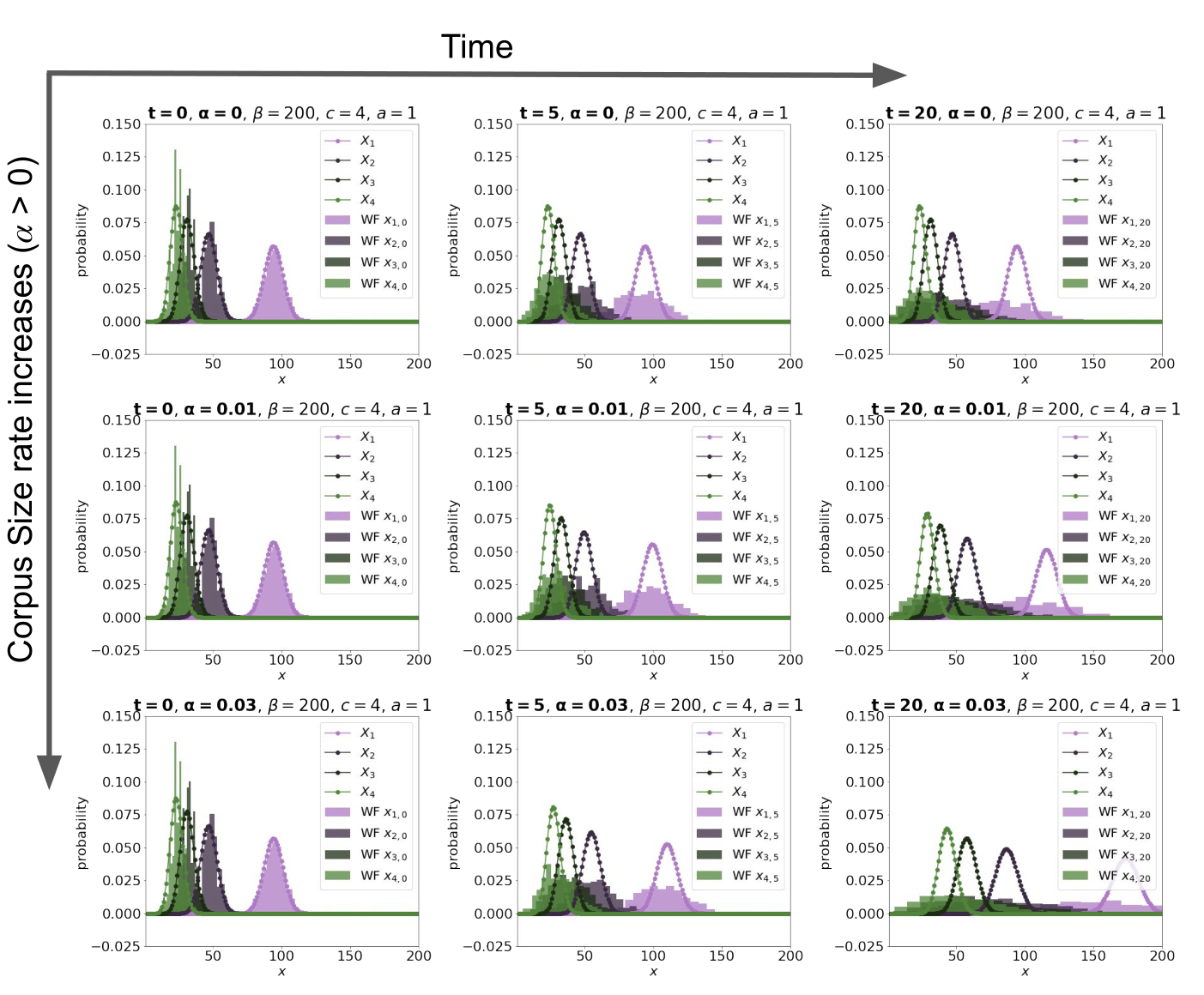}

\section*{S5 Figure}
\label{S5_Fig}
\textbf{100 Simulations of the WF inspired model with $\beta$ varied and fixed $\alpha=0.01$, $c=2$, and $a=1$.} Looking the subplots below from top to bottom, the simulations for the largest initial corpus size showed that the words stayed in their initial ranks in time. In the smallest initial corpus size, there are outcomes where the words changed ranks in time.
\includegraphics[width=\textwidth]{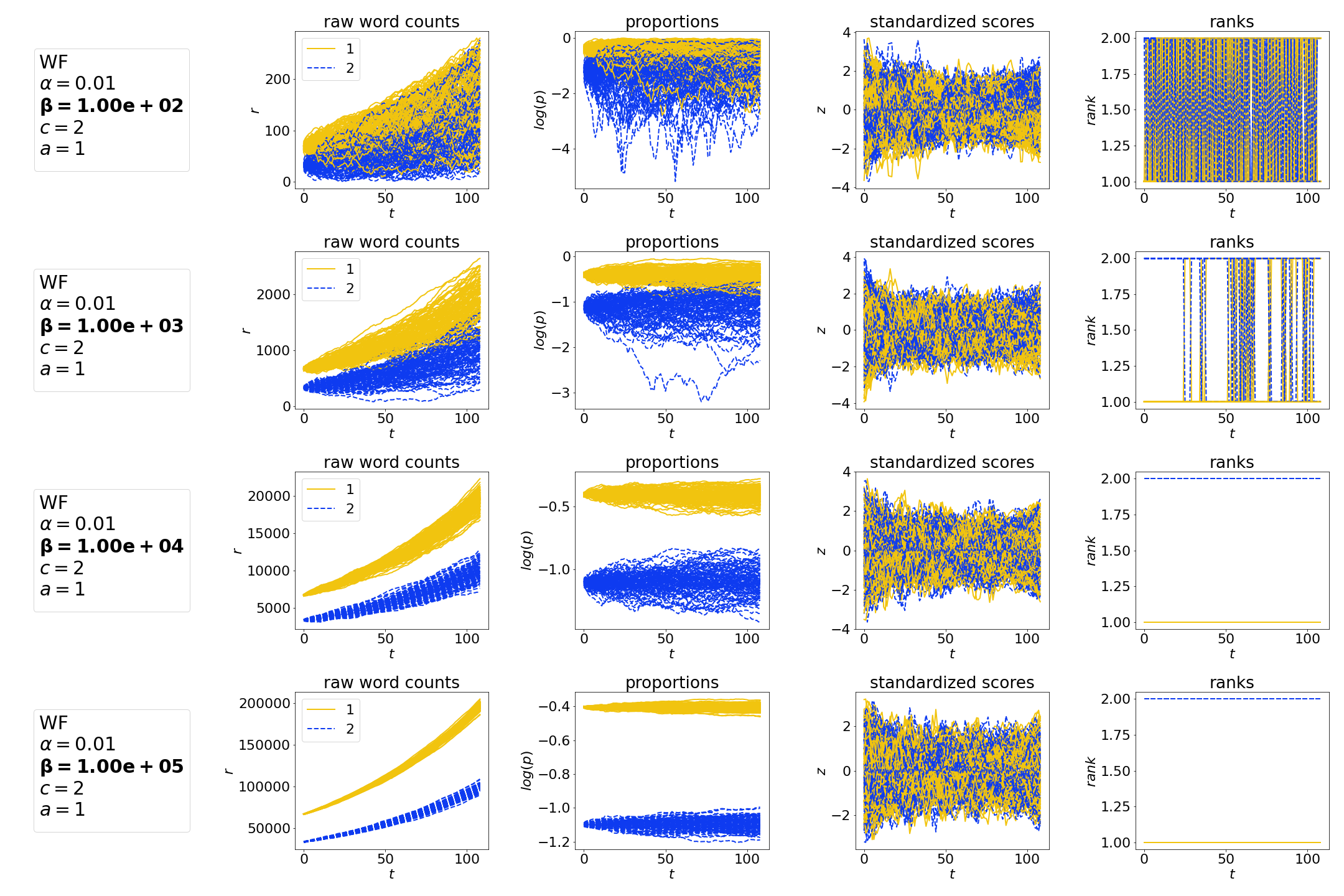}

\section*{S6 Figure}
\label{S6_Fig}
\textbf{100 Simulations of the WF inspired model with $c$ varied and fixed $\alpha=0.01$, $\beta=1.00\times10^4$, and $a=1$.} Looking the subplots from top to bottom, the simulations for the increasing vocabulary size showed that there are outcomes for the lower ranked words that changed ranks in time.
\includegraphics[width=\textwidth]{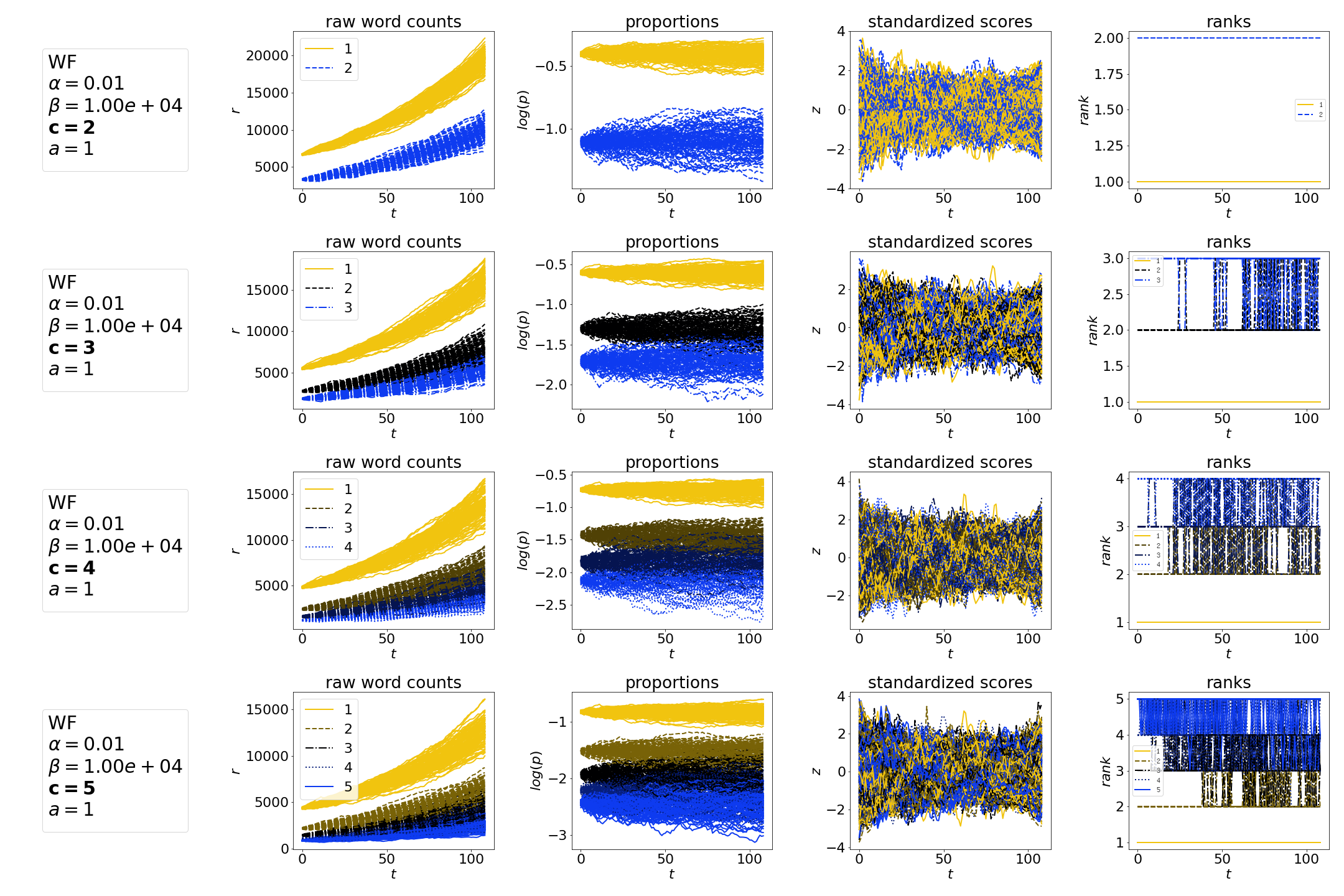}

\section*{S7 Figure}
\label{S7_Fig}
\textbf{WF inspired single model simulations of different cases of parameter sets showing the RBO trends.} The subplots below show six cases of WF simulation where each case has fixed parameter values and a varied parameter. Looking at the subplots by row, the RBO curve shows predictable patterns. Case 1 - with the corpus size rate $\alpha$ is varied - showed the RBO curve to "level-off" or shifted slightly upward as $\alpha$ increases, which means that the ranks are changing less as the corpus size increases. Similarly, Case 2 - with the initial corpus size $\beta$ is varied - showed the RBO curve to shift up more strongly than Case 1. Case 3 - with the vocabulary size $c$ is varied - showed the RBO curve to shift down as $c$ increases. As the vocabulary size increases, the words are more likely to change ranks because there is more competition among words. Cases 4 and 5 - with the ratio $c/\beta$ is constant while $c$ and $beta$ are increasing - showed consistent RBO curves. This means that the overall structure of the words remained largely consistent as long the ratio of $c/\beta$ remained the same regardless $c$ and $\beta$ increases or decreases. Similarly, Case 6 - with the Zipf shape $a$ parameter is varied - showed RBO curves not changing as $a$ increases.
\includegraphics[width=\textwidth]{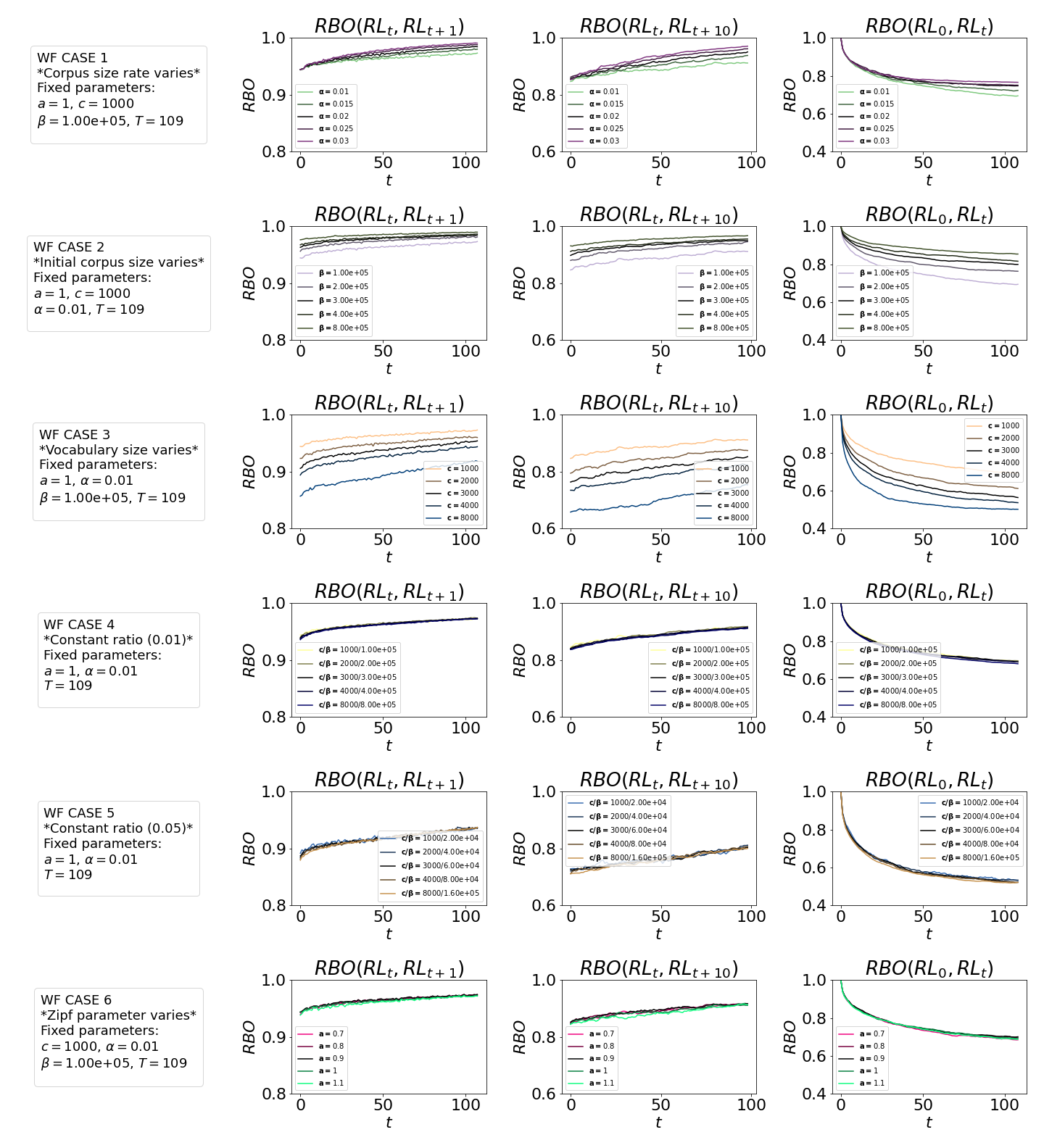}

\section*{S8 Figure}
\label{S8_Fig}
\textbf{Log transformed corpus size function fits against the log transformed language data.} Each yearly corpus size for each language was used to estimate the parameters $\alpha$ and $\beta$ in the corpus size function in Eq.~\ref{eq:corpus-size-function}. The parameter estimation of the log transformed exponential function was done using generalized least squares.
\includegraphics[width=\textwidth]{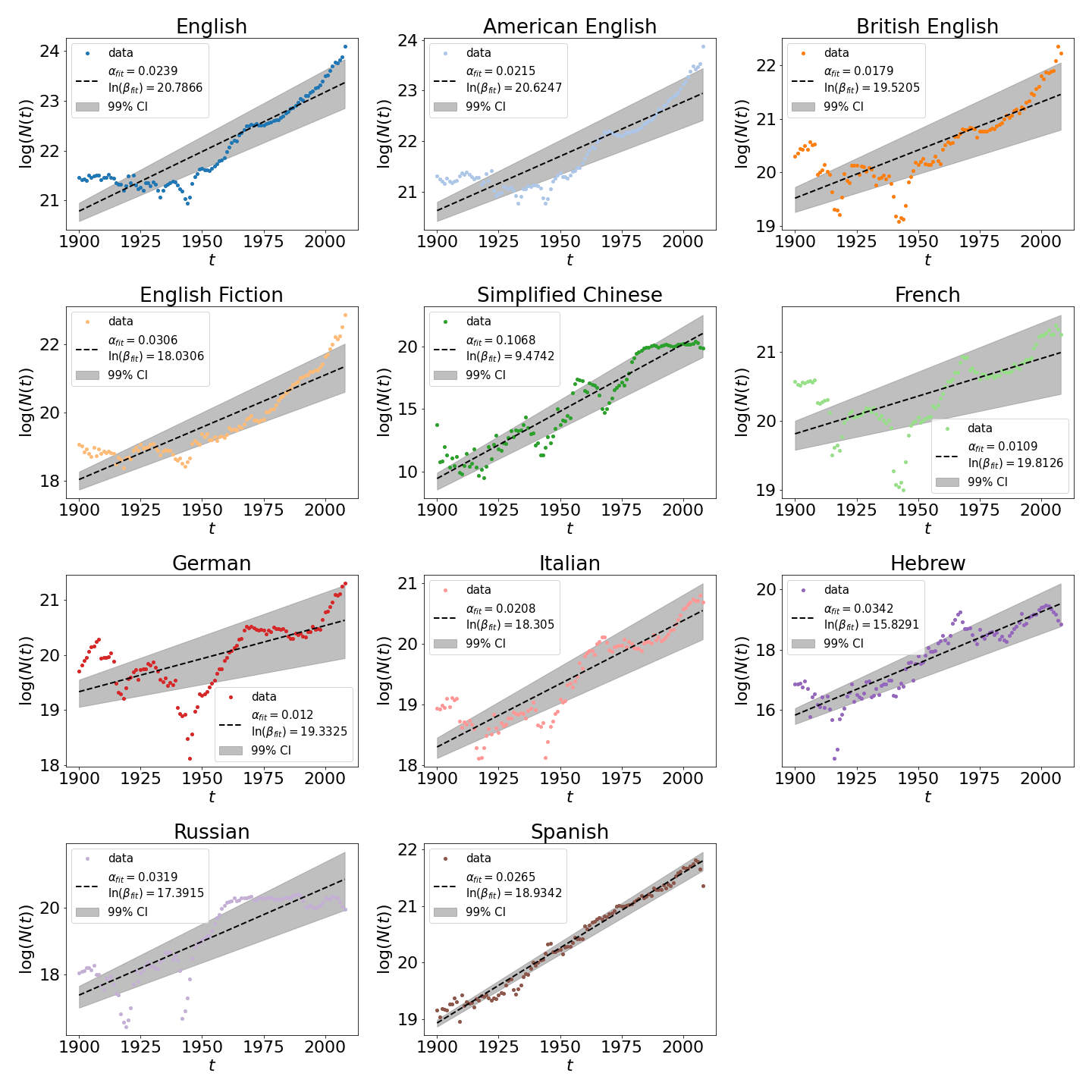}

\section*{S9 Figure}
\label{S9_Fig}
\textbf{Log transformed Zipf function fits against the log transformed language data.} The initial rank distribution was used to estimate the shape parameter $a$ of the Zipf probability mass function in Eq.~\ref{eq:zipf-pmf}. The data and the Zipf function are log transformed prior to fitting using generalized least squares.
\includegraphics[width=\textwidth]{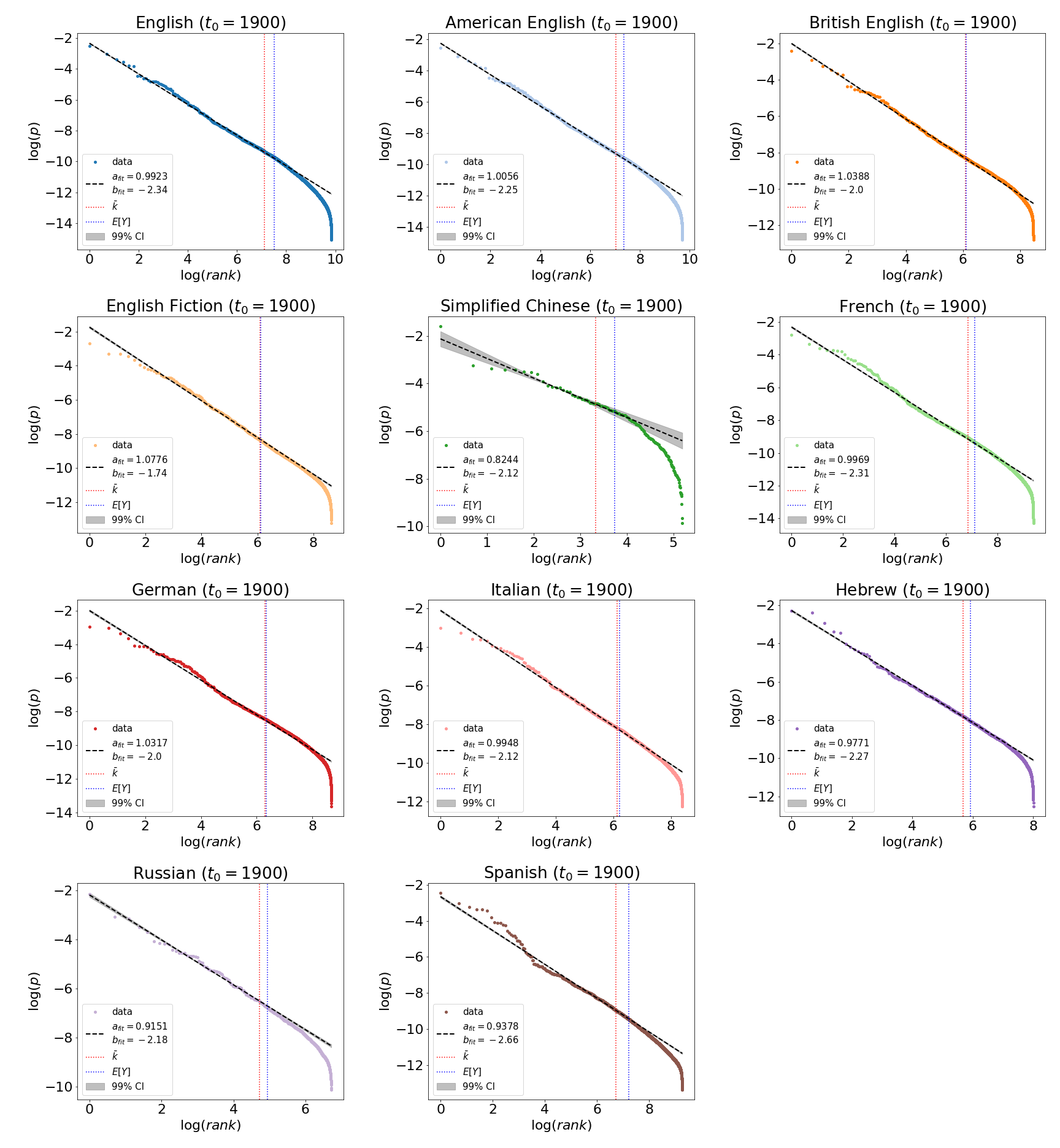}

\section*{S10 Figure}
\label{S10_Fig}
\textbf{The sum of rank change distributions of the languages American English, British English, and English Fiction.} Each distribution below are the sums of the rank change of the words in the languages in the Google Ngram data. For each language, the distributions are annotated on the left tail (\textbf{A}), center (\textbf{B}), and right tail (\textbf{C}) of the distribution to show the list of words corresponding to the values of the sum. The words in list \textbf{A} are words that changed up in ranks. The words in list \textbf{B} are words that have little or no change in ranks. The words in list \textbf{C} are words that changed down in ranks.
\includegraphics[width=\textwidth]{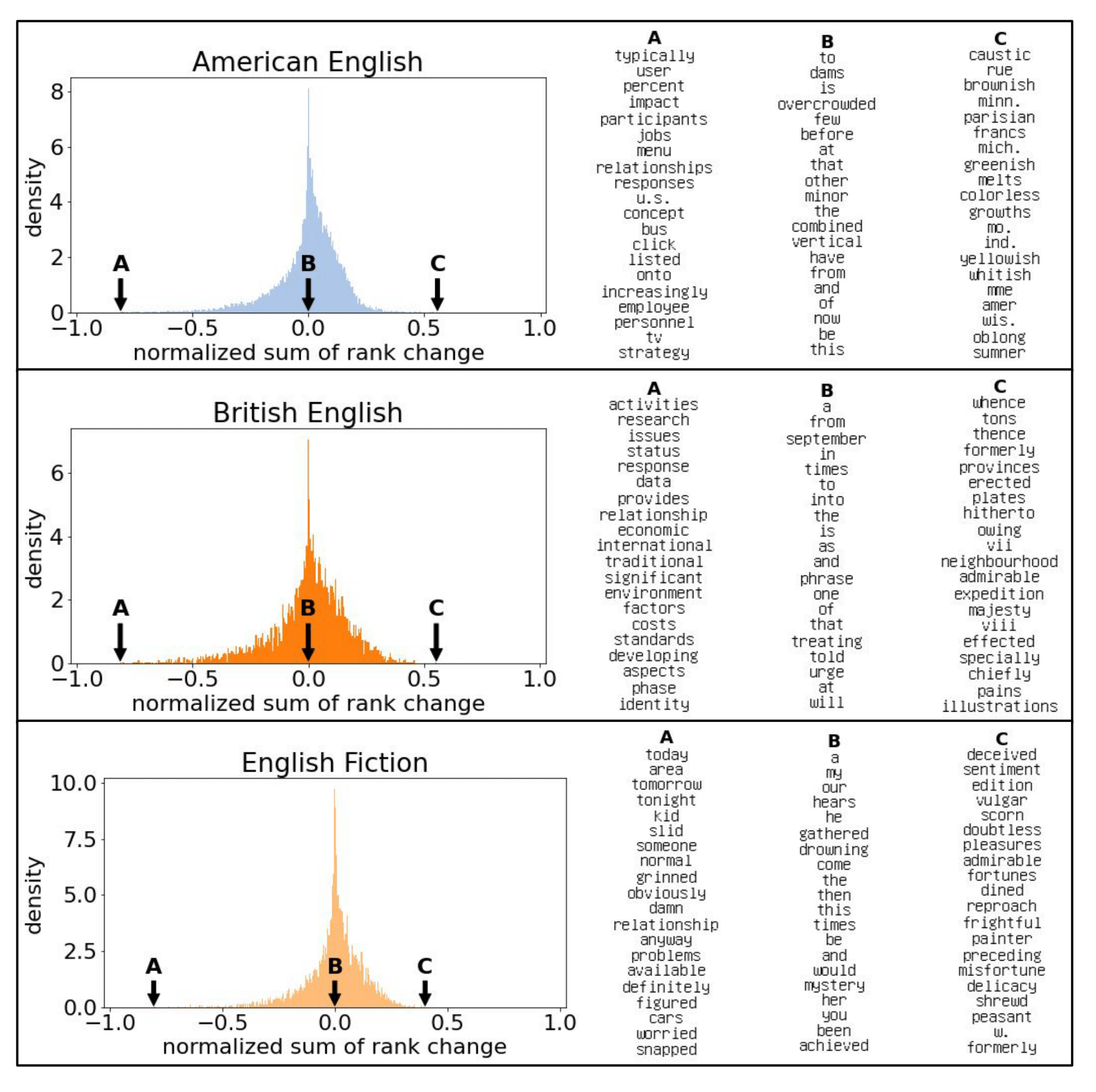}

\section*{S11 Figure}
\label{S11_Fig}
\textbf{The sum of rank change distributions of the languages French, Italian, and Spanish.} Each distribution below are the sums of the rank change of the words in the languages in the Google Ngram data. For each language, the distributions are annotated on the left tail (\textbf{A}), center (\textbf{B}), and right tail (\textbf{C}) of the distribution to show the list of words corresponding to the values of the sum. The words in list \textbf{A} are words that changed up in ranks. The words in list \textbf{B} are words that have little or no change in ranks. The words in list \textbf{C} are words that changed down in ranks.
\includegraphics[width=\textwidth]{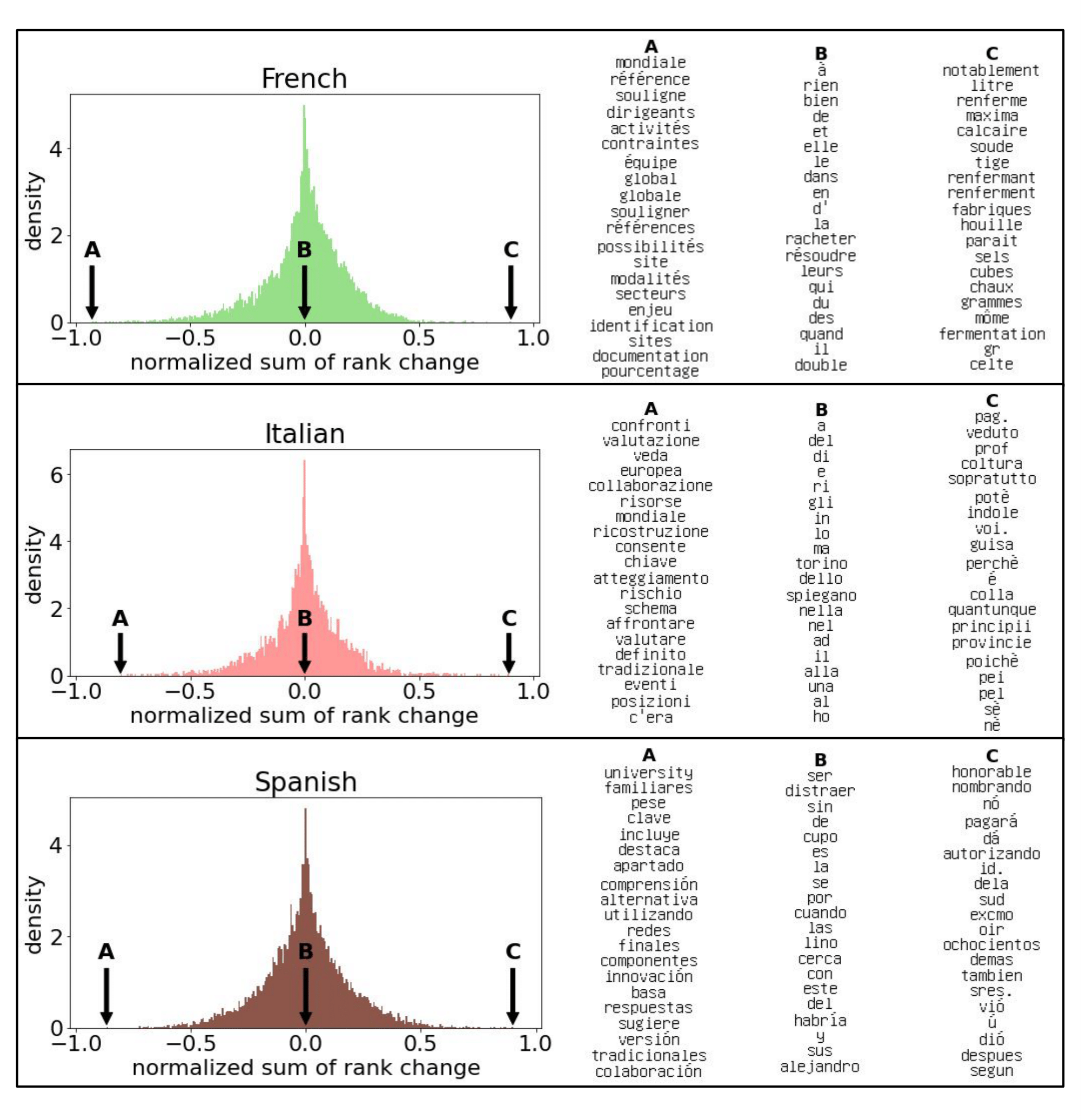}

\section*{S12 Figure}
\label{S12_Fig}
\textbf{The sum of rank change distributions of the languages German and Russian.} Each distribution below are the sums of the rank change of the words in the languages in the Google Ngram data. For each language, the distributions are annotated on the left tail (\textbf{A}), center (\textbf{B}), and right tail (\textbf{C}) of the distribution to show the list of words corresponding to the values of the sum. The words in list \textbf{A} are words that changed up in ranks. The words in list \textbf{B} are words that have little or no change in ranks. The words in list \textbf{C} are words that changed down in ranks.
\includegraphics[width=\textwidth]{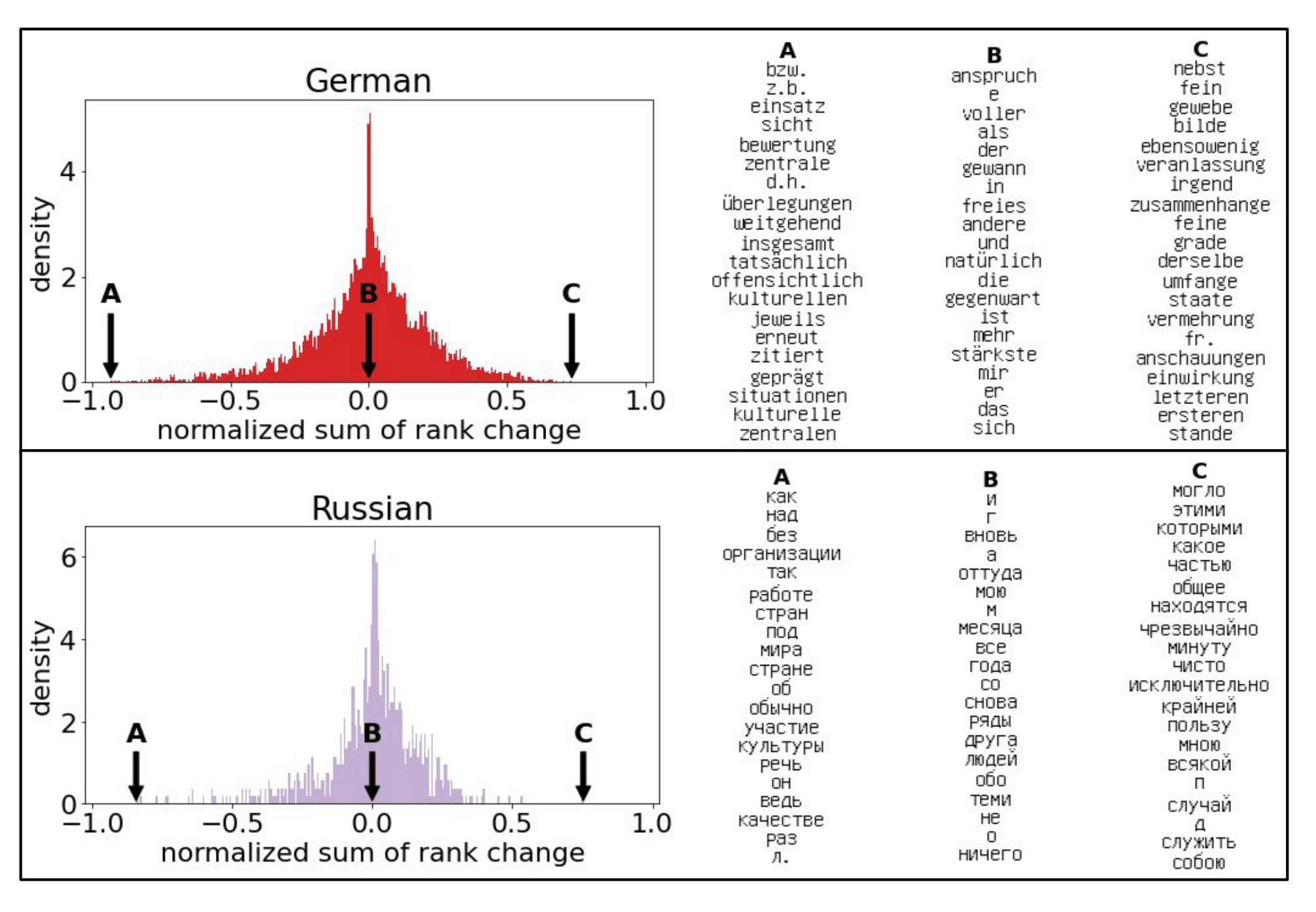}

\section*{S13 Figure}
\label{S13_Fig}
\textbf{The sum of rank change distributions of the languages Hebrew and Simplified Chinese.} Each distribution below are the sums of the rank change of the words in the languages in the Google Ngram data. For each language, the distributions are annotated on the left tail (\textbf{A}), center (\textbf{B}), and right tail (\textbf{C}) of the distribution to show the list of words corresponding to the values of the sum. The words in list \textbf{A} are words that changed up in ranks. The words in list \textbf{B} are words that have little or no change in ranks. The words in list \textbf{C} are words that changed down in ranks.
\includegraphics[width=\textwidth]{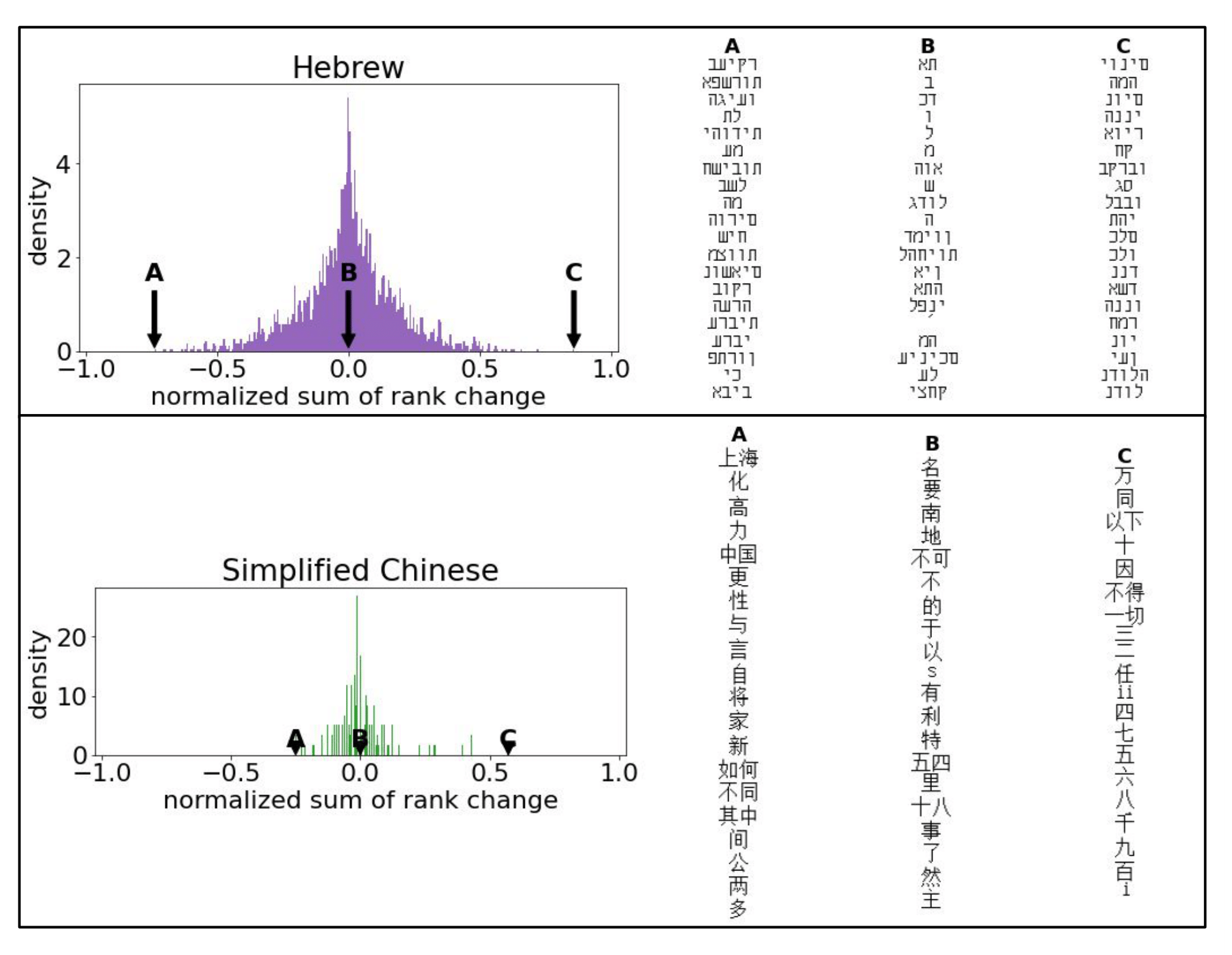}

\section*{S14 Figure}
\label{S14_Fig}
\textbf{The rank change variance distributions of the languages American English, English Fiction, and British English.} Each distribution below are the rank change variances of the words in the languages in the Google Ngram data. For each language, the distributions are annotated on the left tail (\textbf{A}), center (\textbf{B}), and right tail (\textbf{C}) of the distribution to show the list of words corresponding to the values of the variance. The words in list \textbf{A} are words that have little or no variance in their rank change. The words in list \textbf{B} are words with average variances. The words in list \textbf{C} are words that have high variances in their rank change.
\includegraphics[width=\textwidth]{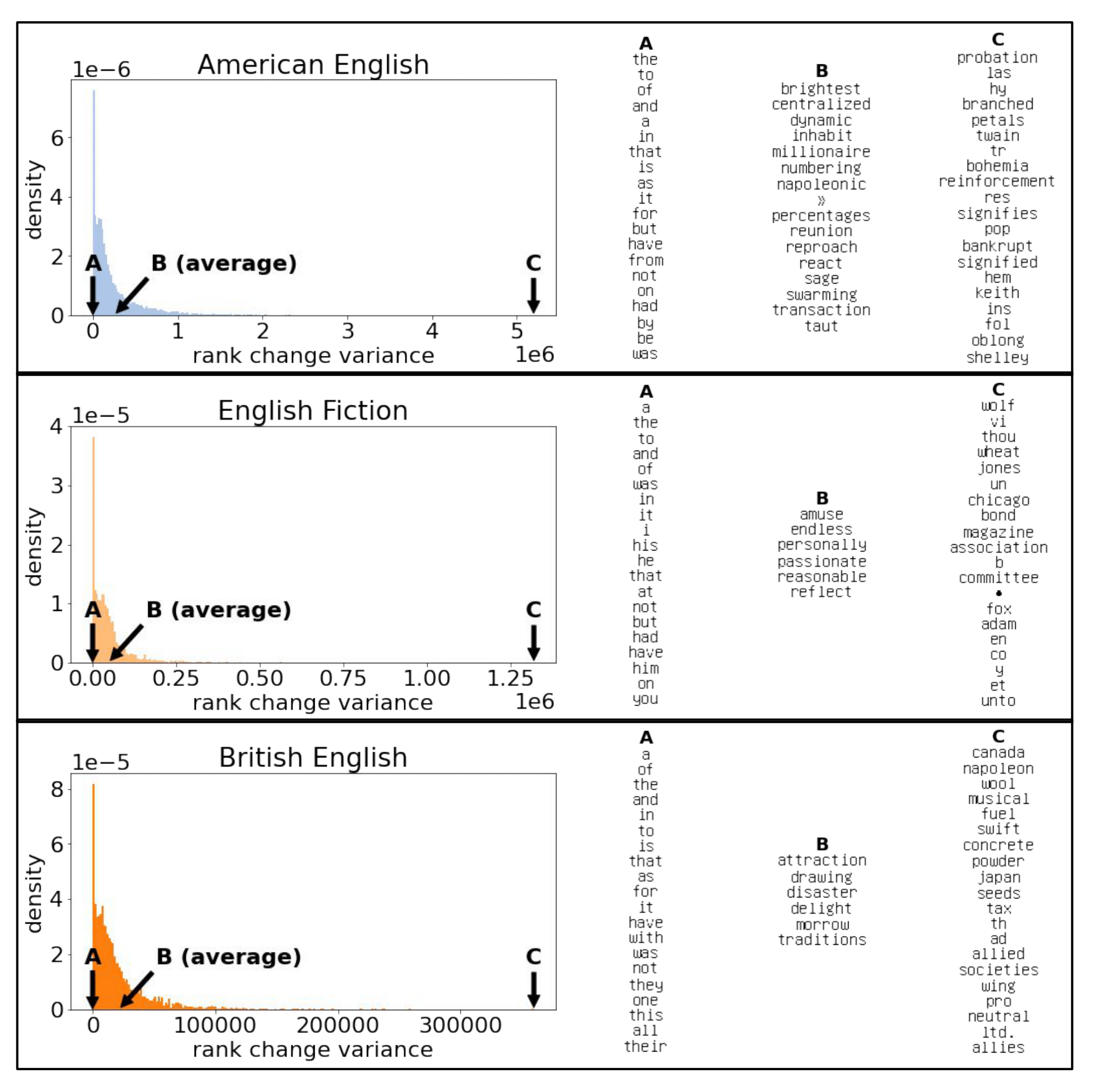}

\section*{S15 Figure}
\label{S15_Fig}
\textbf{The rank change variance distributions of the languages French, Italian, Spanish.} Each distribution below are the rank change variances of the words in the languages in the Google Ngram data. For each language, the distributions are annotated on the left tail (\textbf{A}), center (\textbf{B}), and right tail (\textbf{C}) of the distribution to show the list of words corresponding to the values of the variance. The words in list \textbf{A} are words that have little or no variance in their rank change. The words in list \textbf{B} are words with average variances. The words in list \textbf{C} are words that have high variances in their rank change.
\includegraphics[width=\textwidth]{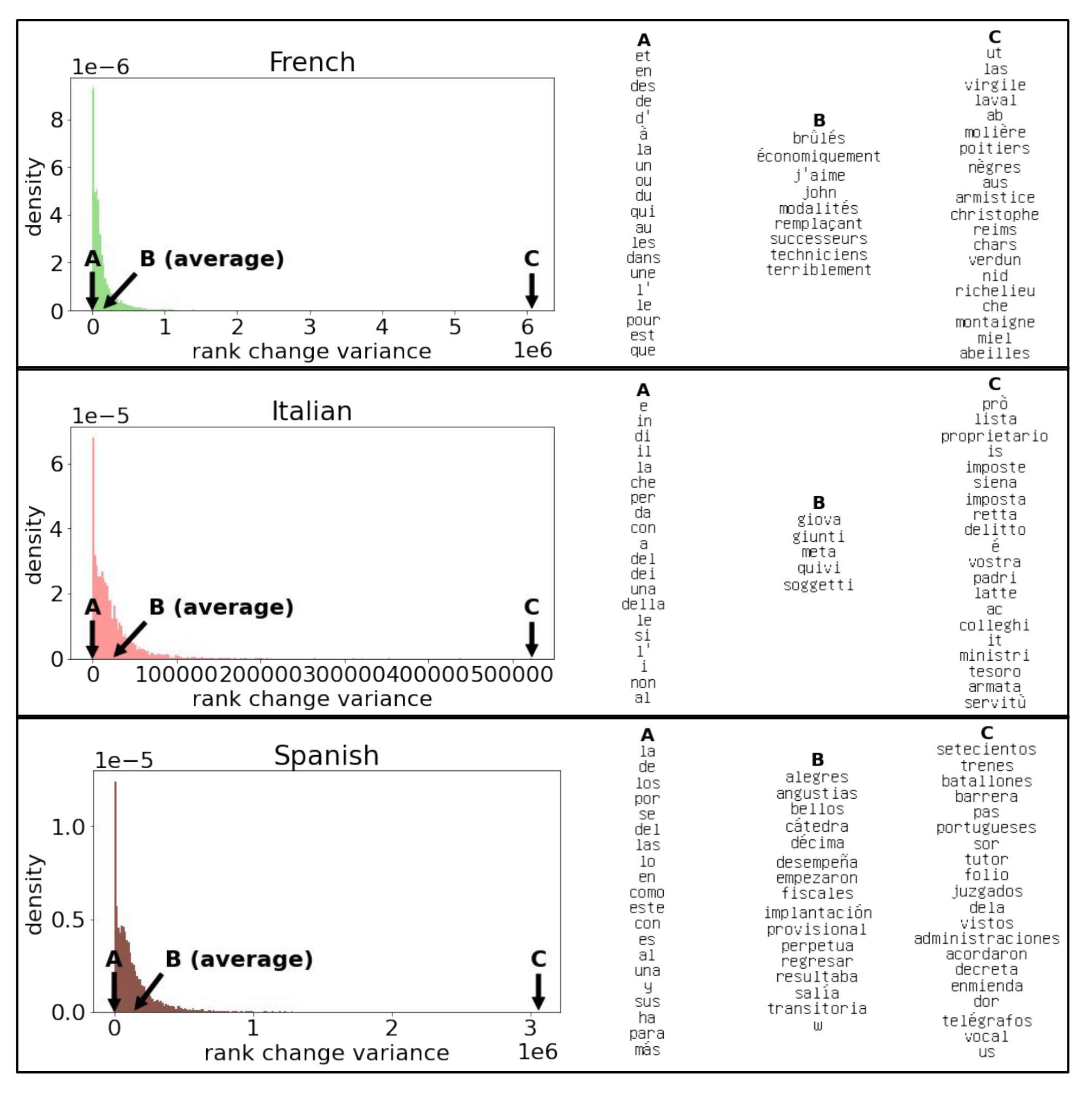}

\section*{S16 Figure}
\label{S16_Fig}
\textbf{The rank change variance distributions of the languages German and Russian} Each distribution below are the rank change variances of the words in the languages in the Google Ngram data. For each language, the distributions are annotated on the left tail (\textbf{A}), center (\textbf{B}), and right tail (\textbf{C}) of the distribution to show the list of words corresponding to the values of the variance. The words in list \textbf{A} are words that have little or no variance in their rank change. The words in list \textbf{B} are words with average variances. The words in list \textbf{C} are words that have high variances in their rank change.
\includegraphics[width=\textwidth]{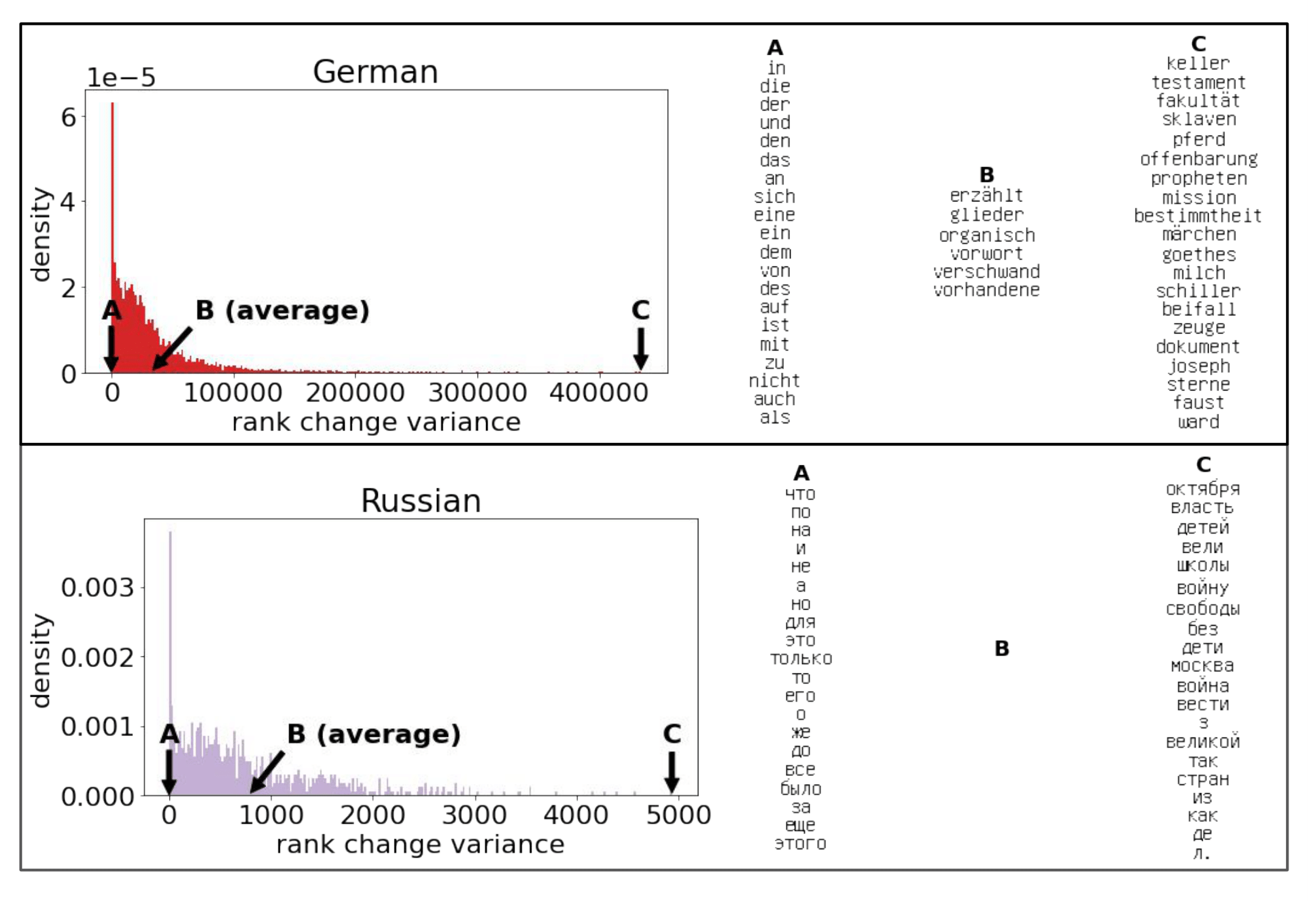}

\section*{S17 Figure}
\label{S17_Fig}
\textbf{The rank change variance distributions of the languages Hebrew and Simplified Chinese.} Each distribution below are the rank change variances of the words in the languages in the Google Ngram data. For each language, the distributions are annotated on the left tail (\textbf{A}), center (\textbf{B}), and right tail (\textbf{C}) of the distribution to show the list of words corresponding to the values of the variance. The words in list \textbf{A} are words that have little or no variance in their rank change. The words in list \textbf{B} are words with average variances. The words in list \textbf{C} are words that have high variances in their rank change.
\includegraphics[width=\textwidth]{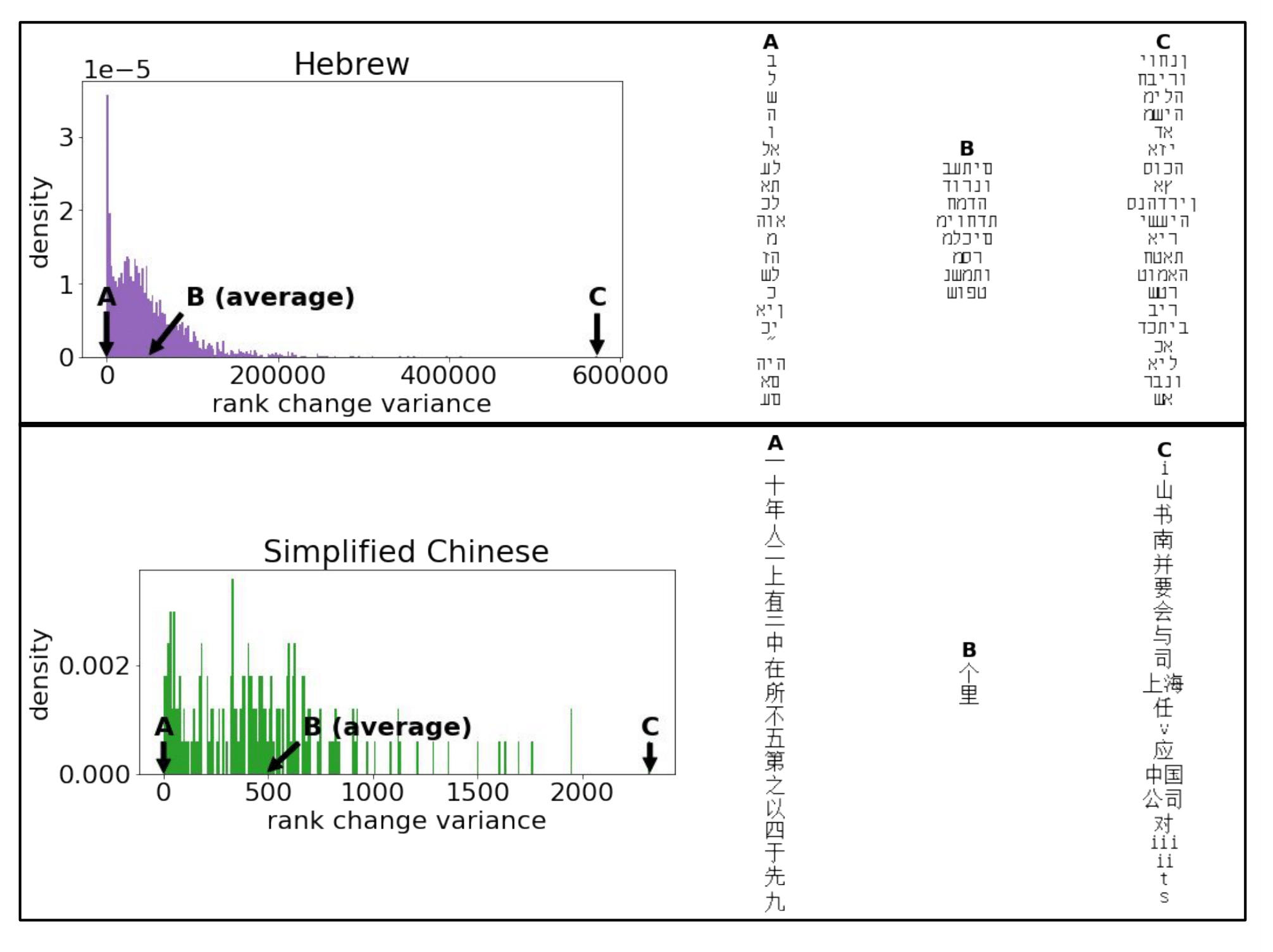}

\section*{S18 Figure}
\label{S18_Fig}
\textbf{Turnover curves of the WF inspired model simulations.} This figure shows the turnover curves for different values of initial corpus size $\beta$ of the WF inspired model. Subfigs from (a)-(d) corresponds to increasing $\beta$, which also means decreasing ratio $c/\beta$. We fit the shape parameter $b$ (dashed lines) of the generic turnover function $z=ay^b$ for each simulation (gray dots) and found that there is an association between the ratio $c/\beta$ and $b$. The turnover curves went from anti-conformist to conformist as $c/\beta$ decreases. Each figure shows the curve (dotted line) if $b=0.86$ (unbiased copying).
\includegraphics[width=\textwidth]{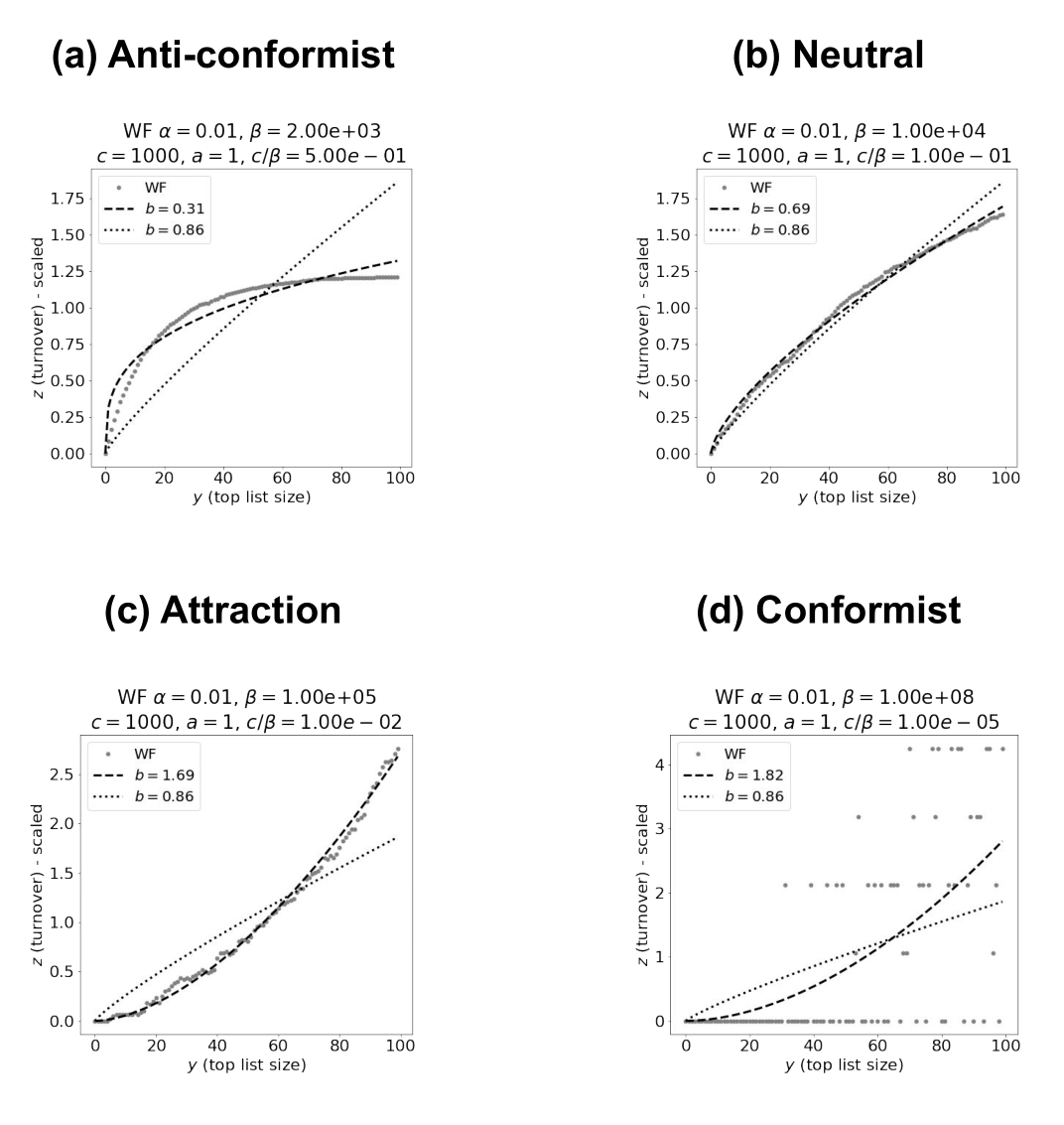}

\section*{S19 Figure}
\label{S19_Fig}
\textbf{Turnover curves of the languages.} This figure shows the turnover curves for the languages. We fit the shape parameter $b$ (dashed lines) of the generic turnover function $z=ay^b$ for each language (colored) and found that all of them shows an attraction turnover curve ($b > 0.86$) except for Simplified Chinese.
\includegraphics[width=\textwidth]{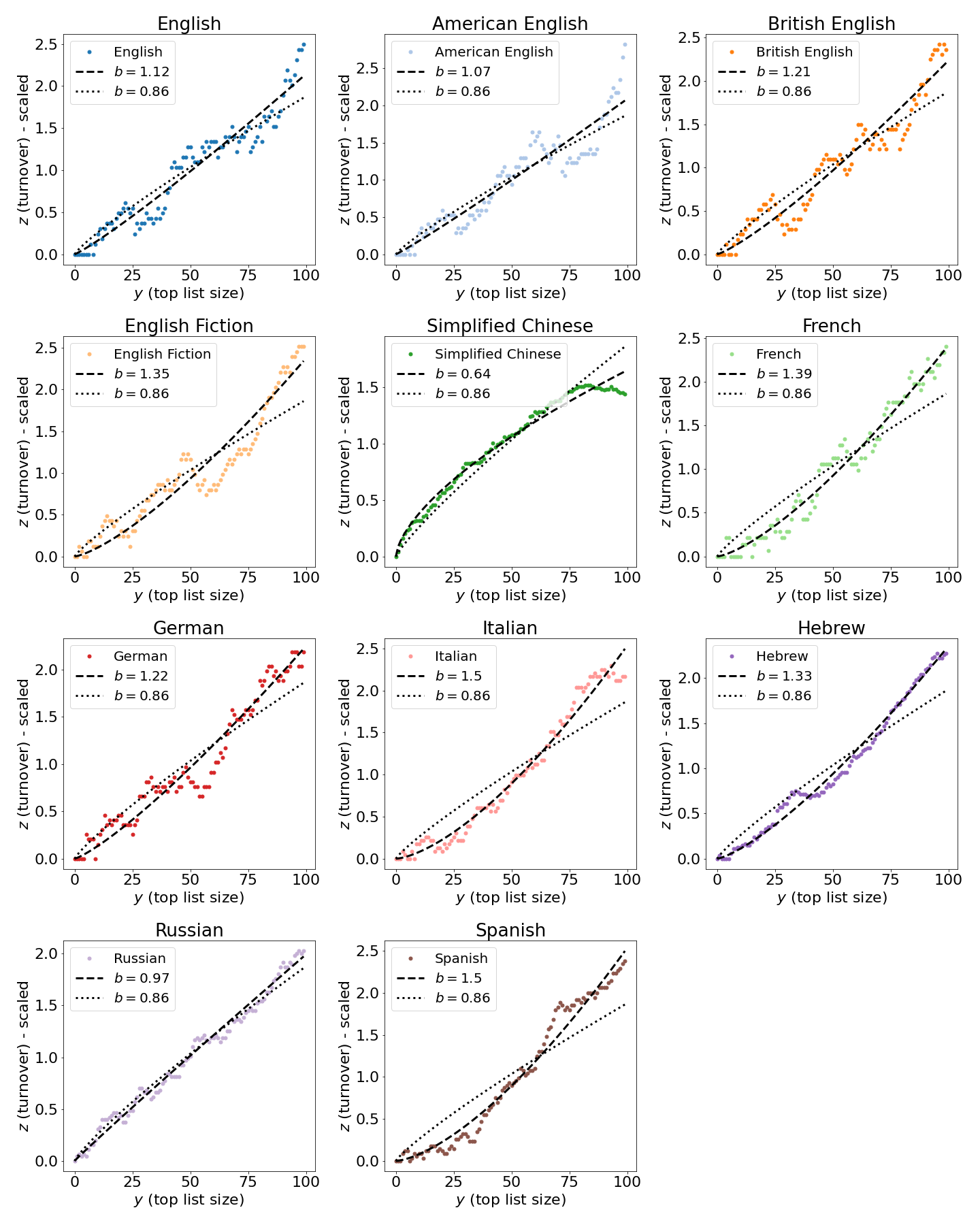}

\section*{Acknowledgments}
We thank the two reviewers for their thorough examination of our paper, as well as their many thoughtful comments and suggestions.

\nolinenumbers

\bibstyle{plos2015}
\bibliography{svv-bibliography}

\end{document}